\documentclass{report}

\usepackage[latin1]{inputenc}
\usepackage[spanish]{babel}
\usepackage{algorithm2e}
\usepackage{latexsym}
\usepackage{amssymb}
\usepackage{epsfig}
\usepackage{subfig}
\usepackage{geometry}
\usepackage{rotating}
\usepackage{titlesec}

\usepackage{graphicx}
\usepackage{color}
\usepackage[all]{xy}
\usepackage{appendix}
\usepackage{multirow}
\usepackage{float}

\definecolor{red}{rgb}{0.8,0.2,0.2}
\definecolor{blue}{rgb}{0,0,0.5}
\definecolor{green}{rgb}{0,0.7,0}
\definecolor{violet}{rgb}{0.5,0.2,0.5}
\definecolor{orange}{rgb}{0.8,0.5,0.2}

\titleformat{\chapter}{\normalfont\huge\bfseries}{\thechapter}{20pt}{\huge}
\titleformat{\section} {\normalfont\Large\bfseries}{\thesection}{1em}{} 
\titleformat{\subsection}{\normalfont\large\bfseries}{\thesubsection}{1em}{} 

\def\C{\cal C}
\def\F{\cal F}

\spanishdecimal{.}

\long\def\comment#1{}

\author{Jorge Alonso Bedoya Puerta}
\title{Aplicaci\'on de distancias entre t\'erminos para datos planos y jer\'arquicos}

\begin{document}

\renewcommand{\contentsname}{Tabla de Contenido}
\renewcommand{\tablename}{Tabla}
\renewcommand{\listtablename}{Lista de Tablas}
\renewcommand{\figurename}{Figura}
\renewcommand{\listfigurename}{Lista de Figuras}
\renewcommand{\appendixname}{Ap\'endices} 
\renewcommand{\appendixtocname}{Ap\'endices} 
\renewcommand{\appendixpagename}{Ap\'endices}

\begin{titlepage}
 \centering

\Large
 UNIVERSIDAD POLIT\'ECNICA DE VALENCIA \\

\small DEPARTAMENTO DE SISTEMAS INFORM\'ATICOS Y COMPUTACI\'ON \\
\vspace{1cm}
\includegraphics[height=3cm]{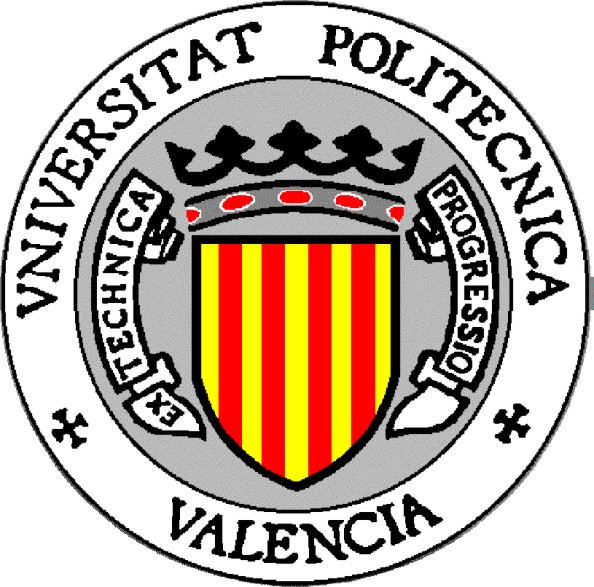}

\begin{center}
\rule{1\textwidth}{1pt}
\end{center}

\LARGE
Aplicaci\'on de distancias entre t\'erminos para datos planos y jer\'arquicos\\

\begin{center}
\rule{1\textwidth}{1pt}
\end{center}

Jorge Alonso Bedoya Puerta\\

\vspace{1cm}

\large  Tesis presentada para optar al t\'itulo de:\\

\vspace{0.5cm}

M\'aster en ingenier\'ia de software, m\'etodos formales y sistemas de informaci\'on\\

\vspace{1.5cm}

Dirigida por:\\

\vspace{0.5cm}

Jos\'e Hern\'andez Orallo\\

\vspace{2cm}

Valencia, septiembre de 2011

\end{titlepage}

 \thispagestyle{empty}

\vspace*{1cm}

\begin{flushright}

En memoria de mi abuela:\\
Mar\'ia Jes\'us Gonz\'alez \'Angel

\end{flushright}

\vspace*{\fill}

\chapter*{Resumen }

\setlength{\parskip}{\baselineskip}

\setcounter{page}{1} 
\pagenumbering{roman}

El aprendizaje autom\'atico se basa en el dise\~no de algoritmos que utilizan la informaci\'on hist\'orica para aprender. Algunos de estos algoritmos usan las distancias para medir el grado de similitud (o disimilitud) entre objetos; estos m\'etodos, basados en distancias, son un enfoque muy conocido y potente para la inferencia inductiva, ya que las distancias son una forma adecuada para medir la disimilitud. Com\'unmente, todos estos algoritmos, basados en distancias y otros enfoques, utilizan informaci\'on que es representada mediante datos proposicionales (tuplas de componentes num\'ericos o categ\'oricos, los cuales son f\'acilmente representados como una tabla); sin embargo, este tipo de representaciones pueden ser un poco restrictivas y, en muchas ocasiones, se requiere de estructuras m\'as complejas que permitan representar los datos de una manera m\'as natural. 

Por otra parte,  los t\'erminos son la base para la representaci\'on de la programaci\'on l\'ogica y funcional. A la vez, la programaci\'on l\'ogica y la programaci\'on funcional pueden ser utilizadas para la representaci\'on del conocimiento en variedad de aplicaciones (sistemas basados en el conocimiento, miner\'ia de datos, etc.). Las distancias entre t\'erminos son una herramienta muy \'util, no solo para comparar t\'erminos, sino tambi\'en para determinar el espacio de b\'usqueda en muchas de esas aplicaciones.

Este trabajo aplica las distancias entre t\'erminos, aprovechando las caracter\'isticas propias de cada distancia y la posibilidad de comparar desde tipos de datos proposicionales hasta representaciones jer\'arquicas. Las distancias entre t\'erminos se aplican por medio del algoritmo de clasificaci\'on $k$-NN ($k$-vecinos m\'as cercanos) utilizando un lenguaje de representaci\'on com\'un; el lenguaje utilizado es XML, que adem\'as de su popularidad, ofrece una alta flexibilidad para representaci\'on de datos proposicionales y t\'erminos desde diferentes niveles de jerarqu\'ia. Para poder representar estos datos en una estructura XML y con el fin de aprovechar los beneficios de las distancias entre t\'erminos, es necesario aplicar algunas transformaciones. Estas transformaciones permiten convertir datos planos en datos jer\'arquicos, representados en XML, aplicando algunas t\'ecnicas basadas en asociaciones intuitivas entre los valores y nombres de las variables y asociaciones basadas en la similitud de los atributos.

Adicionalmente, es posible aplicar estas distancias entre t\'erminos sobre estructuras originalmente jer\'arquicas. Conservando XML como lenguaje de representaci\'on, el proceso de transformaci\'on consiste, principalmente, en mantener el nivel de profundidad de los nombres de las variables y sus valores y garantizar el orden, de acuerdo con caracter\'isticas de formaci\'on de XML.  

Consecuentemente, en este trabajo se realizan experimentos con la distancia entre t\'erminos de Nienhuys-Cheng \cite{NW97} y la distancia de Estruch et al. \cite{estruch2010integrated}. Para el caso de datos originalmente proposicionales, estas distancias son comparadas con la distancia eucl\'idea. En todos los casos, los experimentos se realizan con distancias ponderadas del algoritmo de $k$-vecinos m\'as cercanos, utilizando varios exponentes para la funci\'on de {\em atracci\'on} (distancia ponderada). Se puede ver que, en algunos casos, estas distancias entre t\'erminos pueden mejorar significativamente los resultados sobre enfoques aplicados para representaciones planas. Para el caso de estructuras originalmente jer\'arquicas, tambi\'en se pueden resaltar diferencias obtenidas en el proceso de clasificaci\'on dependiendo del tipo de distancia (entre t\'erminos) aplicada.

 \vspace{1cm}

\noindent \textbf{Palabras clave:} funciones de distancia, clasificaci\'on, representaci\'on basada en t\'erminos, datos jer\'arquicos, datos XML, programaci\'on inductiva.

\chapter*{Abstract }

Machine learning is based on the design of algorithms that use historical information to learn. Some of these algorithms use distances to measure the degree of similarity (or dissimilarity) between objects; these methods, based on distances, are a well known and powerful approach to inductive inference, since the distances are a proper way to measure dissimilarity. Usually, these distance-based algorithms, and other approaches, use information that is represented by propositional data (tuples of numerical or categorical components, which are easily represented as a table), however, this kind of representation can be quite restrictive and, in many cases, it requires more complex structures in order to represent data in a more natural way.

On the other hand, the terms are the basis for functional and logic programming representation. At the same time, logic and functional programming can be used for knowledge representation in a variety of applications (knowledge-based systems, data mining, etc.). Distances between terms are a useful tool not only to compare terms, but also to determine the search space in many of these applications.

This dissertation applies distances between terms, exploiting the features of each distance and the possibility to compare from propositional data types to hierarchical representations. The distances between terms are applied through the $k$-NN ($k$-nearest neighbor) classification algorithm using a common language representation; the language used is XML which, in addition to its popularity, offers high flexibility for propositional data and term representation from different  hierarchy levels. To be able to represent these data in an XML structure and to take advantage of the benefits of distance between terms, it is necessary to apply some transformations. These transformations allow the conversion of flat data into hierarchical data represented in XML, using some techniques based on intuitive associations between the names and values of variables and associations based on attribute similarity.

Additionally, it is possible to apply these distances between terms over originally hierarchical structures. Preserving XML as a representation language, the transformation process is mainly to maintain the depth level of the variables' names and their values and ensure the order, according to XML formation features.

Consequently, several experiments with the distances between terms of Nienhuys-Cheng \cite{NW97} and Estruch et al. \cite{estruch2010integrated} were performed in this dissertation. In the case of originally propositional data, these distances are compared to the Euclidean distance. In all cases, the experiments were performed with the distance-weighted $k$-nearest neighbor algorithm, using several exponents for the {\em  attraction} function (weighted distance). It can be seen that in some cases, the term distances can significantly improve the results on approaches applied to flat representations. In the case of originally hierarchical structures, differences obtained in the classification process depending on the distance type applied (between terms) can also be highlighted.

\vspace{1cm}

\noindent \textbf{Keywords:} Distance functions, classification, term-based representation, hierarchical data, XML data, inductive programming.

\chapter*{Agradecimientos}

\setlength{\parskip}{\baselineskip}

\vspace{-0.6cm}

Este trabajo ha sido realizado con la ayuda, apoyo, comprensi\'on y paciencia de muchas personas y ser\'ia injusto no hacer un reconocimiento a todos quienes, desde diferentes \'ambitos, han aceptado compartir conmigo las dificultades propias de un trabajo de investigaci\'on de fin de m\'aster, m\'axime con las implicaciones del desarraigo cultural que lo hace m\'as complejo, pero al mismo tiempo m\'as satisfactorio.

\vspace{-0.1cm}
A mi madre y mi padre por convertir mis sue\~nos en sus sue\~nos, con todas las consecuencias que esto conlleva. 

\vspace{-0.1cm}
A Jos\'e Hern\'andez Orallo, por aceptar ser mi asesor, en el sentido pleno de la palabra; su calidez humana, conocimiento, disposici\'on, apoyo, diligencia y paciencia han hecho posible que, no s\'olo finalice este proceso satisfactoriamente, sino que ampl\'ie mi conocimiento profesional.

\vspace{-0.1cm}
A Mar\'ia Jos\'e Ram\'irez y C\'esar Ferri, por brindar su experiencia y conocimiento para apoyar activamente este proyecto y ayudarme a mejorarlo.

\vspace{-0.1cm}
A mis amigos en Valencia: Virginia, Antonio, Fabio, Oscar y Pablo por su generosidad, amistad y hospitalidad durante todo este proceso, haci\'endome sentir un miembro de sus familias.  Gracias a ellos, no ha sido tan dif\'icil estar lejos de casa.

\vspace{-0.1cm}
A mi hermano por su consejo simple pero contundente que permite simplificar las dificultades cuando las cosas no van bien.

\vspace{-0.1cm}
A todos los familiares, amigos y compa\~neros del trabajo que de alguna manera me motivaron en este proyecto y de diferentes maneras me apoyaron.

\vspace{-0.1cm}
Y por supuesto a Gloria, mi novia, esta bella persona que con su amor y compa\~n\'ia hace que todo sea m\'as sencillo. 

\vspace{0.5cm}

Agradezco igualmente a Arkix S.A., por permitirme continuar creciendo desde situaciones pr\'acticas; al grupo DMIP ({\em Data Mining and Inductive Programming}) por acogerme en su equipo, apoyarme  y hacerme parte de sus actividades y a la Universidad Polit\'ecnica de Valencia, por ser mi casa lejos de casa y permitirme crecer como ser humano y como profesional.

\vspace*{1cm}

\begin{flushright}

Jorge A. Bedoya P.\\
Valencia, 2011

\end{flushright}

\tableofcontents

\listoftables
\listoffigures

\chapter{Introducci\'on}

\setlength{\parskip}{\baselineskip}

\pagenumbering{arabic}
\setcounter{page}{1} 

\vspace{0.2cm}

\section{Motivaci\'on}
La informaci\'on hist\'orica es fundamental para la aplicaci\'on de algoritmos de aprendizaje autom\'atico; esta informaci\'on puede ser representada de m\'ultiples formas que van desde estructuras r\'igidas y estrictamente ordenadas (por ejemplo datos tabulares) hasta representaciones m\'as naturales (textos, p\'aginas web, etc.). Muchos de estos m\'etodos de aprendizaje, t\'ipicos en aplicaciones de miner\'ia de datos, est\'an dise\~nados para trabajar con representaci\'on de datos proposicionales que consisten en tuplas de atributos num\'ericos o categ\'oricos, com\'unmente representados como una tabla. Los resultados obtenidos por la aplicaci\'on de estos algoritmos se ven afectados, entre otros aspectos, por la manera en que se representan los datos; en muchas ocasiones, los datos deben ser aplanados para ajustarse a representaciones proposicionales; lo anterior hace que se requiera de un lenguaje de representaci\'on m\'as expresivo basado en tipos de datos estructurados como conjuntos, \'arboles, grafos, etc. Dado que el aprendizaje proposicional no se puede tratar con estos tipos de datos, nuevos m\'etodos han sido dise\~nados para extraer patrones desde descripciones m\'as complejas \cite{estruchPhd2009}.

Un subcampo del aprendizaje autom\'atico, relacionado con el aprendizaje a partir de la l\'ogica de primer orden, es conocido como programaci\'on l\'ogica inductiva (ILP), que puede ser aplicada para la representaci\'on de la informaci\'on de una forma m\'as natural y expresiva que otros paradigmas, que solo aprenden de ejemplos con estructuras planas. Este \'area ha sido extendida a otros paradigmas, como la programaci\'on funcional, dando lugar a la programaci\'on funcional inductiva o a la programaci\'on l\'ogico-funcional inductiva \cite{DBLP:conf/agp/Hernandez-OralloR98} \cite{DBLP:conf/ilp/Hernandez-OralloR99} \cite{DBLP:conf/flops/Ferri-RamirezHR01} o al \'area general de programaci\'on inductiva \cite{partridge1997case} \cite{flener2001inductive} \cite{hofmannunifying}. En programaci\'on l\'ogica y en programaci\'on funcional, los datos se representan en base a una estructura de datos que se denomina t\'ermino, a partir de los que se definen los \'atomos , cl\'ausulas o reglas y, en definitiva, los programas.

El aprendizaje basado en instancias utiliza el principio del razonamiento basado en casos (CBR), donde ``problemas similares tienen soluciones similares''. Este aprendizaje utiliza la distancia para conocer la similitud (o disimilitud) entre instancias. Las distancias (tambi\'en llamadas m\'etricas) son medidas de disimilitud con algunas propiedades especiales, como la simetr\'ia y la desigualdad triangular, las cuales son muy ventajosas para muchos algoritmos porque el espacio de b\'usqueda puede ser reducido a una triangulaci\'on. Los m\'etodos basados en distancias son muy eficaces para la inferencia inductiva; la gran ventaja de estos m\'etodos es que tambi\'en son algoritmos o t\'ecnicas que pueden ser aplicadas sobre diferentes ordenes de tipos de datos, siempre que una funci\'on de similitud haya sido definida previamente para estos tipos de datos \cite{Mit97}.

Por otra parte, existen lenguajes muy populares que se basan en \'arboles o jerarqu\'ias, como XML y estructuras relacionadas funcionalmente similares \cite{Cheney08}, que pueden ser utilizados para representar la informaci\'on o el conocimiento (por ejemplo ontolog\'ias). Las estructuras en \'arboles y los t\'erminos funcionales tienen fuertes similitudes; un t\'ermino en programaci\'on funcional (o l\'ogica) puede ser representado como un \'arbol ordenado. Aunque la sem\'antica es crucial para entender el rol de un t\'ermino o \'atomo con relaci\'on a un programa, tambi\'en es posible analizar t\'erminos de manera aislada (analizando \'unicamente su parte sint\'actica). 

Existen distancias para virtualmente cualquier tipo de objeto, incluyendo complejos o altamente estructurados, como tuplas, listas, \'arboles, grafos, im\'agenes, sonidos, p\'aginas web, ontolog\'ias, etc. Un nuevo reto en el aprendizaje autom\'atico, pero m\'as especialmente en el \'area de la programaci\'on inductiva, es la distancia entre t\'erminos o \'atomos de primer orden. Aunque los t\'erminos pueden ser usados para representar muchos de los tipos de datos previos (y consecuentemente, una distancia entre t\'erminos virtualmente llega a ser una distancia para cualquier dato complejo y estructurado), estos tipos de datos son especialmente adaptados para representaciones basadas en t\'erminos o \'arboles. De esta manera, estas distancias no solo pueden ser usadas en el \'area de la programaci\'on l\'ogica inductiva (ILP) \cite{Mu91} (por ejemplo segmentaci\'on de primer orden \cite{BRR98} ) o en general, programaci\'on inductiva, sino tambi\'en para \'areas donde est\'a involucrada la informaci\'on estructurada (jer\'arquica) como aprendizaje de ontolog\'ias o documentos XML. 

En comparaci\'on con las distancias tradicionales, las distancias entre t\'erminos cuentan con propiedades adicionales (como la sensibilidad del contexto, el tama\~no de las diferencias, las diferencias repetidas, etc.) que favorecen la comparaci\'on de estructuras jer\'arquicas; en esta medida, estas distancias, permiten comparar desde tipos de datos proposicionales hasta estructuras jer\'arquicas. Una manera de lograr mejores  resultados en las tareas de aprendizaje basado en distancias es aplicando transformaciones que conviertan datos proposicionales en jer\'arquicos o mantener la jerarqu\'ia original y as\'i obtener un mayor provecho de las propiedades de las distancias entre t\'erminos.

\vspace{0.3cm}

\section{Objetivos}

\vspace{0.2cm}

El prop\'osito general de este trabajo es aplicar distancias entre t\'erminos sobre conjuntos de datos jer\'arquicos  que, por medio de transformaciones apropiadas sobre datos planos, puedan ser utilizadas en una gama m\'as amplia de aplicaciones y permitan mejorar los procesos de clasificaci\'on.  Espec\'ificamente,  se utilizan las distancias entre t\'erminos de Nienhuys-Cheng y Estruch et al. para el algoritmo de $k$-vecinos m\'as cercanos, aplicadas para datos categ\'oricos sobre dos tipos de representaciones: primero, jerarqu\'ias construidas a partir de datos planos, utilizando relaciones entre atributos inducidas por nombre o igualdad de valor y m\'etricas de similitud entre atributos; los resultados son comparados utilizando la distancia eucl\'idea; segundo, datos originalmente jer\'arquicos. Para estas dos situaciones se utiliza XML como lenguaje de representaci\'on.

\noindent Lo anterior es detallado en los siguientes objetivos espec\'ificos:

\begin{itemize}
\item Estudiar y entender las propiedades de las distancias entre t\'erminos utilizando como referencia de inicio de este trabajo el art\'iculo {\em A New Context-Sensitive and Composable Distance for First-Order Terms. Technical Report} \cite{estruchnew}, revisando las distancias entre t\'erminos, particularmente la distancias de  Nienhuys-Cheng y Estruch et al. y sus propiedades.

\item Considerar XML como un lenguaje de representaci\'on de datos sobre el cual se calcule la distancia entre t\'erminos, permitiendo as\'i un formato de entrada m\'as est\'andar y flexible para las transformaciones y jerarqu\'ias propuestas. 

\item Implementar las distancias entre t\'erminos y un algoritmo de clasificaci\'on para evaluar dichas distancias. En concreto, se propone implementar las distancias entre t\'erminos y la distancia eucl\'idea.

\item Explorar enfoques de agrupaci\'on de variables para construir jerarqu\'ias sobre datos planos. Partiendo de los posibles enfoques de agrupaci\'on de atributos (inducida por nombre o igualdad de valor y m\'etricas de similitud entre atributos), se crear\'an jerarqu\'ias que posteriormente puedan ser utilizadas por el algoritmo de clasificaci\'on.

\item Aplicaci\'on de distancias sobre conjuntos de datos y an\'alisis de resultados. Con las jerarqu\'ias inducidas a partir de datos planos y los datos originalmente jer\'arquicos, se realizar\'an experimentos sobre los conjuntos de datos, se analizar\'an los resultados devueltos por el algoritmo de clasificaci\'on y se establecer\'an las conclusiones de cada proceso con el fin de entender mejor el comportamiento de estas distancias y de los procesos de transformaci\'on propuestos.
\end{itemize}

\vspace{0.3cm}

\section{Organizaci\'on de la tesis}

\vspace{0.2cm}

\noindent Este trabajo se encuentra organizado de la siguiente manera:

\begin{itemize}
\item \textbf{Cap\'itulo \ref{prelim}: Antecedentes.} Este cap\'itulo describe aspectos generales del aprendizaje autom\'atico, caracter\'isticas y propiedades de los m\'etodos basados en distancias, algoritmos basados en distancias (particularmente el algoritmo $k$-NN y algunas mejoras) y distancias para diferentes tipos de datos estructurados, representaci\'on de t\'erminos usando XML y, finalmente, se describen las propiedades de algunas distancias entre t\'erminos y su definici\'on formal (profundizando en la distancia de Nienhuys-Cheng y Estruch et al.).

\item \textbf{Cap\'itulo \ref{transformaciones}: Transfomaci\'on de datos semi-estructurados en una representaci\'on basada en t\'erminos usando XML.} En el cap\'itulo \ref{transformaciones} se hace una descripci\'on de las transformaciones realizadas para convertir datos planos en datos jer\'arquicos utilizando como lenguaje de representaci\'on com\'un XML. Para este proceso se consideran tres fuentes de estructuras: igualdad de valor, jerarqu\'ia inducida por nombres y jerarqu\'ia de similitud de atributos. Adicionalmente, se describe brevemente algunos aspectos a considerar para la derivaci\'on de esquemas XML jer\'arquicos a partir de datos originalmente jer\'arquicos.   

\item \textbf{Cap\'itulo \ref{experimentos}: Experimentos.} Este cap\'itulo muestra los resultados obtenidos a partir de la ejecuci\'on del algoritmo de clasificaci\'on $k$-NN, utilizando varios exponentes para la funci\'on de {\em atracci\'on} con diferentes tipos de jerarqu\'ias derivadas por medio de las transformaciones mostradas en el cap\'itulo \ref{transformaciones}, aplicando la distancia entre t\'erminos de Nienhuys-Cheng y Estruch et al.  y compar\'andolas con la distancia eucl\'idea. Adicionalmente, se muestran los resultados sobre la aplicaci\'on de distancias entre t\'erminos sobre un conjunto de datos originalmente jer\'arquico.   
\end{itemize}

\noindent Este trabajo recopila, en el \textbf{cap\'itulo \ref{conclusiones}}, las conclusiones obtenidas a partir de los resultados de los experimentos, algunos trabajos futuros y la publicaci\'on realizada.

\noindent Finalmente, se adjuntan dos \textbf{ap\'endices} con el detalle de los resultados del cap\'itulo \ref{experimentos} y una descripci\'on de la implementaci\'on de las distancias entre t\'erminos.

\chapter{Antecedentes}\label{prelim} 

\setlength{\parskip}{\baselineskip}

En este capitulo se describen algunos conceptos b\'asicos del aprendizaje autom\'atico, adem\'as de algunos m\'etodos, representaciones y medidas basadas en la similitud de instancias que sirven para comprender e ilustrar los cap\'itulos porteriores sobre la aplicaci\'on de distancias entre t\'erminos para datos planos y jer\'arquicos.

\section{Aprendizaje autom\'atico}

El aprendizaje autom\'atico es el estudio de m\'etodos computacionales que pueden aprender  y mejorar a partir de la experiencia. Durante los \'ultimos a\~nos, varios m\'etodos de aprendizaje han sido desarrollados y utilizados en muchas aplicaciones pr\'acticas. En \cite {Mit97} se presentan algunos de estos m\'etodos: \'arboles de decisi\'on, redes neuronales, aprendizaje bayesiano y m\'etodos estad\'isticos, aprendizaje basado en instancias, algoritmos gen\'eticos, reglas de decisi\'on y programaci\'on l\'ogica inductiva (ILP); existen otras t\'ecnicas tambi\'en mencionadas por otros autores como son los conjuntos de clasificadores \cite{Dietterich00}  y las m\'aquinas de soporte vectorial \cite{cristianini2002support}.

Dentro de estos m\'etodos es posible encontrar diferentes tipos de algoritmos de acuerdo con su funci\'on de salida o tarea. De ellos se pueden destacar los algoritmos de aprendizaje supervisado y algoritmos de aprendizaje no supervisado. El aprendizaje supervisado produce una funci\'on que establece una correspondencia entre las variables de entrada y la variable de salida deseada; el objetivo de su funci\'on es predecir esta variable de salida. Si la variable de salida es discreta es llamado clasificaci\'on y si la variable de salida es  un valor continuo es regresi\'on. Los algoritmos de aprendizaje no supervisado est\'an conformados \'unicamente por variables de entrada, su objetivo es identificar relaciones entre variables o entre ejemplos; las relaciones entre las variables pueden ser asociaciones, dependencias o correlaciones; las relaciones entre ejemplos son agrupamientos.  

Otro tipo de clasificaci\'on de algoritmos de aprendizaje autom\'atico se basa en la inteligibilidad de los modelos generados. La principal ventaja de un modelo inteligible es que ofrece una explicaci\'on parcial de la realidad proporcionando conocimiento sobre c\'omo las predicciones est\'an hechas. Los algoritmos con esta caracter\'istica pueden ser divididos en dos grupos, dependiendo del lenguaje de representaci\'on que ellos usen. El primer grupo est\'a formado por aprendizaje de \'arboles de decisi\'on y m\'etodos de aprendizaje de reglas cl\'asicas; todos estos emplean reglas proposicionales que tienen una expresividad restrictiva porque no pueden tener variables. Un segundo grupo est\'a compuesto por m\'etodos que usan lenguajes de representaci\'on muy expresivos como cl\'ausulas de Horn de primer orden, que permiten representar las reglas con caracter\'isticas avanzadas como variables, predicados y llamado de funciones. Esta segunda familia es usualmente llamada programaci\'on l\'ogica inductiva (ILP) \cite{ramirez2003multi} o, m\'as en general, programaci\'on inductiva \cite{partridge1997case} \cite{flener2001inductive} \cite{hofmannunifying}. 

Finalmente,  los algoritmos de aprendizaje tambi\'en son clasificados como m\'etodos {\em perezosos} o {\em ansiosos}. Los algoritmos {\em perezosos} son m\'etodos basados en instancias, como por ejemplo el $k$-NN ($k$-vecinos m\'as cercanos). Este tipo de algoritmos utilizan enfoques conceptualmente sencillos para las aproximaciones de valores reales o discretos de las funciones de salida. Aprender en estos modelos consiste en almacenar los datos de entrenamiento presentados y, cuando una nueva instancia es encontrada, un grupo de ejemplos similares relacionados son recuperados de memoria y usados para clasificar la nueva instancia consultada. Una diferencia clave en estos enfoques, con respecto a otros m\'etodos, es que pueden construir una aproximaci\'on diferente de la funci\'on de salida para cada ejemplo que debe ser clasificado. De hecho, muchas t\'ecnicas construyen solo una aproximaci\'on local de la funci\'on de salida que se aplica en la vecindad de una nueva instancia y nunca construyen una aproximaci\'on dise\~nada para tener un buen rendimiento sobre todo el espacio de instancias de entrada. Esto tiene ventajas significantes cuando la funci\'on objetivo es muy compleja y puede ser descrita por una colecci\'on de aproximaciones locales menos complejas \cite {Mit97}.

Una desventaja de los enfoques basados en instancias \cite {Mit97}, es que el costo de clasificaci\'on de un ejemplo nuevo puede ser muy alto; esto se debe al hecho de que casi todo el c\'alculo tiene lugar en tiempo de clasificaci\'on y no cuando los ejemplos de entrenamiento son encontrados previamente. Por lo tanto, las t\'ecnicas para la indexaci\'on eficiente  de ejemplos de entrenamiento son un tema significativamente pr\'actico en la reducci\'on del c\'omputo requerido en el momento de la consulta. Una segunda desventaja para muchos de estos enfoques basados en instancias, especialmente los enfoques del $k$-NN, es que se suelen considerar todos los atributos cuando se trata de recuperar en memoria los ejemplos de entrenamiento similares. Si el valor de salida depende \'unicamente de unos pocos de los muchos atributos disponibles, entonces las instancias que son realmente m\'as ``similares" pueden tener una gran distancia de separaci\'on.

La tabla \ref{tab:clasificacion-algoritmos-aprendizaje} muestra una clasificaci\'on de algunos algoritmos de aprendizaje autom\'atico de acuerdo con su inteligibilidad y la manera en que construyen su funci\'on de salida.

\vspace{0.2cm}

\begin{table}[h]
\begin{center}
\begin{tabular}{|l|l|l|}\hline\hline
   & & \\ 
   &  \makebox[5.5cm][c]{Inteligible}  & \makebox[5.5cm][c]{No inteligible} \\
   &  & \\
\hline\hline
{\em Ansioso} &-\'Arboles de decisi\'on &-Redes neuronales artificiales (ANN) \\
& -Aprendizaje de reglas & -M\'aquinas de soporte vectorial \\
& -Programaci\'on L\'ogica inductiva (ILP) & -M\'etodos Kernel \\
& -$k$-medias & -Clasificadores de bayesianos \\
&  & -Perceptr\'on \\
\hline
{\em Perezoso} &  & -$k$-vecinos m\'as cercanos ($k$-NN) \\
&  & -Razonamiento basado en casos (CBR) \\
&  & -Regresi\'on lineal  ponderada local \\
\hline\hline
\end{tabular}
\end{center}
\caption{Clasificaci\'on de algoritmos de aprendizaje autom\'atico } 
\label{tab:clasificacion-algoritmos-aprendizaje}
\end{table}

\section{M\'etodos basados en distancias}

El razonamiento basado en casos (CBR) es un tipo de m\'etodo basado en instancias que est\'a relacionado con el aprendizaje basado en distancias. CBR supone que problemas similares tienen soluciones similares.  En el aprendizaje autom\'atico es muy com\'un utilizar la similitud entre objetos y existen muchos m\'etodos que se fundamentan en establecer si un nuevo objeto es similar a uno previamente conocido. Una manera de hacerlo es cuantificando la similitud (o disimilitud) entre dos objetos por medio de una medida de distancia. 

M\'as precisamente, una funci\'on $s: X \times  X \rightarrow \mathbb{R}$  se dice que es una funci\'on de similitud si $s(x_i,x_j)$ es mayor cuando los objetos $x_i$ y $x_j$ son m\'as similares. Por ejemplo, a continuaci\'on se define la funci\'on $s$, para atributos nominales, que cuenta el n\'umero de coincidencias ocurridas en la misma posici\'on de una tupla \cite{estruchPhd2009}:

\[
\begin{array}{rcl}
s: X \times  X \ \ \ & \rightarrow & \ \ \ \mathbb{R} \\
s(x_i,x_j) \ \ \ & \rightarrow  & \ \  \displaystyle  \sum_{k=1,2} s^\prime (x_{ik},x_{jk})
\end{array}
\]

\noindent donde $x_i= (x_{i1}, x_{i2})$, $x_j =(x_{j1},x_{j2})$ y 

\[ s^\prime (x_{ik},x_{jk}) =\left\{
\begin{array} {ll}
1, & \ \ \ \mbox{si $x_{ik} = x_{jk}$}\\
0, & \ \ \ \mbox{otro caso}
\end{array}
\right.
\]

\noindent Suponiendo que $x_1=(a,a)$, $x_2=(a,b)$ y $x_3=(b,c)$ entonces $s(x_1,x_2)=1$ y $s(x_1,x_3)= s(x_1,x_3) = 0$. Por lo tanto, $x_1$ y $x_2$ son m\'as similares entre ellos que $x_1$ \'o $x_2$ con respecto a $x_3$. 

Al igual que la similitud, es posible cuantificar la disimilitud de dos objetos. Una funci\'on $s: X \times  X \rightarrow \mathbb{R}$  se dice que es una funci\'on de disimilitud si $d(x_i,x_j)$ es mayor cuando $x_i$ y $x_j$ son menos similares \cite{estruchPhd2009}. 

La relaci\'on entre la similitud y la disimilitud es m\'as clara cuando se trabaja con una funci\'on de similitud (o disimilitud) normalizada \cite{estruchPhd2009}; es decir cuando $0 \leq s, d \leq 1$. Esto adem\'as permite expresar una funci\'on en t\'erminos de la otra; es decir $d=1-s$ \'o $s=1-d$. 

La disimilitud es llamada distancia o m\'etrica $d$ cuando satisface las propiedades de {\em no negatividad}, {\em reflexivilidad}, {\em simetr\'ia} y {\em desigualdad triangular}, definidas de la siguiente manera: 

\begin{enumerate}
\item No negatividad:  $d(x_i, x_j ) \geq  0,  \forall x_i,x_j  \in X$.
\item Reflexividad: $d(x_i, x_j ) = 0 \Leftrightarrow  x_i = x_j$.
\item Simetr\'ia:  $d(x_i, x_j ) = d(x_j, x_i)$.
\item Desigualdad triangular: $d(x_i ; x_j ) \leq d(x_i, x_k) + d(x_k, x_j ), \forall x_i, x_j ; x_k \in X$.
\end{enumerate}

Estas propiedades muestran la interpretaci\'on de la distancia de manera m\'as natural que una funci\'on de disimilitud. La propiedad de no negatividad nos garantiza que la distancia entre dos objetos $x_i$ y $x_j$ siempre es mayor o igual a cero;  la reflexividad asegura que si la distancia entre dos objetos es cero, entonces $x_i$ y $x_j$ son iguales. De igual manera, la simetr\'ia garantiza que la distancia entre dos objetos no depende del orden en que son comparados. Finalmente, si no se cumple la desigualdad triangular puede suceder que se obtienen resultados no deseables, por ejemplo, si los objetos $x_i$ y $x_j$ son muy similares a $x_k$, debe esperarse que $x_i$ y $x_j$ tambi\'en sean muy similares.

Una de las principales ventajas de los m\'etodos basados en distancias es que el algoritmo se puede ajustar para un problema espec\'ifico por medio de la definici\'on de una distancia adecuada; es decir, la distancia es un par\'ametro del algoritmo de aprendizaje. Igualmente, un algoritmo puede ser usado para cualquier representaci\'on de datos si una funci\'on de distancia se define sobre \'el. Por ejemplo, si se utiliza las distancias para datos proposicionales, en caso de tener datos estructurados, la mayor\'ia de los m\'etodos de aprendizaje se pueden usar directamente utilizando una definici\'on de distancia para el tipo de datos manejado.  Lo anterior implica que el rendimiento del algoritmo puede variar en funci\'on de la distancia utilizada \cite{estruchPhd2009}.

Las funciones de distancia m\'as comunes utilizadas para representaciones basadas en tuplas son la distancia eucl\'idea, la distancia de Manhattan, la distancia de Chebychev, la distancia Minkowski y  la distancia de Mahalanobis. Formalmente, definimos estas distancias de la siguiente manera: 

\noindent Sea $x$ una instancia arbitrariamente descrita por el vector de caracter\'isticas \cite {Mit97}:

\[
\begin{array}{l}
\langle a_{1}(x), a_{2}(x),\ldots, a_{n}(x)\rangle 
\end{array}
\]

\noindent donde $a_{r}(x)$ denota el valor de $r$-\'esimo atributo de la instancia $x$; entonces la distancia entre dos instancias $x_{i}$ y $x_{j}$, es definida por $d(x_{i}, x_{j})$, para cada una de las siguientes distancias:

\begin{itemize}

\item  Distancia eucl\'idea

\[
d(x_i,x_j) = \sqrt{ \sum_{i=1}^{n}  ( a_{r}(x_{i}) - a_{r}(x_{j}))^2} 
\]

\item Distancia de Manhattan

\[
d(x_i,x_j) = \sum_{i=1}^{n} \left | ( a_{r}(x_{i}) - a_{r}(x_{j})) \right |
\]

\item Distancia de Chebychev

\[
d(x_i,x_j) =  max_{i=1, \dots ,n} \left| ( a_{r}(x_{i}) - a_{r}(x_{j})) \right|
\]

\item Distancia de Minkowski

\[
d(x_i,x_j) = \left( \sum_{i=1}^{n} \left| ( a_{r}(x_{i}) - a_{r}(x_{j})) \right|^p \right)^{1/p}
\]

\item Distancia de Mahalanobis

\[
d(x_i,x_j) = \left(   \left( a_{r}(x_{i}) - a_{r}(x_{j})\right)^t S^{-1} \left( a_{r}(x_{i}) - a_{r}(x_{j}) \right)  \right)^{1/2}
\]

\noindent donde $S$ es la matriz de covarianza

\end{itemize}

La distancia de Minkowski es una generalizaci\'on de las diatancias eucl\'idea, Manhattan y Chebychev, donde un par\'ametro $p$ debe ser definido. Si $p=1$, es la distancia de Manhattan, si $p=2$, es la distancia eucl\'idea y finalmente si $p=\infty$, es la distancia de Chebychev. Adicionalmente, la distancia eucl\'idea es un caso particular de la distancia de Mahalanobis: en la distancia eucl\'idea no se tiene en cuenta la correlaci\'on entre los atributos.

Los m\'etodos basados en distancias consideran que cada ejemplo corresponde a un punto en el espacio m\'etrico y el rendimiento de sus predicciones, mediante la comparaci\'on de la proximidad, difieren, en gran parte, en la manera en que son elegidos los ejemplos. A continuaci\'on se hace una breve descripci\'on de los m\'etodos m\'as comunes; el $k$-vecinos m\'as cercanos, el discriminante de Fisher y el aprendizaje por cuantificaci\'on vectorial son m\'etodos supervisados y la agrupaci\'on jer\'arquica y $k$-medias son m\'etodos no supervisados. 

\begin{itemize}
\item \textbf{$k$-vecinos m\'as cercanos:} es uno de los algoritmos m\'as analizados en aprendizaje autom\'atico. Este algoritmo permite aproximar funciones de salida para valores discretos o continuos. Asume que las instancias corresponden a un punto en un espacio m\'etrico $n$-dimensional. El valor de la funci\'on de salida para un nuevo ejemplo es estimado de los valores conocidos de los $k$ ejemplos de entrenamiento m\'as cercanos \cite{Mit97}. Este algoritmo se describe con m\'as detalle en la secci\'on  \ref{KNN}.

\item \textbf{Discriminante de Fisher:} cada clase es representada por medio un centroide que es un punto que minimiza la suma de la distancia para los elementos que pertenecen a una clase; un nuevo ejemplo es marcado de acuerdo con la etiqueta de su centroide m\'as cercano.  Este m\'etodo es {\em ansioso} y los ejemplos son removidos de la memoria una vez que el modelo (conjunto de centroides) haya sido obtenido. Dado que el espacio de representaci\'on est\'a compuesto por tuplas de datos nominales o num\'ericos, se puede derivar un modelo m\'as expresivo. Los centriodes de las clases adyacentes son unidos por l\'ineas rectas y la mediana de estas l\'ineas son las reglas discriminantes \cite{estruchPhd2009}.

\item \textbf{Aprendizaje por cuantificaci\'on vectorial (LVQ):} El modelo aprendido es una colecci\'on de prototipos donde un ejemplo es clasificado de acuerdo con la proximidad a estos prototipos. Inicialmente los prototipos son elegidos al azar, y sus posiciones se actualizan hasta que se alcanza un umbral. Este proceso se realiza de manera iterativa. Primero, un ejemplo del conjunto de entrenamiento es elegido y el prototipo m\'as cercano es determinado. Dependiendo de las etiquetas, tanto del ejemplo como las del prototipo, la posici\'on del prototipo cambia. Si las ambas etiquetas emparejan, el prototipo se acerca m\'as al ejemplo \cite{estruchPhd2009}.

\item  \textbf{Agrupamiento jer\'arquico:} este algoritmo se basa en la construcci\'on de un \'arbol en el que las hojas son los elementos del conjunto de ejemplos, el resto son nodos con subconjuntos de ejemplos que pueden ser utilizados como particionamiento del espacio. Este \'arbol de grupos es tambi\'en llamado dendrograma. Existen dos m\'etodos para construir el \'arbol: 1) {\em Aglomerativo}. El \'arbol se va construyendo empezando por la hojas, hasta llegar a la ra\'iz. Inicialmente, cada ejemplo es un grupo y se van aglomerando los grupos para formar conjuntos m\'as numerosos hasta la ra\'iz, que contiene todos los ejemplos. 2) {\em Desaglomerativo}. Se parte de la ra\'iz, que es un solo grupo conteniendo a todos los ejemplos, y se hacen divisiones paulatinas hasta llegar a las hojas que representa a la situaci\'on en que cada ejemplo es un grupo \cite{hernandez2004introduccion}.    

La forma en que los grupos son divididos o fusionados est\'a guiada por el principio en que los elementos m\'as cercanos deben permanecer en el mismo grupo. En consecuencia, es necesario conocer si dos grupos son cercanos. Teniendo en cuenta que la funci\'on de distancia es definida solo para un par de elementos, es necesario extenderla para medir la distancia entre grupos \cite{estruchPhd2009}\cite{hernandez2004introduccion}. Las m\'as comunes son:
\\
\begin{itemize}
\item \textbf{Enlace simple:} En este m\'etodo, la distancia entre dos grupos es la distancia entre sus miembros m\'as pr\'oximos, es decir, si $U$ y $V$ son dos grupos, entonces:

\[
\begin{array}{l}
d_{UV} = min \left \{   d_{ij} : i \in U, j \in V \right \}
\end{array}
\]

\item \textbf{Enlace completo:} la distancia entre dos grupos es la distancia entre sus miembros m\'as alejados, es decir: 

\[
\begin{array}{l}
d_{UV} = max \left \{   d_{ij} : i \in U, j \in V \right \}
\end{array}
\]

\vspace{0.4cm}

\item \textbf{Enlace promedio:}  la distancia entre dos grupos es la distancia media entre todos los pares de unidades, donde un elemento del par es de un grupo, y el otro elemento pertenece al otro grupo, es decir, si $n_u$ es el n\'umero de unidades en $U$, y $n_v$ es el n\'umero de unidades en $V$ , entonces:

\vspace{0.4cm}

\[
\begin{array}{l}
d_{UV} = \displaystyle\frac{1}{n_un_v} \sum_{i \in U} \sum_{j \in V} d_{ij} 
\end{array}
\]

\vspace{0.4cm}

\end{itemize}

En cuanto a los m\'etodos {\em aglomerativos} , el par de grupos que est\'an m\'as cerca se fusionaran de acuerdo a la distancia de enlace. Si la distancia es el enlace promedio entonces  los centroides deben ser recalculados. El proceso contin\'ua hasta que un n\'umero previamente fijado de los grupos es logrado o el dendrograma completo es construido \cite{estruchPhd2009}\cite{hernandez2004introduccion}.

\item \textbf{$k$-medias:} Los m\'etodos de particionamiento son otro tipo importante de t\'ecnicas de agrupamiento. Los grupos son mejorados gradualmente de acuerdo con una funci\'on de optimizaci\'on. Entre ellos el m\'as conocido es $k$-medias. Inicialmente un conjunto de centroides son elegidos al azar; luego, los ejemplos son asignados a su centroide m\'as cercano formando grupos y el centriode de los grupos nuevos es calculado. Este proceso se repite hasta que los centroides no cambien. Aunque este algoritmo ha tenido \'exito en aplicaciones industriales y cient\'ificas, tiene algunos inconvenientes; el m\'as significativo es que su desempe\~no depende fuertemente de los supuestos iniciales y del n\'umero de centroides propuestos \cite{estruchPhd2009}.
\end{itemize}

Como se ha dicho, el algoritmo de aprendizaje de $k$-vecinos m\'as cercanos, es uno de los algoritmos m\'as analizados en para el aprendizaje autom\'atico. A continuaci\'on se introduce este algoritmo, incluyendo algunas variantes com\'unmente utilizadas para este m\'etodo.

\section{$k$-vecinos m\'as cercanos $k$-NN}\label{KNN} 

El algoritmo $k$-NN, presentado en \cite{Mit97}, asume que todas las instancias corresponden a puntos en un espacio $n$-dimensional $\Re ^{n}$, aunque el algoritmo funciona igualmente para cualquier otro tipo de espacio, incluso sino es m\'etrico. Los vecinos m\'as cercanos de un ejemplo son definidos en t\'erminos de una distancia. Usualmente se utiliza la distancia eucl\'idea; sin embargo, como se mencion\'o anteriormente, por ser un algoritmo basado en distancias, es posible utilizar cualquier otra distancia: la distancia de Manhattan, la distancia de Chebychev, etc.

En el aprendizaje del vecino m\'as cercano, la funci\'on de salida puede ser un valor discreto (clasificaci\'on) o continuo (regresi\'on). Considerando primero el aprendizaje para funciones de salida de valores discretos, la tabla \ref{tab:algoritmo-knn} muestra el algoritmo de $k$-NN para aproximar esta funci\'on definida de la forma $f : \Re^{n}\longrightarrow V$, donde $V$ es un conjunto finito de clases $\{v_1, \ldots, v_s\}$. El valor $\hat{f}(x_q)$, devuelto por el algoritmo como su estimaci\'on de  $f(x_q)$, es el valor m\'as com\'un de $f$ entre los $k$ ejemplos de entrenamiento m\'as cercanos a $x_q$. Si se elige $k=1$, entonces el algoritmo asigna a $\hat{f}(x_q)$ el valor de $f(x_i)$ donde $x_i$ es la instancia de entrenamiento m\'as cercana a $x_q$. Para valores grandes de $k$, el algoritmo asigna el valor m\'as com\'un entre los $k$ ejemplos m\'as cercanos.

\vspace{0.3cm}

\begin{table}[htp]
\begin{center}
\begin{tabular}{|l|}\hline
\\
   Algoritmo de entrenamiento: \\
{\hspace{0.3cm}} - Por cada ejemplo de entrenamiento $\langle x,f(x) \rangle$, se adiciona un ejemplo  a la lista de\\ 
{\hspace{0.5cm}}   ejemplos de entrenamiento.\\ 
\\
Algoritmo de clasificaci\'on: \\
{\hspace{0.3cm}} - Dada una instancia de consulta $x_q $ para ser clasificada, \\
{\hspace{0.6cm}} - Si $x_1, \dots,x_k$ denota las $k$ instancias de ejemplos de entrenamiento que son m\'as \\ 
{\hspace{0.8cm}} cercanas a $x_q$.\\
{\hspace{0.6cm}} - Retorna\\
{\hspace{3cm}}  $\hat{f}(x_q) \longleftarrow argmax_{v \in V }\sum_{i=1}^{k} \delta (v,f(x_i))$ \\
\\
{\hspace{0.8cm}}  donde $\delta(a,b)=1$ si $a=b$ y donde $\delta(a,b)=0$ en otro caso.\\

   \\
\hline
\end{tabular}
\end{center}
\caption{Algoritmo de $k$-NN para aproximar una funci\'on de valores discretos} 
\label{tab:algoritmo-knn}
\end{table}

Para funciones de salida de valores continuos, el algoritmo es f\'acilmente adaptable; el algoritmo calcula el valor medio de los $k$ ejemplos de entrenamiento m\'as cercanos,  en lugar de calcular el valor m\'as com\'un. M\'as precisamente, para aproximar la funci\'on de salida de valores reales $f : \Re^{n}\longrightarrow \Re$ se remplaza $\hat{f}(x_q)$, utilizado en el algoritmo de valores discretos, por: 

\vspace{0.4cm}

\[
\begin{array}{l}
\hat{f}(x_q) \longleftarrow \displaystyle \frac{ \displaystyle \sum_{i=1}^{k} f(x_i)} {\displaystyle k}
\end{array}
\]

\vspace{0.4cm}

Un refinamiento para el algoritmo de $k$-NN puede ser realizado por medio de la funci\'on de {\em atracci\'on}, denominada $w$; este refinamiento consiste en poderar la contribuci\'on de cada uno de los $k$-vecinos de acuerdo con su distancia para el punto de consulta $x_q$, dando mayor peso a los vecinos m\'as cercanos. Por ejemplo, en el algoritmo de la tabla \ref{tab:algoritmo-knn}, el cual aproxima funciones de salida de valores discretos, es posible ponderar el voto de cada vecino de acuerdo con el cuadrado inverso de su distancia de $x_q$. Esto puede ser logrado reemplazando  $\hat{f}(x_q)$ por: 

\[
\begin{array}{l}
\hat{f}(x_q) \longleftarrow argmax_{v \in V }\displaystyle \sum_{i=1}^{k} w_i \delta (v,f(x_i))
\end{array}
\]

\noindent donde,

\vspace{0.4cm}
\[
\begin{array}{l}
\displaystyle w_i \equiv \frac{1}{d(x_q,x_i)^2}
\end{array}
\]

\vspace{0.4cm}

Esta mejora de ponderar los $k$ vecinos m\'as cercanos para un nuevo ejemplo, puede suavizar el impacto de los ejemplos de entrenamiento aislados o con ruido. Adicionalmente, es posible considerar que el exponente cuadr\'atico, utilizado en el denominador de $w$,  puede ser modificado por una variable denominada {\em par\'ametro de atracci\'on} que permite incrementar o decrementar la ponderaci\'on $w_i$ de manera que, mientras mayor sea esta variable, menos importante es $k$. 

De igual manera, para funciones de salida con valores reales, la distancia ponderada de instancias utiliza un denominador constante que normaliza las contribuciones de varias ponderaciones. De este modo, la funci\'on $\hat{f}(x_q)$ puede ser remplazada  por:

\[
\begin{array}{l}
\displaystyle \hat{f}(x_q) \longleftarrow \frac{\displaystyle \sum_{i=1}^{k} w_i f(x_i)}{ \displaystyle \sum_{i=1}^{k} w_i}
\end{array}
\]

\vspace{0.4cm}

El algoritmo $k$-vecinos m\'as cercanos es uno de los algoritmos m\'as analizados en el aprendizaje autom\'atico \cite{Mit97}, esto se debe a su simplicidad y, por otra parte, a la edad que tiene. Es un m\'etodo inductivo de inferencia muy eficaz en muchos problemas pr\'acticos, especialmente cuando utiliza mejoras como la distancia ponderada;  es robusto ante los ruidos de datos y suficientemente efectivo en conjuntos de datos grandes.

\vspace{0.4cm}

\section{Distancias para datos estructurados y semi"=estructurados}

En el aprendizaje autom\'atico, los m\'etodos basados en distancias pueden incorporar funciones de similitud definidas sobre datos estructurados (por ejemplo una tabla), semi-estructurados (por ejemplo un documento XML) o no estructurados (por ejemplo un documento de texto). Algunos m\'etodos basados en similitud de datos estructurados son aplicados, en gran medida, a semi-estructurados; sin embargo, estos datos que no son completamente estructurados, es necesario aplicar t\'ecnicas que requieren procesos m\'as complejos. Las distancias entre grafos (o en particular, distancias entre \'arboles) son com\'unmente utilizadas como una medida de similitud para datos semi-estructurados. A continuaci\'on se describen algunos aspectos generales sobre datos estructurados y semi-estructurados y, finalmente, se detallan m\'etodos basados en similitud de acuerdo con varios tipos de datos.

\subsection{Datos estructurados y semi-estructurados}

\vspace{0.4cm}

La principal diferencia que existe entre los datos estructurados y semi-estructurados est\'a relacionada, obviamente, con la rigidez de su estructura; para los datos estructurados, un conjunto de instancias son definidas por variables con tipos de datos iguales y \'unicamente cambia el valor de esta variable (por ejemplo una tupla de tama\~no fijo de valores reales, una imagen de $n\times m$ p\'ixeles, etc.); los datos semi-estructurados estan conformados fundamentalmente por etiquetas que, aunque ofrecen cierto nivel estructural, tambi\'en son altamente flexibles, con diferentes tipos de datos y dominios, adem\'as de datos no estructurados (como textos, imagenes, etc.)  que adicionan dificultad a tareas de consulta y por lo tanto a la aplicaci\'on de m\'edotos de aprendizaje (por ejemplo un documento en XML, un t\'ermino l\'ogico-funcional, una secuencia de ADN, etc.).

La definici\'on de funciones de similitud entre objetos semi-estructurados se basa, en gran medida, en su parte estructural y la utilizaci\'on de aquellas etiquetas existentes para un objeto; no obstante, existen t\'ecnicas para tratar los segmentos no estructurados de acuerdo con su contenido; por ejemplo, el procesamiento de lenguaje natural ofrece t\'ecnicas que permiten medir la similitud ling\"u\'istica de objetos por medio de an\'alisis l\'exico, sint\'actico y sem\'antico. 

En un documento semi-estructurado, en aquellas partes que contienen alg\'un tipo de estructura, se pueden aplicar funciones de similitud existentes para datos estructurados; aunque es posible encontrar diferentes dominios sobre variables conceptualmente iguales que obliga, en algunas ocasiones, a utilizar una serie de transformaciones y aplicar t\'ecnicas que permitan crear dominios comparables.

En los datos semi-estructurados, por sus caracter\'isticas de etiquetado, es usual que se creen estructuras jer\'arquicas. Esto ha hecho que muchos casos hayan sido abordados mediante representaciones basadas en grafos o \'arboles; en consecuencia, es com\'un que las funciones de distancia empleadas para medir la similitud entre instancias se basen tambi\'en en distancias para este mismo tipo de datos. La similitud sobre datos semi-estructurados, han sido utilizadas principalmente para agrupaci\'on de documentos, y para detectar cambios entre estos. Algunos enfoques pueden ser vistos en \cite{XXG07}, \cite{FGOT03}, \cite{WDC03}.

\subsection{Distancias entre datos}

A continuaci\'on se decriben algunos m\'etodos basados en similitud que son empleados acuerdo con los tipos datos m\'as comunes (prensentados por \cite{estruchPhd2009}, \cite{EFHR06d} y \cite{BS97}): conjuntos, listas, \'arboles y grafos. Las distancias entre t\'erminos son un tema central en este trabajo, por esta raz\'on son tratadas en la secci\'on \ref{distTerminos}.  
 
\begin{itemize}

\item\textbf{Conjuntos}:\\ \\
La cardinalidad de la diferencia sim\'etrica entre dos conjuntos finitos $A$ y $B$ es una funci\'on de distancia  $\mid(A-B)\cup(B-A)\mid$ que satisface la propiedad de identidad y simetr\'ia. La desigualdad triangular puede verse demostrando que para cualquier conjunto finito $C$, si el elemento $x$ est\'a en $\mid(A-B)\cup(B-A)\mid$ entonces $x$ est\'a en $\mid(A-C)\cup(C-A)\mid$ o en $\mid(C-B)\cup(B-C)\mid$, lo que implica que $\mid(A-B) \cup (B-A)\mid \leq \mid(A-C) \cup (C-A)\mid+\mid(C-B)\cup(B-C)\mid$. Cuando esta m\'etrica es empleada, la distancia entre dos conjuntos es dada por el n\'umero de elementos que tienen en com\'un. Los elementos m\'as comunes y los elementos m\'as cercanos. Por ejemplo, dado el siguiente conjunto de secuencias, $A = \left\{ab, a^4 \right\}$, $B = \left\{ab, d^4 \right\}$ y $C = \left\{ab, a^3 \right\}$ entonces $d(A,B) = d(A,C) = 2$. Esta funci\'on asume que la distancia $0$ \'o $1$ ($d(x, y) = 1$ si $x\not=y$, $0$ otro caso) ha sido previamente definida sobre los elementos construidos en el conjunto; por lo tanto, la diferencia sim\'etrica es una distancia que considera que cualquier elemento del conjunto es igualmente diferente al resto. Por ejemplo, esta distancia muestra que el elemento $d^4 \in B$ y $a^3 \in C$ son diferentes a $a^4 \in A$, cuando intuitivamente $a^3$ es m\'as similar a $a^4$ que $d^4$.

En algunos contextos es importante capturar las diferencias entre los elementos de los conjuntos de una manera m\'as precisa a como lo hace la diferencia sim\'etrica. La distancia de {\em Hausdorff}, por ejemplo, es una distancia que tiene en cuenta los elementos; y  esta distancia est\'a definida como:

\vspace{0.6cm}

\[
d_H(A,B) =\textrm{m\'ax} \left\{
\begin{array} {llll}
\textrm{sup inf} & d(a,b), & \textrm{sup inf} & d(a,b)\\
^{a \in A \; b \in B} & & ^{ b \in B \; a \in A} &
\end{array}
\right \}
\]

\vspace{0.6cm}

\noindent donde $A$ y $B$ son dos conjuntos y $d(\cdot,\cdot)$ es una distancia definida sobre elementos. 
\\
\item\textbf{Listas}:\\ \\
Una de las funciones para listas es la distancia de {\em Hamming}; Esta funci\'on es usada cuando las listas tienen longitud igual; la distancia entre dos listas es el n\'umero de posiciones para las cuales los s\'imbolos son diferentes. Por ejemplo, dada la secuencia $s_1=aabb$ y $s_2=accb$ entonces la distancia $d(s_1,s_2)= 2$. 
La distancia de edici\'on (tambien conocida como {\em Levenshtein}), es una generalizaci\'on de la distancia de {\em Hamming}, que permite manejar secuencias con longitudes variables. Esta distancia cuenta el n\'umero m\'inimo de operaciones de eliminaci\'on, inserci\'on y sustituci\'on requeridas para transformar una secuencia en otra. Estas operaciones pueden ser ponderadas de acuerdo con el contexto del problema. Por lo tanto, la distancia corresponde a la transformaci\'on con el costo m\'as bajo. A continuaci\'on se describe un ejemplo para esta distancia:
\vspace{0.2cm}

Sea alfabeto $\Sigma  = \{ a,b,c \}$ y $\Sigma^\ast$ un espacio finito de todas las listas construidas sobre los s\'imbolos $a$,$b$ y $c$.  Se asigna un costo de $1$ para inserciones y eliminaciones y, $2$ para las sustituciones. La distancia de edici\'on entre dos secuencias $s_1=aabb$ y $s_2=bbcc$ puede ser especificada por:

\vspace{0.4cm}

\[
\begin{array} {cccccc}
a & a & b& b & &\\
&&b&b&c&c
\end{array}
\]

\vspace{0.5cm}

\noindent Entonces $s_1$ es transformada en $s_2$ borrando los dos primeros s\'imbolos y luego adicionando dos s\'imbolos al final:

\vspace{0.4cm}

\[
s_1 = aabb \rightarrow  abb \rightarrow  bb \rightarrow  bbc \rightarrow  bbcc = s_2
\]

\vspace{0.5cm}

\noindent Lo anterior hace que $d(s_1, s_2) = 4$, esto es equivalente a transformar $s_2$ en $s_1$   borrando los dos s\'imbolos $b$ y luego adicionando los dos s\'imbolos $a$. Las otras transformaciones pueden no ser \'optimas, por ejemplo:

\vspace{0.5cm}

\[
\begin{array} {ccccc}
a & a & b& b & \\
&b&b&c&c
\end{array}
\]

\vspace{0.5cm}

\noindent La secuencia $s_1$ es transformada a $s_2$ removiendo el primer s\'imbolo $a$ (con costo $1$); se sustituye el segundo s\'imbolo $a$ por $b$ (con costo $2$), luego se cambia el tercer s\'imbolo $b$ por $c$ (con costo $2$), finalmente se adiciona $c$ al final de la secuencia (con costo $1$)

\vspace{0.6cm}

\[
s_1 = aabb \rightarrow  abb \rightarrow  bbb \rightarrow  bbc \rightarrow  bbcc = s_2
\]

\vspace{0.5cm}

El costo total de esta transformaci\'on es $6$; que es mayor a la primera transformaci\'on que ten\'ia un costo de $4$; por esa raz\'on, esta \'ultima no es una trasformaci\'on \'optima.
\\
\item\textbf{\'Arboles y grafos}:\\ \\
La distancia de edici\'on tambi\'en puede ser utilizada para grafos y \'arboles etiquetados. Formalmente, un grafo puede ser definido $g(V, \alpha, \beta )$ donde $V$ es un conjunto finito de v\'ertices y $\alpha: V \rightarrow L$ es un nodo  y $\beta: V \times  V \leftarrow L$ es una funci\'on de etiquetado de aristas. En este contexto, los grafos son siempre completos si las aristas faltantes son consideradas como una etiqueta especial vac\'ia. 

\vspace{0.3cm}

Frecuentemente, se utiliza el concepto de {\em ecgm} ({\em error-correcting graph matching}) el cual es un conjunto de operaciones de edici\'on para transformar un grafo en otro y el costo asociado a un {\em ecgm}.

\vspace{0.3cm}

Formalmente, sean $g_1 = (V_1, \alpha_1, \beta_1)$ y $g_2 = (V_2, \alpha_2, \beta_2)$ dos grafos. El {\em ecgm} de $g_1$ a $g_2$ es una funci\'on biyectiva   $f : \hat{V}_1 \rightarrow \hat{V}_2$ donde $\hat{V}_1 \subset V_1$ y $\hat{V}_2 \subset V_2$.

\vspace{0.3cm}

El costo de un  {\em ecgm}  $f : \hat{V}_1 \rightarrow \hat{V}_2$ de un grafo  $g_1 = (V_1, \alpha_1, \beta_1)$ y$g_2 = (V_2, \alpha_2, \beta_2)$ es dado por:

\vspace{0.5cm}

\[
\begin{array} {lll}
c(f) & = & \sum_{v \in \hat{V}_1}c_{ns}(v) + \sum_{v \in V_1 - \hat{V}_1}c_{nd}(v) + \sum_{v \in V_2 - \hat{V}_2}c_{ni}(v)  \\ \\
& & + \sum_{e \in \hat{E}_1}c_{es}(e)  +  \sum_{e \in E_1 - \hat{E}_1}c_{ed}(e) +  \sum_{e \in E_2 - \hat{E}_2}c_{ei}(e)
\end{array}
\]

\vspace{1cm}

donde $c_{ns}$, $c_{nd}$ y $c_{ni}$ son el costo de sustituci\'on, eliminaci\'on e inserci\'on de un nodo, respectivamente; e igualmente, $c_{es}$, $c_{ed}$ y $c_{ei}$ son el costo de sustituci\'on, eliminaci\'on e inserci\'on de una arista.

\vspace{0.2cm}

Finalmente, la m\'inima distancia entre dos grafos $g_1$ y $g_2$ es el m\'inimo costo obtenido sobre todos los {\em ecgm} de $g_1$ y $g_2$. Por ejemplo:

\vspace{0.2cm}

Sean $g_1 = (V_1, \alpha_1, \beta_1)$ y $g_2 = (V_2, \alpha_2, \beta_2)$ dos grafos representados en la figura \ref{fig:distancia_grafos}:

\vspace{0.5cm}

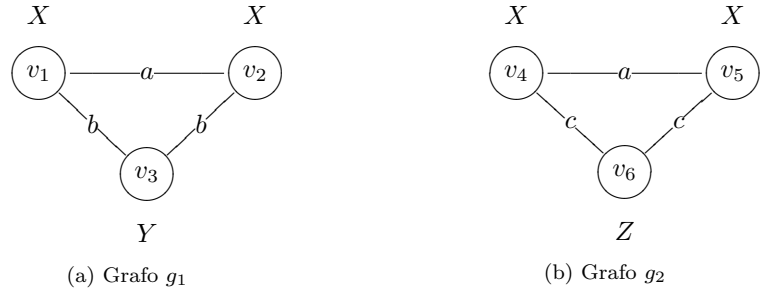
\begin{figure*}[h]
  \centering
   \subfloat[Grafo $g_1$]{\label{fig:grafo1}
     \xymatrix @R=0.2cm @C=0.2cm {
      &X&&&&X\\
      & *+<0.5cm>[o][F]{v_1} \ar@{-}[dr]  \ar@{-}[r] & *{} \ar@{-}[r] & *{a} \ar@{-}[r] &  *{} \ar@{-}[r] &*+<0.5cm>[o][F]{v_2} \\ 
      &&  *{b} \ar@{-}[dr] & & *{b} \ar@{-}[ur]& \\
      && &*+<0.5cm>[o][F]{v_3} \ar@{-}[ur] & & \\
      &&&Y&&\\
    } }
    \hspace{2cm}
   \subfloat[Grafo $g_2$]{\label{fig:grafo2}
     \xymatrix @R=0.2cm @C=0.2cm {
     &X&&&&X\\
     & *+<0.5cm>[o][F]{v_4} \ar@{-}[dr]  \ar@{-}[r] & *{} \ar@{-}[r] & *{a} \ar@{-}[r] &  *{} \ar@{-}[r] &*+<0.5cm>[o][F]{v_5} \\ 
     &&  *{c} \ar@{-}[dr] & & *{c} \ar@{-}[ur]& \\
     && &*+<0.5cm>[o][F]{v_6} \ar@{-}[ur] & & \\
     &&&Z&&\\
   } }
   \caption{Ejemplo de distancia entre dos grafos}
   \label{fig:distancia_grafos}
 \end{figure*}

\vspace{0.6cm}

\begin{itemize}
\item $V_1 = {1,2,3}$; $V_2 = \{4,5,6\}$; $L = \{X,Y,Z,a,b,c, null\}$.
\item $\alpha_1: 1 \mapsto X, 2 \mapsto X, 3 \mapsto Y$
\item $\alpha_2: 4 \mapsto X, 5 \mapsto X, 6 \mapsto Z$
\item $\beta_1: (1,2) \mapsto a, (1,3) \mapsto b, (2,3) \mapsto b$
\item $\beta_2: (4,5) \mapsto a, (4,6) \mapsto a, (5,6) \mapsto c$

\end {itemize}

\vspace{0.6cm}

Un posible {\em ecgm} es $f:1 \mapsto 4, 2 \mapsto 5$ con $\hat{V}_1 = \{1,2\}$ y $\hat{V}_2 = \{4,5\}$. Aplicando el {\em ecgm} los nodos $1$ y $2$ son sustituidos por $4$ y $5$, respectivamente. En consecuencia, la arista $(1,2)$ es sustituida por $(4,5)$. Todas estas sustituciones son id\'enticas en el sentido en que no hay cambios en la etiquetas; El nodo $3$ y las aristas $1,3$ y $2,3$ son eliminadas y el nodo $6$ junto con sus aristas $(4,6)$ y $(5,6)$ son insertados. 
\vspace{0.2cm}

Si el costo de la funci\'on es definido de la siguiente manera:
\vspace{0.5cm}

\[c_{ns}(v) =\left\{
\begin{array} {ll}
0, & \mbox{si $\alpha_1(v) = \alpha_2(v)$}\\
\infty, & \mbox{otro caso}\\
\end{array}
\right.\]
\[
c_{nd}(v) = 1, \forall v \in V_1 - \hat{V}_1
\]
\vspace{0.5cm}
\[
c_{ni}(v) = 1, \forall v \in V_2 - \hat{V}_2
\]
\vspace{0.5cm}
\[
c_{es}(e) =\left\{
\begin{array} {ll}
0, & \mbox{si $\beta_1(v) = \beta_2(v)$}\\
\infty, & \mbox{otro caso}\\
\end{array}
\right.
\]
\vspace{0.5cm}
\[
c_{ed}(v) = 0, \forall e \in E_1 - \hat{E}_1
\]
\vspace{0.5cm}
\[
c_{ci}(v) = 0, \forall e \in E_2 - \hat{E}_2
\]

\vspace{0.8cm}

Se puede ver f\'acilmente que $d(g_1, g_2) = c(f) = 2$, puesto que el costo de eliminaci\'on o inserci\'on de un nodo es igual a $1$ (mientras que el costo de inserci\'on o eliminacion de una arista es $0$).
\end{itemize}

\vspace{0.5cm}

\section{Representaci\'on basada en t\'erminos y XML}\label {Representacion-Basada-Terminos}

Adem\'as de la popularidad alcanzada por XML para la representaci\'on y el intercambio de datos en la Web \cite{goldman1999semistructured}, el principal inter\'es de utilizar XML en este trabajo es la capacidad de representar estructuras como listas y \'arboles; por lo anterior, XML permite representar objetos basados en t\'erminos. A continuaci\'on se hace una corta descripci\'on de este tipo de representaciones.

\subsection{Representaci\'on basada en t\'erminos}

La representaci\'on basada en t\'erminos busca obtener representaciones compactas donde los objetos est\'an representados por medio de funtores los cuales son s\'imbolos de funci\'on no-evaluadas cuyos argumentos son t\'erminos tan complejos como un problema lo requiera. En consecuencia, los funtores pueden ser usados para representar tama\~nos flexibles de tipos de datos ordenados tales como listas y \'arboles. Esto tiene la ventaja de que todos los datos relativos a un objeto se mantienen unidos \cite{estruchPhd2009}\cite{lachiche2000first}. El siguiente ejemplo muestra una representaci\'on basada en t\'erminos de un objeto del conjunto de datos de {\em Mushroom} (UCI {\em machine learning repository} \cite{UCI}):

\noindent
{
\ttfamily
\footnotesize
\indent Mushroom(\\
\indent \indent cap(CONVEX, SMOOTH, WHITE), \\ 
\indent \indent BRUISES, ALMOND, \\
\indent \indent gill(FREE, CROWDED, NARROW, WHITE), \\
\indent \indent stalk(TAPERING, BULBOUS, surface(SMOOTH, SMOOTH), color(WHITE, WHITE)), \\ 
\indent \indent veil(PARTIAL, WHITE), \\
\indent \indent ring(ONE, PENDANT), \\
\indent \indent spore(print(BROWN)), \\
\indent \indent SEVERAL, WOODS) \\
}

\begin{figure*}[h]
\centering
\includegraphics[width=0.73\textwidth]{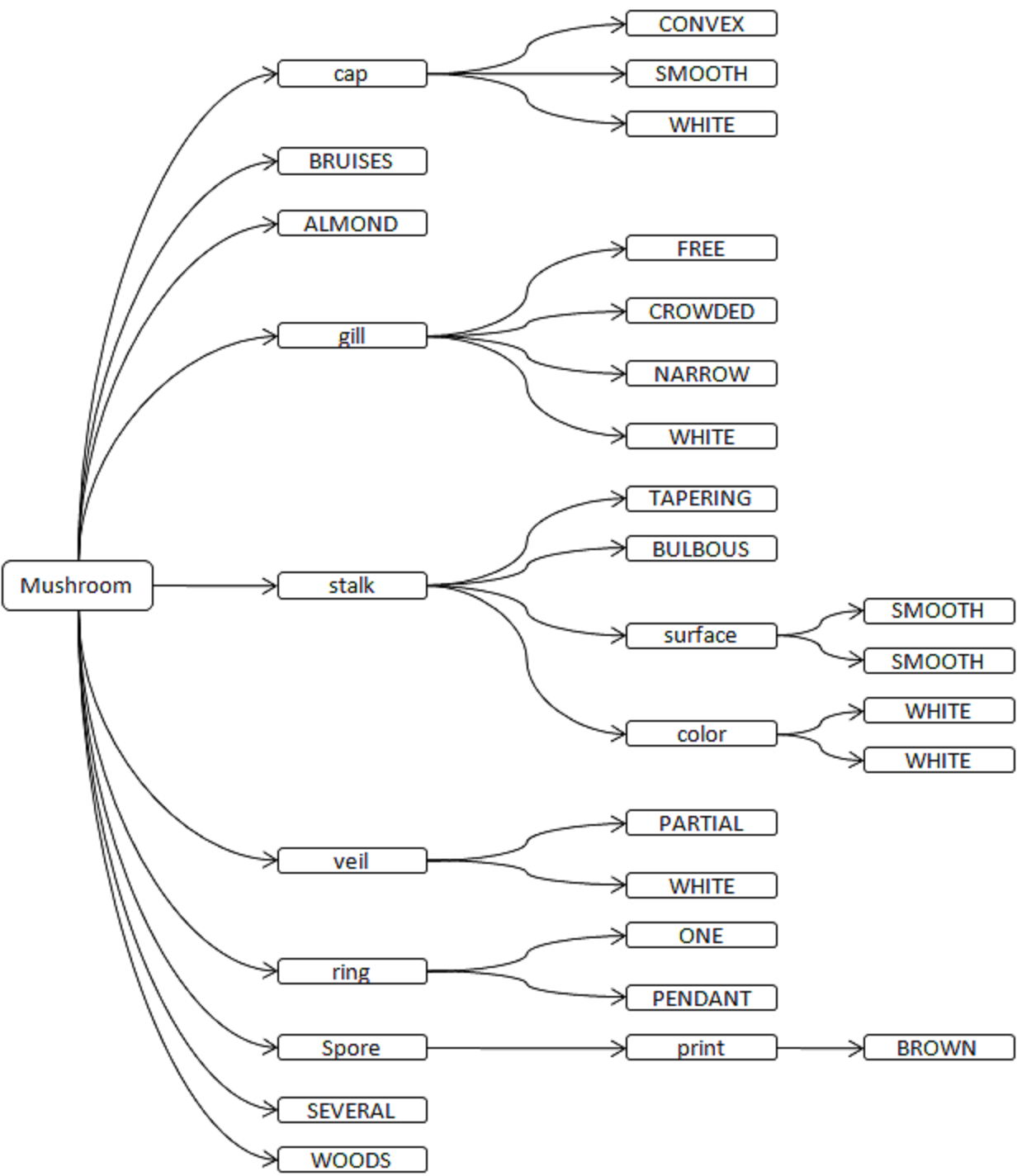}
\vspace{-0.2cm}
\caption{Representaci\'on en \'arbol de un objeto de {\em Mushroom}.}
\label{fig:jeraquiaMushrrom}
\end{figure*}

El s\'imbolo de predicado solo se refiere al objeto, mientras los funtores se refieren a partes de los objetos y las constantes se refieren a propiedades de estas partes. Por lo tanto, {\em Mushroom} es una tupla compuesta por {\em cap}, {\em gill}, {\em stalk}, {\em veil}, {\em ring} y {\em spore}, y varios atributos adicionales. A su vez, cada tupla puede estar compuesta de manera anidada por otras tuplas y atributos. La figura \ref{fig:jeraquiaMushrrom} muestra una representaci\'on en \'arbol del ejemplo anterior.

Una de las principales caracter\'isticas de la representaci\'on basada en t\'erminos es que puede ser aplicada de una manera m\'as natural que otros paradigmas que utilizan ejemplos con estructuras planas como datos tabulares, textos, etc. Las estructuras planas pueden obligar a tener campos en blanco (por ausencia de caracter\'isticas) y no representar claramente relaciones jer\'arquicas entre variables.

\subsection{XML}

XML es un lenguaje textual que ha ganado popularidad para la representaci\'on y el intercambio de datos en la Web. Los componentes b\'asicos son elementos etiquetados que tienen una secuencia de cero o m\'as pares atributo-valor y una secuencia de cero o m\'as subelementos. Los subelementos pueden ser ellos mismos los elementos marcados o pueden ser segmentos de datos de texto sin etiqueta. Debido a que XML se define como un lenguaje textual y no como un modelo de datos, un documento XML siempre tiene un orden impl\'icito (el orden puede o no ser relevante, sin embargo, es inevitable en una representaci\'on textual) \cite{goldman1999semistructured}.

XML permite jerarquizar y estructurar la informaci\'on, describir los contenidos dentro del mismo documento y reutilizar partes del mismo; por esta raz\'on un documento XML permite la representaci\'on de jerarqu\'ias y estructuras complejas que, al igual que la representaci\'on basada en t\'erminos, es m\'as natural que estructuras planas; sin embargo, XML no es un t\'ermino y deben considerarse aspectos como el orden de los elementos y la ausencia de caracter\'isticas. 

\begin{figure*}[h]
\centering
\includegraphics[width=0.43\textwidth]{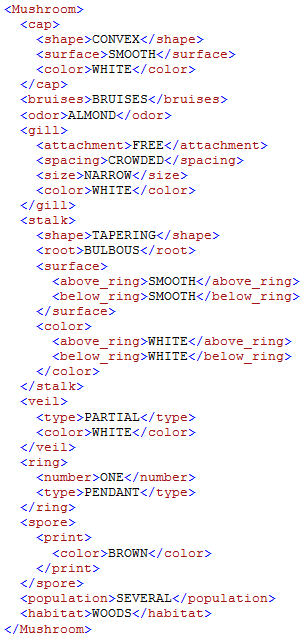}
\vspace{-0.2cm}
\caption{Representaci\'on en XML de un objeto basado en t\'erminos.}
\label{fig:MushroomXML}
\end{figure*}

Un documento XML puede ser convertido en un \'arbol de t\'erminos funcionales usando listas y este puede ser tratado en el marco de programaci\'on l\'ogico-funcional \cite{hernandez2004introduccion}. La figura \ref{fig:MushroomXML} muestra una representacion en XML del objeto del ejemplo anterior.

\section{Distancias entre t\'erminos} \label{distTerminos} 

Las distancias entre t\'erminos son distancias muy adecuadas para la inferencia inductiva porque incorporan propiedades adicionales sobre las distancias tradicionales. Las distancias entre t\'erminos m\'as relevantes son la distancia propuesta por Nienhuys-Cheng \cite{NW97} y la distancia de J. Ramon et al. \cite{ramon98b}. Cada una de estas dos distancias cuenta con propiedades espec\'ificas que las hacen muy diferentes entre s\'i. Por otra parte, V. Estruch \cite{estruch2010integrated} propone una distancia entre \'atomos/t\'erminos que busca integrar los beneficios de las distancias anteriores. 

A continuaci\'on se describen las propiedades generales de las distancias entre t\'erminos y a qu\'e distancias est\'an asociadas estas propiedades; posteriormente, se presenta la notaci\'on para definir las distancias de Nienhuys-Cheng, J. Ramon et al. y Estruch et al.; finalmente se definen estas distancias (la distancia de Nienhuys-Cheng y Estruch et al. son definidas en las secciones \ref{distancia-Nienhuys-Cheng} y \ref{distancia-Estruch et al}, respectivamente). 

\subsection{Propiedades generales para las distancias entre t\'erminos}\label{propiedades-distancias-entre-terminos}

\noindent Las propiedades generales de las distancias entre t\'erminos (definidas en \cite{estruch2010integrated}) son:

\begin{itemize}

\item\textbf{Sensibilidad del contexto:} Es la posibilidad de considerar d\'onde ocurren las diferencias entre dos t\'erminos, de manera que las diferencias de las ocurrencias de s\'imbolos que ocurren en posiciones m\'as profundas cuentan menos, ya a que estas ofrecen menos informaci\'on. Por ejemplo, la distancia entre $p(a)$ y $p(b)$ debe ser mayor que la distancia entre $p(f(a))$ y $p(f(b))$. 

\item\textbf{F\'acil normalizaci\'on:} es muy \'util trabajar con distancias que puedan ser normalizadas. Por ejemplo aquellas que devuelven un valor real pueden ser f\'acilmente normalizadas. 

\item\textbf{Diferencias repetidas:} consiste en manejar apropiadamente las diferencias repetidas entre los t\'erminos. Por ejemplo si $r=p(a,a)$,  $s=p(b,b)$ y $t=p(c,d)$, intuitivamente se espera que los t\'erminos $r$ y $s$ sean m\'as cercanos que $r$ y $t$ (\'o $s$ y $t$). 

\item\textbf{Tama\~no de las diferencias:} esta propiedad consiste en que si el tama\~no de los t\'erminos se incrementa, entonces la distancia debe ser mayor. Por ejemplo, si la distancia $d(p(a),p(b))= 1/2$ entonces la distancia $d(p(a),p(f(c)))$ debe ser mayor a $1/2$.

\item\textbf{Manejo de las variables:} las variables son muy \'utiles cuando falta parte de la estructura de un objeto, ya que en caso de no existir, se requiere de conceptos extras para manejar s\'imbolos de variables. 

\item\textbf{Componibilidad:} permite definir las funciones de distancia para tuplas, combinando las funciones de distancia definidas sobre tipos b\'asicos sobre los cuales se construye esta tupla. Com\'unmente, esta combinaci\'on es hecha como una combinaci\'on de distancias base.  

\item\textbf{Ponderaciones:} consiste en dar mayor o menor ponderaci\'on para posiciones espec\'ificas de un t\'ermino sobre otros. Por ejemplo, la distancia entre $f(a)$ y $f(b)$ puede incrementarse si estos t\'erminos son reescritos como $f(d_1(d_2(a)))$ y  $f(d_1(d_2(b)))$.

\end{itemize}

La tabla \ref{tab:Ventajas-distancias_terminos} muestra las propiedades para las distancias de Nienhuys-Cheng, J. Ramon et al. y Estruch et al. La distancia de Nienhuys-Cheng solo tiene en cuenta la componibilidad, la normalizaci\'on y, aunque no siempre, el contexto en el que las diferencias se producen. Por el contrario, la distancia de J. Ramon et al. a pesar de cumplir con muchas de las propiedades, adolece de no cumplir f\'acilmente la propiedad de normalizaci\'on y componibilidad y, al igual que la distancia de Nienhuys-Cheng, no siempre cumple la sensibilidad de contexto. Obviamente, la distancia de Estruch et al., por tratarse de un distancia que integra las propiedades de las dos distancias anteriores cumple, directa o indirectamente, todas las propiedades.

\vspace{0.6cm}
\begin{table}[htp]
\begin{center}
\begin{tabular}{|l|c|c|c|}\hline
& Nienhuys-Cheng & J. Ramon et al. & Estruch et al.
\\\hline\hline
{\em Contexto} & No siempre & No siempre & S\'i \\
& & (dependiendo de la ponderaci\'on usada) & \\
\hline
{\em Normalizaci\'on} & S\'i & No es f\'acil & S\'i \\
\hline
{\em Repeticiones} & No & S\'i & S\'i \\
\hline
{\em Tama\~no} & No & S\'i & S\'i \\
\hline
{\em Variables} & Indirectamente & S\'i & Indirectamente \\
\hline
{\em Componibilidad} & S\'i & Dif\'icil & S\'i \\
\hline
{\em Ponderaciones} & No & S\'i & Indirectamente \\
\hline
\end{tabular}
\end{center}
\caption{Propiedades de las distancias entre t\'erminos \cite{estruch2010integrated}.
\label{tab:Ventajas-distancias_terminos}}
\end{table}

\subsection{Notaci\'on \cite{estruch2010integrated}}

Sea $\cal{L}$ un lenguaje de primer orden definido sobre $\Sigma= \langle {\cal C}, {\cal F},\Pi\rangle$, donde $\C$ es un conjunto de constantes, y $\F$ ($\Pi$, respectivamente) es una familia indexada en $\mathbb{N}$ (enteros no negativos), donde ${\F}_n$ ($\Pi_n$) es un conjunto de la funci\'on $n-$aria de s\'imbolos de predicado. Los \'atomos y t\'erminos son construidos, como es usual, de $\Sigma$. El s\'imbolo ra\'iz y la aridad de una expresi\'on $t$ est\'a dado por las funciones $Root(t)$ y $Arity(t)$, respectivamente. Por lo tanto, siendo $t=p(a,f(b))$, $Root(t)=p$ y $Arity(t)=2$.

Considerando la representaci\'on habitual de $t$ como un \'arbol etiquetado, las ocurrencias son secuencias finitas de n\'umeros positivos (separados por puntos) que representan una ruta de acceso en $t$. Se supone que cada ocurrencia siempre est\'a encabezada por un s\'imbolo especial (impl\'icito) $\lambda$, el cual denota la ocurrencia vac\'ia. El conjunto de todas las ocurrencias de $t$ es denotado por $O(t)$. Para este caso, $O(t)=\{\lambda,1,2,2.1\}$. Se utilizan letras min\'usculas (indexadas) $o',o,o_1,o_2,\ldots$ para representar las ocurrencias. La longitud de una ocurrencia $o$, $Length(o)$, es el n\'umero de elementos en $o$ (excluyendo $\lambda$). Por ejemplo, $Length(2.1)=2$, $Length(2)=1$ y $Length(\lambda)=0$. Adicionalmente, si $o \in O(t)$ entonces $t|_o$ representa el subt\'ermino de $t$ en la ocurrencia $o$. En el ejemplo anterior, donde $t=p(a,f(b))$, $Root(t)=p$, $t|_{1}=a$, $t|_{2}=f(b)$, $t|_{2.1}=b$. En cualquier caso, siempre se tiene que $t|_{\lambda}=t$. Para $Pre(o)$, se denota el conjunto de todas los prefijos de las ocurrencias de $o$ diferentes de $o$. Por ejemplo, $Pre(2.1)=\{\lambda,2\}$, $Pre(2)=\{\lambda\}$ and $Pre(\lambda)=\emptyset$. Dos expresiones $s$ y $t$ son compatibles (denotado por la funci\'on booleana $Compatible(s,t)$) si $Root(s)=Root(t)$ y $Arity(s)=Arity(t)$. De lo contrario, se dice que $s$ y $t$ son incompatibles ($\neg Compatible(s,t)$).

\subsection{Definici\'on de las distancias entre t\'erminos}

A continuaci\'on se define la distancia de J. Ramon et al. Las distancias de Nienhuys-Cheng y Estruch et al. son definidas en el secci\'on \ref{distancia-Nienhuys-Cheng} y \ref{distancia-Estruch et al}.

\noindent \textbf{Distancia de J. Ramon et al.}

La distancia J. Ramon et al. \cite{estruch2010integrated}, se basa en las diferencias sint\'acticas con relaci\'on a su operador {\em lgg} ({\em least general generalisation} \cite{plotkin1981structural}); adicionalmente utiliza una funci\'on de tama\~no para calcular esta distancia. El tama\~no es definido como $Size(t)=(F,V)$, donde $F$ cuenta el n\'umero de predicados y s\'imbolos de funci\'on que ocurren en $t$; $V$ es la suma de la frecuencia al cuadrado de la aparici\'on de cada variable en $t$. 

Dados dos t\'erminos $s$ y $t$ la distancia de J. Ramon et al., denotada por $d_R$, es definida de la siguiente manera: 

\[
d_R(s,t) = [Size(s) - Size(lgg(s,t))] + [Size(t) - Size(lgg(s, t))]
\]

Esta distancia devuelve un par de valores $(F,V)$ que expresa qu\'e tan diferentes son los t\'erminos de funci\'on y los s\'imbolos de variables. Por ejemplo, si $s = p(a,b)$ y $t= p(c,d)$ y se conoce que el $lgg(s,t)=p(X,Y)$, entonces:

\[
\begin{array}{lll}
Size(s)=(3, 0) \\
Size(t)=(3, 0) \\
Size(lgg(s,t))=(1,2) \\
d_R(s,t)=[(3,0)-(1,2)]+[(3,0)-(1,2)]=(2,-2)+(2,-2)=(4,-4) \\ 
\end{array}
\]

Debido a que su funci\'on de salida no devuelve un valor num\'erico, sino un par de valores, es dif\'icil la aplicaci\'on directa de esta distancia sobre algoritmos de clasificaci\'on tradicionales basados en distancias.

\section{Distancia de Nienhuys-Cheng \cite{NW97}}\label{distancia-Nienhuys-Cheng}

La distancia de Nienhuys-Cheng, como se mencion\'o anteriormente en la secci\'on \ref{propiedades-distancias-entre-terminos}, tiene en cuenta la profundidad de las ocurrencias de s\'imbolos, de manera que aquellas diferencias que ocurren m\'as cerca del elemento ra\'iz cuenta m\'as. 

Dadas dos expresiones, $s = s_0(s_1, \ldots, s_n)$ y $t =t_0(t_1, \ldots, t_n)$, la distancia de Nienhuys-Cheng, denotada por $d_N(s,t)$, es definida recursivamente como:

\[d_N(s, t) =\left\{
\begin{array} {ll}
0, & \mbox{si $s = t$}\\
1, & \mbox{si $\neg Compatible(s,t)$}\\
\frac{1}{2n}\sum_{i=1}^{n} d(s_i, t_i), & \mbox{otro caso}
\end{array}
\right.\]

Por ejemplo, sean $s=p(a,a)$ y $t=p(f(b),f(b))$ dos expresiones; entonces, 

\[
d_N(s,t) = \frac{1}{4}\cdot \bigl(d(a,f(b))+d(a,f(b))\bigr) = \frac{1}{4}\bigl(1+1\bigr) = \frac{1}{2} 
\]

Debe destacarse que la funci\'on de salida de esta distancia devuelve un valor n\'umerico que puede ser facilmente utilizado por algoritmos tradicionales de aprendizaje basados en distancias.

\section{Distancia de Estruch et al. \cite{estruch2010integrated}}\label{distancia-Estruch et al}

La distancia entre \'atomos/t\'erminos de Estruch et al. integra las propiedades de las distancias de Nienhuys-Cheng y J. Ramon et al. para ello los autores definen el concepto de diferencias sint\'acticas de la expresi\'on, el tama\~no y el valor del contexto. A continuaci\'on se definen estos aspectos y finalmente esta distancia. 

Sean $s$ y $t$ dos expresiones, el conjunto de las diferencias sint\'acticas, denotado por  $O^\star(s,t)$, es definido como:

\[
\begin{array}{llll}
O^\star(s,t) & = &\{o \in O(s)\cap O(t):&\neg Compatible(s|_{o},t|_{o}) \textrm{ y } \\
& & & Compatible(s|_{o'},t|_{o'}),\forall o' \in Pre(o)\}\\
\end{array}
\]

Luego, la complejidad de las diferencias sint\'acticas entre $s$ y $t$ son calculadas sobre el n\'umero de s\'imbolos de subt\'erminos (en $s$ y $t$) en las ocurrencias que tiene $o\in O^\star(s,t)$. Para este prop\'osito, se utiliza una funci\'on especial de tama\~no de una expresi\'on denotada por $Size'(t)$ definida de la siguiente manera: 

Dada una expresi\'on $t=t_0(t_1,\ldots,t_n)$ entonces $Size'(t)=\frac{1}{4}Size(t)$ donde,

\[
Size(t_0(t_1,\ldots,t_n))= \left \{
\begin{array}{l}
1,\,n=0 \\
1+\frac{\sum_{i=1}^n Size(t_i)}{2(n+1)}, \,n>0
\end{array}\right .
\]

Por ejemplo, considerando $s=f(f(a),h(b),b)$, entonces $Size(a)=Size(b)=1$, $Size(f(a))=Size(h(b))=1+1/4=5/4$, $Size(s)=1+(5/4+5/4+1)/8=23/16$ y finalmente, $Size'(s)=23/64$. 

El valor de contexto de una ocurrencia $o$ en una expresi\'on $t$, $C(o;t)$, es usado para considerar la relaci\'on entre $t|_o$ y $t$ en el sentido que, un alto valor de $C(o;t)$ corresponde a una posici\'on de profundidad de $t|_o$ en $t$ o la existencia de supert\'erminos de $t|_o$ con un gran n\'umero de argumentos. Este concepto se formaliza de la siguiente manera: 

Sea $t$ una expresi\'on. Dada una ocurrencia $o \in O(t)$, el valor de contexto de $o$ en $t$, denotado por $C(o;t)$, es definido como:

\[
C(o;t)=\left \{
\begin{array}{l}
1,\,o=\lambda \\
2^{Length(o)}\cdot\prod_{\forall o' \in Pre(o)}(Arity(t|_{o'})+1),\,\textrm{otro caso} \\
\end{array}\right .
\]

Est\'a demostrado en \cite{estruch2010integrated} que si $o \in O^\star(s,t)$ entonces $C(o;s)=C(o;t)$. De esta manera, en estos casos, el contexto de una ocurrenca $o \in O^\star(s,t)$ es denotado por $C(o)$.
Las difrencias repetidas se manejan a trav\'es de una relaci\'on de equivalencia ($\sim$) sobre el conjunto $O^\star(s,t)$ definido de la siguiente manera:

\[
\forall o_i,o_j \in O^\star(s,t),\,\, o_i\sim o_j \Leftrightarrow s|_{o_i}=s|_{o_j} \textrm{ y } t|_{o_i}=t|_{o_j}
\]

\noindent el cual produce una particion $O^\star(s,t)$ que no se sobrepone en la equivalencia de clases. Tambi\'en para la relaci\'on $(\leq)$, en cada equivalencia de clase $O^\star_i(s,t)$, es definida como $\forall o_j,o_k \in O^\star_i(s,t),\,\, o_j \leq o_k \Leftrightarrow C(o_j) \leq C(o_k)$.

Adicionalmente, los conceptos previos son usados para definir otra funci\'on la cual simplemente asocia pesos para las ocurrencias de manera que al mayor $C(o)$, se asigna el menor de los pesos de $o$, es decir, el menos significativo de la diferencia sint\'actica referidos en $o$. Por lo tanto, dadas dos expresiones $s$ y $t$, la funci\'on de peso $w$ es:

\[\forall o\in O^\star_i(s,t),\,\, w(o)=\frac{3f_i(o)+1}{4f_i(o)}\]

\noindent donde $i=\pi(o)$, $\pi(o)$ es el \'indice de la equivalencia de clase perteneciente a $o$, y $f_i(o)$ es la posici\'on que tiene $o$ de acuerdo con $\leq$. 

\noindent Finalmente, la distancia de Estruch et al. es definida como:

\noindent Sean $s$ y $t$ dos expresiones, la distancia entre $s$ y $t$ es,

\[
d_E(s,t) = \sum_{o \in O^\star(s,t)} \frac{w(o)}{C(o)}\bigl ( Size'(s|_{o})+Size'(t|_{o})\bigr )
\]

Por ejemplo, sean $s=p(a,a)$ y $t=p(f(b),f(b))$ dos expresiones. Entonces, $O^\star(s,t)=\{1,2\}$. Adem\'as,

\[
C(1)=C(2)=2\cdot(2+1)=6
\]

\noindent Los tama\~nos de los subt\'erminos involucrados en el c\'alculo de la distancia son: 

\[
Size'(a)=1/4 \textrm{ and } Size'(f(b))=5/16
\]

\noindent Hay solo una clase de equivalencia $O^\star=O^\star_1(s,t)$. Se supone que la ocurrencia $1$ ocupa el primer lugar,

\[
w(1)=1 \textrm{ and } w(2)=7/8
\]

Finalmente, 

\[
d_E(s,t)=\frac{1}{6}\bigl(\frac{1}{4}+\frac{5}{16}\bigr)+\frac{7}{48}\bigl(\frac{1}{4}+\frac{5}{16}\bigr)
\]

\chapter{Transformaci\'on de datos semi"=estructurados en una representaci\'on basada en t\'erminos usando XML. }\label{transformaciones}

\setlength{\parskip}{\baselineskip}

Para aplicar distancias entre t\'erminos para diferentes tipos de datos se requiere un lenguaje com\'un que permita su representaci\'on. XML, por sus caracter\'isticas propias de su estructura, es un lenguaje flexible que permite la representaci\'on desde datos planos hasta datos jer\'arquicos y sobre el cual es posible aplicar distancias entre t\'erminos; sin embargo, es necesario controlar algunos aspectos su estructura; por consiguiente, con respecto a la estructura, dos situaciones extremas pueden ser identificadas (obviamente, existen casos intermedios que pueden ser manejados utilizando una mezcla de estas dos situaciones):
  
\begin{itemize}

\item \textbf{Datos planos:} muchos conjuntos de datos son dados como una tabla de pares atributo-valor. Sin embargo, si se hace una inspecci\'on detallada es posible identificar que algunos atributos se relacionan entre s\'i y pueden inducirse jerarqu\'ias sobre ellos.

\item \textbf{Datos jer\'arquicos:} otros conjuntos de datos consisten en datos con una estructura jer\'arquica que est\'a representada por un \'arbol o un t\'ermino funcional (por ejemplo en Haskell o LISP).
\end{itemize}

A continuaci\'on se hace una descripci\'on de c\'omo controlar los aspectos relacionados con la estructura de XML y c\'omo derivar esquemas XML a partir de datos planos y jer\'arquicos.

\section{Definici\'on del esquema}

Debido a que los documentos XML y los t\'erminos funcionales no son lo mismo, se deben tomar algunas decisiones con el fin de utilizar XML para representar los t\'erminos funcionales y adaptar los tipos de datos anteriores en una representaci\'on com\'un. Un aspecto clave es c\'omo manejar la profundidad, la repetici\'on y el orden; los elementos en XML deben tener orden y considerar la posibilidad de permitir repeticiones sobre diferentes niveles de profundidad.

Al aplicar estas distancias entre t\'erminos sobre una jerarqu\'ia, se debe garantizar su correcta correspondencia entre los diferentes elementos, ya que algunas partes pueden estar vac\'ias, y determinar la manera en que se deben comparar. En otras palabras, la jerarqu\'ia puede tener diferente n\'umero de elementos, en el mismo nivel, por lo que es necesario garantizar que los c\'alculos de la distancia no sean afectados por la ausencia de caracter\'isticas o por su orden. Esta dificultad requiere de la creaci\'on de un esquema general que permita a cada instancia, con sus caracter\'isticas propias, ajustarse apropiadamente en un orden definido y sin perder ning\'un elemento o contenido. El esquema propuesto es un documento XML que contiene etiquetas de todos los elementos que pueden existir para un ejemplo dado.

\begin{figure*}[h]
  \centering
   \subfloat[]{\label{fig:hierarchy1}\includegraphics[width=0.55\textwidth]{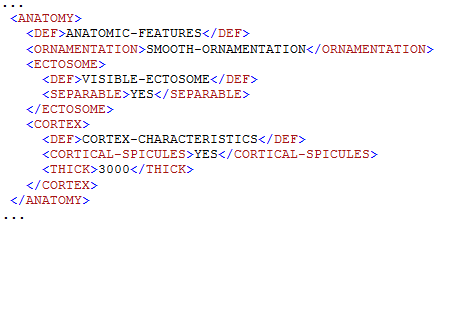}}
   \subfloat[]{\label{fig:hierarchy2}\includegraphics[width=0.55\textwidth]{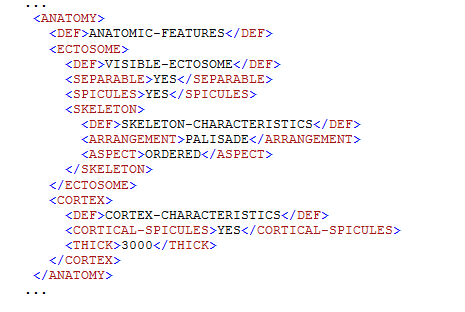}}  
   \caption{Ejemplo de jerarqu\'ias utilizando diferentes elementos}
   \label{fig:schema-definition}
 \end{figure*}

\begin{figure*}[h]
\centering
\includegraphics[width=0.3\textwidth]{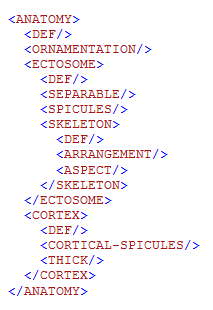}
\caption{Ejemplo de todas las posibles etiquetas utilizando las jerarqu\'ias de la figura \ref{fig:schema-definition}.}
\label{fig:schema}
\end{figure*}

Por ejemplo, la figura  \ref{fig:schema-definition} muestra dos jerarqu\'ias donde 
{
\ttfamily
\footnotesize
$<$SPICULES$>$
} y 
{
\ttfamily
\footnotesize
$<$SKELETON$>$ 
}
son elementos que existen en la jerarqu\'ia (b), pero no en (a); de la misma manera, el elemento 
{
\ttfamily
\footnotesize
$<$ORNAMENTATION$>$
}
existe para la jerarqu\'ia (a), pero no para (b). Por lo tanto, es necesario crear un esquema que contenga una estructura integrada para ambos casos; es decir, un esquema para las jerarqu\'ias de la figura \ref{fig:schema-definition} tendr\'ia, no s\'olo aquellos elementos comunes, sino tambi\'en aquellos que est\'an en alguna de las dos jerarqu\'ias como 
{
\ttfamily
\footnotesize
$<$SPICULES$>$
},
{
\ttfamily
\footnotesize
$<$SKELETON$>$ 
} y 
{
\ttfamily
\footnotesize
$<$ORNAMENTATION$>$.
} La figura   \ref{fig:schema} muestra todas las posibles etiquetas de un esquema utilizando las jerarqu\'ias (a) y (b) de la figura  \ref{fig:schema-definition}.

El esquema permite, no solo tener una estructura general, sino tambi\'en garantizar el formato, de manera que cada caso debe seguir esta estructura. Sin embargo, es necesario tener en cuenta que no todos los casos se pueden adaptar f\'acilmente a un ejemplo XML, por ejemplo, aquellos elementos que son hojas con contenido no pueden ser adaptados directamente cuando el esquema requiere un \'arbol. En esta situaci\'on el elemento es agregado independientemente de la estructura del esquema. Sin embargo, este hecho no afecta los c\'alculos de la distancia ya que se consideran como caracter\'isticas diferentes.

A pesar de que existen diferencias entre una representaci\'on basada en t\'erminos y una representaci\'on en XML es posible convertir una representaci\'on en otra (como se present\'o en la secci\'on \ref{Representacion-Basada-Terminos}). En XML, cada t\'ermino se representa como un elemento compuesto por una etiqueta y un valor; de esta manera, es posible convertir un t\'ermino en un elemento XML a\~nadiendo una etiqueta e insertando el t\'ermino como un valor. Tambi\'en es posible convertir un documento XML en un t\'ermino sin tener en cuenta la etiqueta y representado el valor como un t\'ermino. Por ejemplo, las figuras \ref{fig:name-induced-hierarchy} y \ref{fig:sponge-hierarchy-LISP-XML}  muestran c\'omo un documento XML puede representar dos tipos de jerarqu\'ia diferentes.
 
\section{Derivaci\'on de esquemas XML jer\'arquicos a partir de datos planos}

Los datos planos se refieren a problemas atributo-valor que son comunes en bases de datos, aprendizaje autom\'atico, miner\'ia de datos y otras \'areas.  En otras palabras, los datos son presentados en forma de una tabla de valores escalares. Sin embargo, en muchos casos, alg\'un tipo de estructura puede ser inferida a partir de estos conjuntos de datos, ya sea porque originalmente fueron as\'i, despu\'es de un proceso de aplanamiento, o porque algunas caracter\'isticas pueden ser agrupadas de acuerdo con alguna raz\'on. El hecho derivar una jerarqu\'ia a partir de datos planos es uno de los problemas del \'area del conocimiento de computaci\'on granular tratados en \cite{bargiela2003granular}.

Para este caso se consideran tres posibles fuentes de estructuras de representaci\'on de datos planos:

\begin{enumerate}

\item \textbf{Igualdad de valor:} muchos conjuntos de datos son dados como datos planos, pero al hacer una revisi\'on detallada es posible encontrar que algunos atributos est\'an relacionados por los valores que toman. Por ejemplo, si dos variables  $X_1$ y $X_2$ pueden tomar el valor $Este$, $Oeste$, $Norte$ y $Sur$, hay claramente una conexi\'on entre ellas que puede ser explotada, especialmente a trav\'es del uso de igualdades. Por ejemplo, es posible definir una condici\'on o regla usando  $X_1$ y $X_2$ que solo tiene sentido si los tipos de datos son iguales; de la misma manera en que las repeticiones de una variable son permitidas en t\'erminos funcionales, como $f(a,X,X)$.

\item \textbf{Jerarqu\'ia inducida por nombre:} en muchos casos es posible encontrar simples estructuras en la jerarqu\'ia de atributos de acuerdo con sus nombres o su sem\'antica. Por ejemplo {\em cap-shape}, {\em cap-surface} y {\em cap-color} son atributos que contienen sub-caracter\'isticas de ``{\em cap}''. De esta manera, puede ser creada una jerarqu\'ia de un grupo de caracter\'isticas.

\item  \textbf{Jerarqu\'ia de similitud de atributos:} en otras ocasiones donde los nombres y los valores  pueden ser diferentes, es posible establecer relaciones entre atributos los cuales pueden ser usados para inducir a una estructura. Un enfoque es conocido como un \'arbol de aglomeraci\'on de variables de Watanabe-Kraskov  \cite{watanabe1969knowing}\cite{kraskov2003hierarchical}, el cual construye un dendrograma (un \'arbol de jerarqu\'ias) usando la m\'etrica de similitud entre atributos. 

\end{enumerate}

\noindent A continuaci\'on se detallan estos tres aspectos utilzando algunos ejemplos.

\subsection{Igualdad de valor}

La igualdad de valor utiliza la relaci\'on existente entre los valores posibles de dos o m\'as variables; para el caso en que existan dominios similares, estas variables pueden ser marcadas para ser tenidas en cuenta como un criterio adicional al momento de un proceso de clasificaci\'on o agrupamiento. La tabla \ref{tab:igualdad-valor} muestra el dominio de algunos atributos del conjunto de datos de {\em mushroom} (de UCI $machine$ $learning$ $repository$ \cite{UCI}), en el cual los atributos {\em stalk-color-above-ring} y {\em stalk-color-below-ring} est\'an relacionados entre ellos; adicionalmente, es posible ver como {\em cap-color}, aunque no es exactamente igual, es muy similar son respecto a los otros dos atributos.

\vspace{0.8cm}

\begin{table}[htp]
\begin{center}
\begin{tabular}{|l|l|}\hline
 \makebox[3cm][c]{Atributo} &  \makebox[8cm][c]{Dominio} \\
\hline\hline
cap-color & brown, buff, cinnamon, gray, green, pink, purple, red, white, yellow\\
\hline
stalk-color-above-ring & brown, buff, cinnamon, gray, orange, pink, red, white, yellow \\
\hline
stalk-color-below-ring &  brown, buff, cinnamon, gray, orange, pink, red, white, yellow\\
\hline
\end{tabular}
\end{center}
\caption{Ejemplo similaridad entre dominos de variables} 
\label{tab:igualdad-valor}
\end{table}

\vspace{0.6cm}

La forma de relacionar estos atributos en XML depende de c\'omo  las repeticiones y relaciones entre estas variables son tenidas cuenta por el algoritmo utilizado para calcular la distancia. Este algoritmo podr\'ia tener en cuenta solo los valores de las variables o tener en cuenta el par atributo-valor. En el primer caso no es requerido ning\'un tipo de transformaci\'on; para el segundo caso es necesario transformar el nombre de las etiquetas de las variables relacionadas, en un nombre com\'un.

\subsection{Jerarqu\'ia inducida por nombre}

\begin{figure*}[htp]
  \centering
   \subfloat[Atributos originales]
   {\label{fig:mushroom-variables}\includegraphics[width=0.4\textwidth]{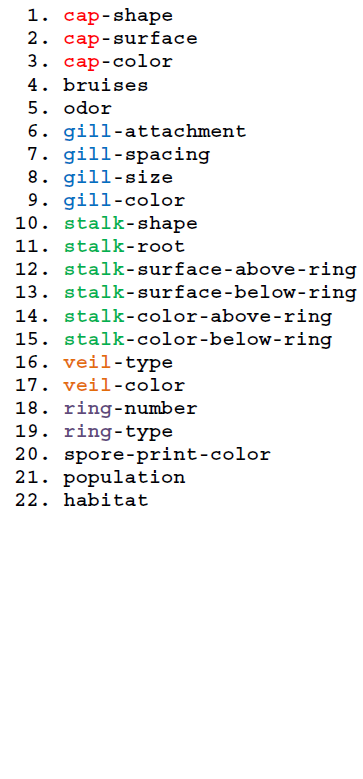}}
   \subfloat[Jerarqu\'ia inducida usando nombres comunes]
   {\label{fig:mushroom-XML}\includegraphics[width=0.42\textwidth]{images/mushroomXML.png}}  
   \caption{Jeraqu\'ia inducida por nombres para el conjunto de datos de mushroom}
   \label{fig:name-induced-hierarchy}
 \end{figure*}

La figura \ref{fig:name-induced-hierarchy} muestra un ejemplo de jerarqu\'ia la cual es inducida por los nombres de los atributos originales utilizando el conjunto de datos plano de {\em mushroom} de UCI \cite{UCI}, el cual originalmente no tiene una estructura (como se muestra en \ref{fig:name-induced-hierarchy}(a)). Despu\'es de agrupar por prefijos de nombres comunes se obtiene una jerarqu\'ia la cual es finalmente representada como un documento XML (mostrado en la figura \ref{fig:name-induced-hierarchy}(b)).  Este documento XML puede ser procesado como un t\'ermino funcional (como es presentado en \ref{Representacion-Basada-Terminos}); en concreto:

\noindent
{
\ttfamily
\footnotesize
\indent Mushroom(\\
\indent \indent cap(CONVEX, SMOOTH, WHITE), \\ 
\indent \indent BRUISES, ALMOND, \\
\indent \indent gill(FREE, CROWDED, NARROW, WHITE), \\
\indent \indent stalk(TAPERING, BULBOUS, surface(SMOOTH, SMOOTH), color(WHITE, WHITE)), \\ 
\indent \indent veil(PARTIAL, WHITE), \\
\indent \indent ring(ONE, PENDANT), \\
\indent \indent spore(print(BROWN)), \\
\indent \indent SEVERAL, WOODS) \\
}

\subsection{Jerarqu\'ia de similitud de atributos}

Este enfoque se basa en la idea de encontrar la similitud entre los atributos. Cuando los atributos son num\'ericos, normalmente se realiza a trav\'es de medidas de correlaci\'on. En el caso de atributos nominales, se pueden utilizar otras medidas de asociaci\'on como una prueba Chi-cuadrado; de esta manera se construye una matriz de similitud de atributos y, como se mencion\'o anteriormente, un dendrograma.

\begin{table}[htp]
\begin{center}
\begin{tabular}{|l|l|r|}\hline
 \makebox[2.3cm][c]{Atributo 1} &  \makebox[2.3cm][c]{Atributo 2}  & \makebox[2.3cm][c]{Prueba} \\
  & &\makebox[2.3cm][c]{chi-cuadrado} \\
\hline\hline
cap-shape & cap-surface & 83.49\\
\hline
cap-shape & cap-color & 215.39\\
\hline
cap-shape & bruises & 87.10\\
\hline
cap-shape & odor & 327.91\\
\hline
cap-shape & gill-attachment & 4.30\\
\hline
cap-shape  & \makebox[2.3cm][c]{\begin{sideways} ... \ \end{sideways}} & \makebox[2.3cm][c]{\begin{sideways}  ... \end{sideways}} \\
\hline
\makebox[2.3cm][c]{\begin{sideways} ... \ \end{sideways}} & \makebox[2.3cm][c]{\begin{sideways} ... \end{sideways}} & \makebox[2.3cm][c]{\begin{sideways}  ... \end{sideways}} \\
\hline
gill-attachment & cap-shape & 4.30\\
\hline
gill-attachment & cap-surface & 3.71\\
\hline
gill-attachment & cap-color & 201.46\\
\hline
gill-attachment & bruises & 3.33\\
\hline
gill-attachment & odor & 358.79\\
\hline
gill-attachment& \makebox[2.3cm][c]{\begin{sideways} ... \ \end{sideways}} & \makebox[2.3cm][c]{\begin{sideways} ... \end{sideways}}\\
\hline
\makebox[2.3cm][c]{\begin{sideways} ... \ \end{sideways}} & \makebox[2.3cm][c]{\begin{sideways} ... \end{sideways}} & \makebox[2.3cm][c]{\begin{sideways} ... \end{sideways}} \\
\hline
\end{tabular}
\end{center}
\caption{Ejemplo de asociaci\'on entre variables utilizando la prueba Chi-cuadrado} 
\label{tab:asociacion-entre-variables}
\end{table}

La tabla \ref{tab:asociacion-entre-variables} es un ejemplo de asociaci\'on entre atributos utilizando una muestra del conjunto de datos {\em mushroom} con la prueba Chi-cuadrado de Weka \cite{hall2009weka} como medida de similitud. Posteriormente, la tabla \ref{tab:matriz-similitud-atributos} muestra la matriz de similitud resultante constrida a partir de los resultados de la tabla \ref{tab:asociacion-entre-variables}. 

\begin{table}[htp]
\begin{center}
\begin{tabular}{|l|r|r|r|r|r|r|c|}\hline
 &  \makebox[1.4cm][c]{cap-shape} & \makebox[1.4cm][c]{cap-surface}  & \makebox[1.4cm][c]{cap-color} & \makebox[1.4cm][c]{bruises} & \makebox[1.4cm][c]{odor}& \makebox[1.4cm][c]{gill-}  & \makebox[0.6cm][c]{...}\\
&   &  & & & & \makebox[1.3cm][c]{attachment}  & \\
\hline\hline
cap-shape &  0 & 83.49 & 215.39 & 87.10 & 327.91 & 4.30 & ...\\
\hline
cap-surface &  83.49 & 0 & 167.81 & 4.04 & 135.83 & 3.71 & ...\\
\hline
cap-color &  215.39 & 167.81 & 0 & 56.75 & 520.88 & 201.46 & ...\\
\hline
bruises &  87.10 & 4.04 &  56.75 & 0 & 242.75 & 3.33 & ...\\
\hline
gill-attachment &  327.91 & 135.83 & 520.88 & 242.75 & 0 &358.79 & ...\\
\hline
odor &  4.30 & 3.71 & 201.46 & 3.33 & 358.79 & 0 & ...\\
\hline
\makebox[1.4cm][c]{\begin{sideways} ... \ \end{sideways}} &  \makebox[1.4cm][c]{\begin{sideways} ... \end{sideways}} & \makebox[1.4cm][c]{\begin{sideways} ... \end{sideways}} & \makebox[1.4cm][c]{\begin{sideways} ... \end{sideways}} & \makebox[1.4cm][c]{\begin{sideways} ... \end{sideways}} & \makebox[1.4cm][c]{\begin{sideways} ... \end{sideways}} & \makebox[1.4cm][c]{\begin{sideways} ... \end{sideways}} & \makebox[0.6cm][r]{0} \\
\hline
\end{tabular}
\end{center}
\caption{Matriz de similitud entre atributos} 
\label{tab:matriz-similitud-atributos}
\end{table}

\begin{figure*}[htp]
\centering
\includegraphics[width=0.72\textwidth]{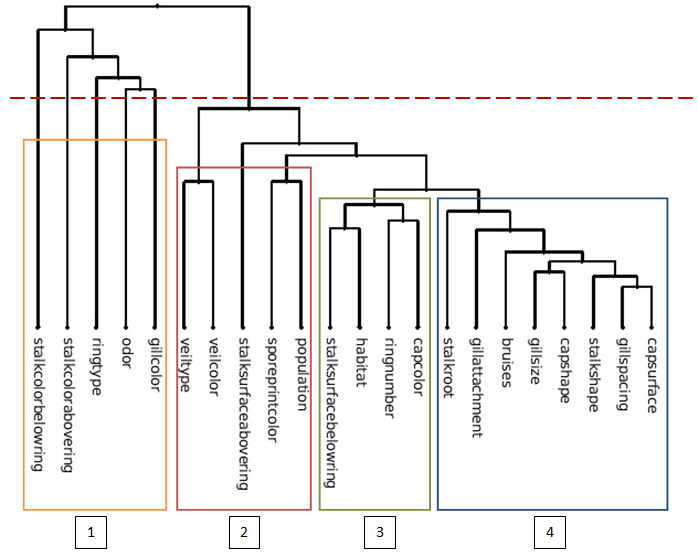}
\caption{Jerarquia de similitud de atributos del conjunto de datos de {\em mushroom}}
\label{fig:similarity-induced-hierarchy}
\end{figure*}

\begin{figure*}[htp]
\centering
\includegraphics[width=0.6\textwidth]{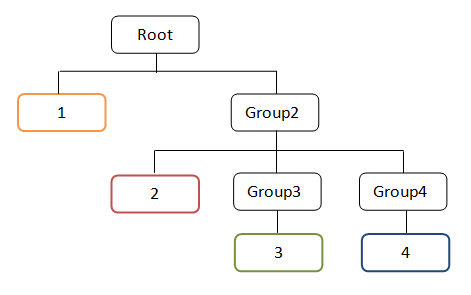}
\caption{Jerarquia simplificada basada en la figura \ref {fig:similarity-induced-hierarchy}}
\label{fig:similarity-induced-hierarchy2}
\end{figure*}

Con los datos de la matriz de similitud (mostrada en la tabla \ref{tab:matriz-similitud-atributos}) se construye un dendrograma utilizando los 22 atrubutos del conjunto de datos anterior. En este dendrograma, en lugar de usar toda la jerarqu\'ia, s\'olo se utilizan los segmentos m\'as grandes, debido a que algunas variables quedar\'ian muy profundas y con muy baja ponderaci\'on. La figura \ref{fig:similarity-induced-hierarchy} muestra el dendrograma resultante (construido por medio la utilidad `DendroUPGMA' \cite{DendroUPGMA}) e identificando cuatro grupos como se muestra en la figura \ref{fig:similarity-induced-hierarchy2}.

\vspace{0.4cm}

A partir de esto, es colocada cada variable en un grupo que puede ser representado como un t\'ermino funcional:

\noindent
{
\ttfamily
\footnotesize
\indent Mushroom(WHITE,WHITE, PENDAT, AMOND, WHITE, \\
\indent \indent  Group2(PARTIAL, WHITE, SMOOTH, BROWN, SEVERAL, \\
\indent \indent \indent  Group3(SMOOTH, WOODS, ONE, WHITE), \\
\indent \indent \indent Group4(TAPERING, FREE, CROWDED, NARROW, BRUISES, BULBOUS, CONVES, SMOOTH))) 
}

\section{Derivaci\'on de esquemas XML jer\'arquicos a partir de datos  jer\'arquicos}

En este punto se trata de la transformaci\'on necesaria cuando los datos ya tienen originalmente una rica estructura jer\'arquica. Este tipo de derivaciones son m\'as directas; consisten en transformaciones simples que deben garantizar una estructura bien formada para el documento XML; es necesario establecer si el orden, repeticiones y etiquetas son pertinentes  para determinar las caracter\'isticas y la ubicaci\'on correcta en la estructura XML. Un aspecto espec\'ifico, como se mencion\'o anteriormente, es la forma de tratar las partes vac\'ias de la jerarqu\'ia y la forma de compararlas con los t\'erminos que no est\'en vac\'ios. Este enfoque se basa en un esquema com\'un para todos los ejemplos y la suposici\'on de que la distancia entre una parte vac\'ia y una no vac\'ia es 1 (no est\'a dado por el tama\~no de una parte vac\'ia).

\begin{figure*}[htp]
\centering
\includegraphics[width=1\textwidth]{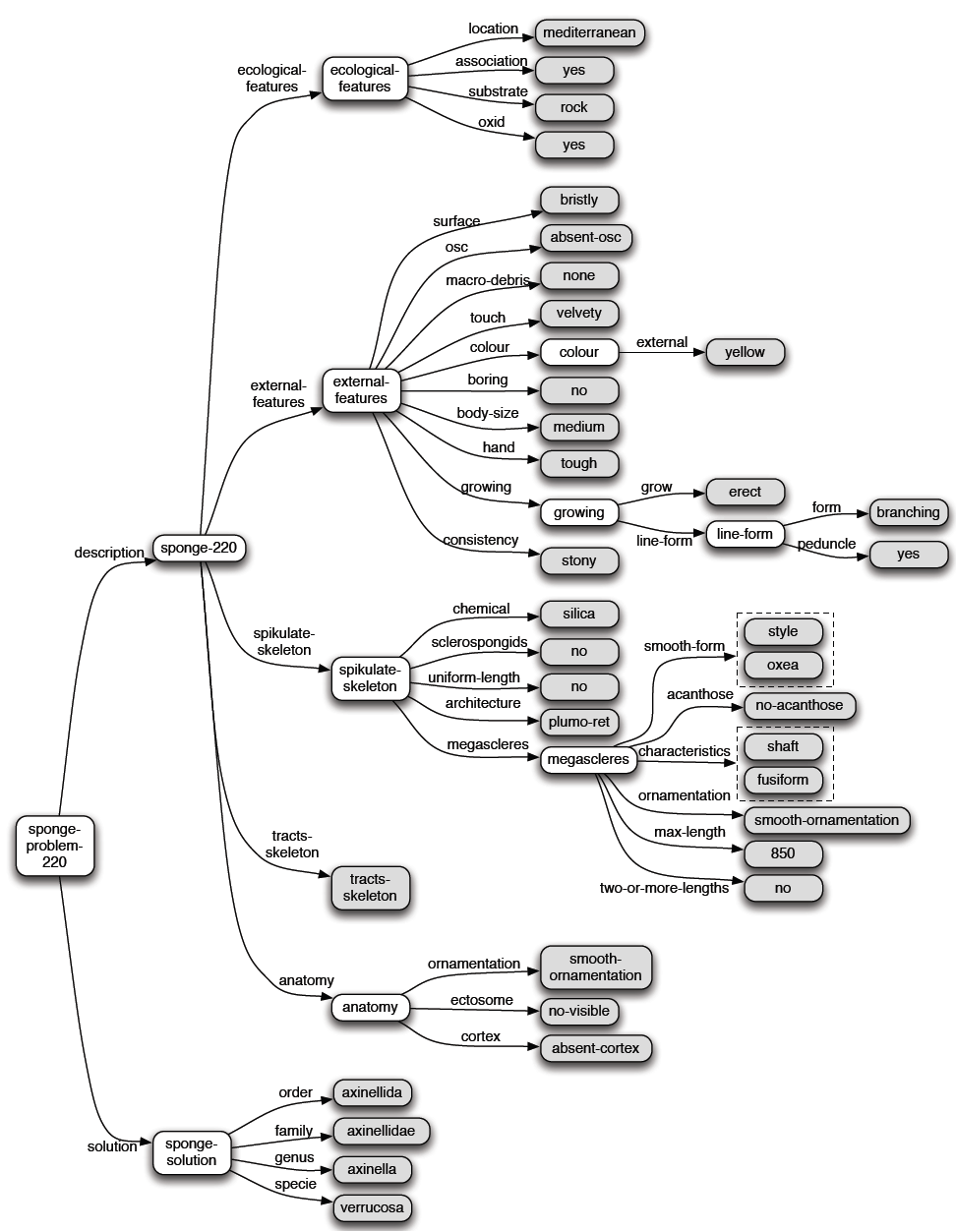}
\vspace{-0.8cm}
\caption{Jerarqu\'ia compleja (ejemplo 220) del conjunto de datos de sponge}
\label{fig:sponge-hierarchy}
\end{figure*}

\begin{figure*}[htp]
  \centering
   \subfloat[Jerarqu\'ia en LISP]
   {\label{fig:original-sponge-hierarchy-LISP}\includegraphics[width=0.57\textwidth]{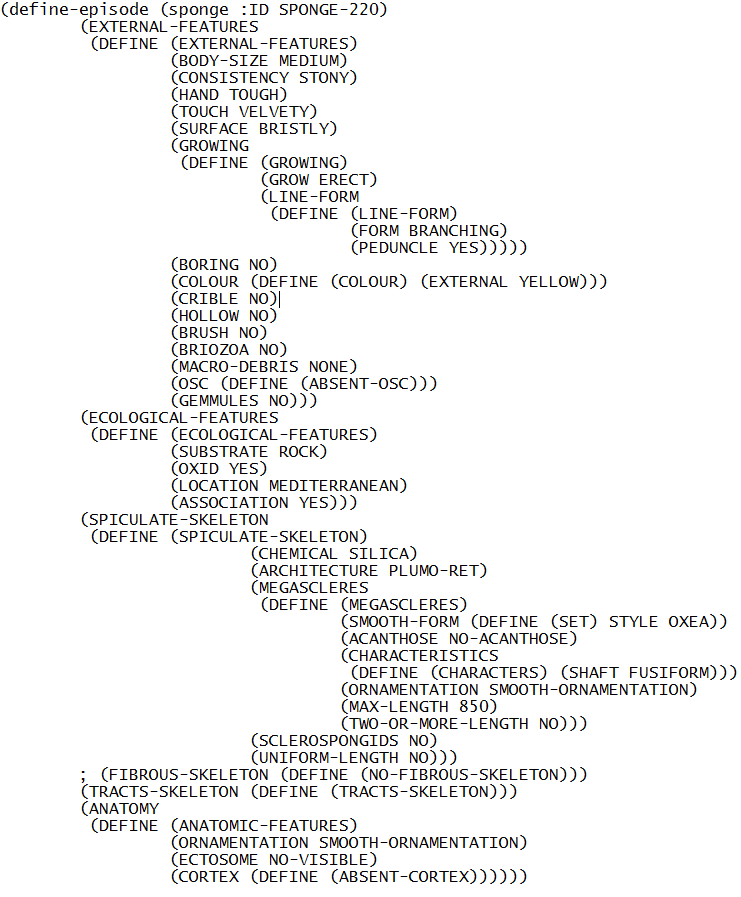}}
   \subfloat[Jerarqu\'ia representada como un documento XML]
   {\label{fig:sponge-hierarchy-XML}\includegraphics[width=0.42\textwidth]{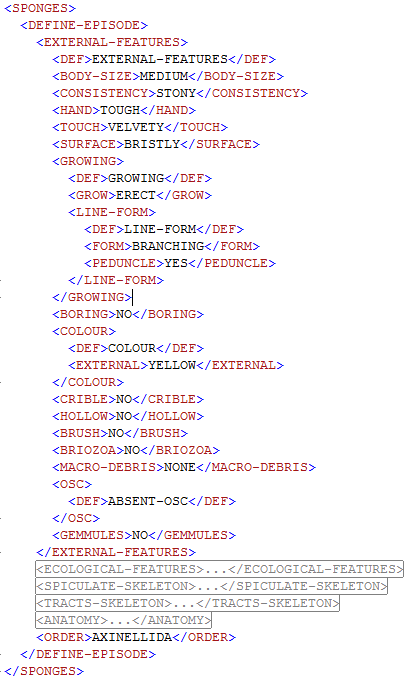}}  
   \caption{T\'ermino en LISP y su representaci\'on en XML del conjunto de datos de sponge}
   \label{fig:sponge-hierarchy-LISP-XML}
   \vspace{-0.4cm}
 \end{figure*}

La figura \ref{fig:sponge-hierarchy} muestra la estructura del ejemplo 220 del conjunto de datos `sponge' del UCI repository \cite{UCI}. La figura \ref{fig:sponge-hierarchy-LISP-XML} muestra este mismo objeto representado como un t\'ermino en LISP y su represntaci\'ion en un documento XML. Este es un conjunto de datos complejo que cuenta con una rica estructura, donde es posible resaltar que los m\'etodos proposicionales cl\'asicos no son aplicables. 

\chapter{Experimentos}\label{experimentos}

En este cap\'itulo se incluyen experimentos sobre un proceso de clasificaci\'on usando algunas distancias entre t\'erminos introducidas en la cap\'itulo \ref{prelim}, con las transformaciones presentadas en la cap\'itulo previo (cap\'itulo \ref {transformaciones}).

Para el proceso de clasificaci\'on se utiliza el algoritmo $k$-NN con la variante de ponderaci\'on, usando la funci\'on de {\em atracci\'on}. $k$-NN es un algoritmo simple y adecuado para analizar el efecto de las distancias y las transformaciones. Como se mencion\'o en la secci\'on \ref{KNN}, este algoritmo clasifica una instancia en la clase m\'as com\'un de los $k$ ejemplos m\'as cercanos usando la distancia. El valor de $k$ para los siguientes experimentos es calculado como $\sqrt{n}$ (donde $n$ es el n\'umero de ejemplos), que es una manera com\'unmente utilizada para definir el valor de $k$ en este algoritmo. La funci\'on de {\em atracci\'on}, definida como $\frac{1}{d^i}$ (donde $d$ es la distancia e $i$ es el {\em par\'ametro de atracci\'on}), da mayor ponderaci\'on a los ejemplos m\'as cercanos de $k$ (en $k$-NN no ponderado, $i$ es igual a $0$); en otras palabras, mientras mayor sea $i$, menor importancia tiene $k$. En los experimentos se usan varios valores para $i$ que var\'ian de $0$ a $3$.

Se consideraron tres distancias: la distancia de Nienhuys-Cheng, la distancia de Estruch et al. y la distancia eucl\'idea; las cuales son denotadas por $d_{N}$, $d_{E}$ y $d_{U}$, respectivamente.

Para la evaluaci\'on experimental,  se utilizaron tres conjuntos de datos del UCI {\em repository} \cite{UCI}; dos de ellos, {\em mushroom} y {\em soybean}, son conjuntos de datos planos que son usados de diferentes maneras: inicialmente con su estructura plana y, porteriormente, jerarquizados utilizando dos m\'etodos diferentes. El tercer conjunto de datos, {\em Demospongiae} ({\em sponge}), es un documento LISP el cual fue transformado en un documento XML, preservando la jerarqu\'ia y los valores entre los atributos. 

Las tablas de resultados, para los conjuntos de datos de {\em mushroom} y {\em soybean}, incluyen los valores medios de los experimentos utilizando validaci\'on cruzada con 10 pliegues. Para la prueba de significancia entre los resultados de los m\'etodos, se utiliza {\em t-Student} con una confianza de 95\%; si la diferencia de un m\'etodo con respecto a otro es significativo, se incluyen su acr\'onimo ($d_N, d_E, d_U$) en la celda. En el conjunto de datos de {\em sponge} se utiliza 60\% de ejemplos de entrenamiento y 40\% para ejemplos de prueba.

\section{Conjunto de datos {\em Mushroom}}

Este conjunto de datos incluye descripciones de muestras hipot\'eticas correspondientes a 23 especies de setas; cada especie es identificada como comestible o venenosa. Este conjunto de datos cuenta con $8416$ ejemplos, de los cuales se extrajo una muestra aleatoria de 1000 de ellos para el proceso de clasificaci\'on.

\begin{figure*}[htp]
\centering
\includegraphics[width=0.7\textwidth]{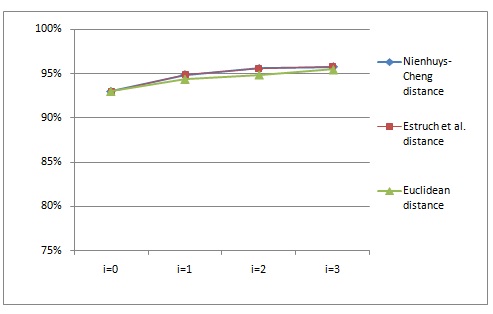}
\caption{Porcentajes de aciertos para el conjunto de datos de {\em mushroom} sin jerarqu\'ias.}
\label{fig:mush-without-hierarchies}
\end{figure*}

Primero se comparan las tres distancias usando un esquema plano (sin usar ning\'un tipo de jerarqu\'ia) y con diferentes valores para el par\'ametro $i$. Los resultados, presentados en la tabla \ref{tab-mush-without-hierarchies} y la figura \ref{fig:mush-without-hierarchies}, muestran que las diferencias entre las tres distancias no son significativas.

\vspace{0.4cm}

\begin{table}[htp]
\begin{center}
\begin{tabular}{|l|c|c|c|c|}\hline
  & $i=0$ & $i=1$ & $i=2$ & $i=3$ \\
  &$(\%)$ &$(\%)$ &($\%)$ &$(\%)$\\
\hline\hline
 & & & &  \\
Distancia de Nienhuys-Cheng& $93.0$ & $94.9$ & $95.6$ & $95.8$   \\
\makebox[4cm][c]{$d_{N}$} &  &  &  &  \\
\hline
  & & & & \\
Distancia de Estruch et al.& $93.0$ & $94.9$ & $95.6$ & $95.8$   \\
\makebox[4cm][c]{$d_{E}$} &  &  &  &  \\
\hline
 & & & & \\
Distancia eucl\'idea& $93.0$ & $94.4$ & $94.9$ & $95.5$   \\
\makebox[4cm][c]{$d_{U}$}&  &  &  &  \\
\hline
\end{tabular}
\caption{Porcentajes de aciertos para el conjunto de datos de {\em mushroom} sin jerarqu\'ias.}
\label{tab-mush-without-hierarchies}
\end{center}
\end{table}

\vspace{0.4cm}

\vspace{0.4cm}

\begin{table}[htp]
\begin{center}
\begin{tabular}{|l|c|c|c|c|}\hline
  & $i=0$ & $i=1$ & $i=2$ & $i=3$ \\
  &$(\%)$ &$(\%)$ &($\%)$ &$(\%)$\\
\hline\hline
 & & & & \\
Distancia de Nienhuys-Cheng& $99.8$ & $99.8$ & $99.9$ & $99.9$   \\
\makebox[4cm][c]{$d_{N}$}  &  &  &  &  \\
\hline
  & & & & \\
Distancia de Estruch et al. & $99.8$ & $99.8$ & $99.8$ & $99.8$   \\
\makebox[4cm][c]{$d_{E}$}&  &  &  &  \\
\hline
& & & & \\
Distancia eucl\'idea& $93.0$ & $94.4$ & $94.9$ & $95.5$   \\
\makebox[4cm][c]{$d_{U}$}&  &  &  &  \\
\hline
\end{tabular}
\caption{Porcentajes de aciertos para el conjunto de datos de {\em mushroom} con una jerarqu\'ia derivada del nombre de los atributos.}
\label{tab-mush-att-associated}
\end{center}
\end{table}

\begin{figure*}[htp]
\centering
\includegraphics[width=0.7\textwidth]{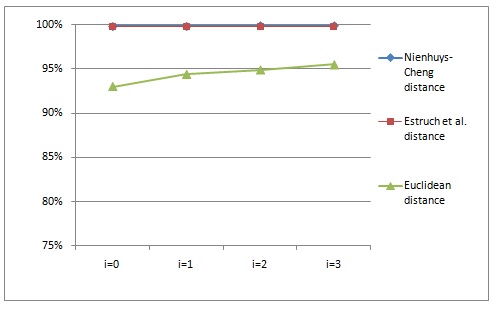}
\caption{Porcentajes de aciertos para el conjunto de datos de {\em mushroom} con una jerarqu\'ia derivada del nombre de los atributos.}
\label{fig:mush-att-associated}
\end{figure*}

La tabla \ref{tab-mush-att-associated} y la figura \ref{fig:mush-att-associated} muestra los resultados usando una jerarqu\'ia inducida por los nombres de los atributos. Con esta estructura, es posible encontrar diferencias entre las distancias; el desempe\~no de las distancias de Estruch et al. y Nienhuys-Cheng es \'optimo, mientras que los resultados de distancia eucl\'idea no se ven afectados por la estructura y, al igual que la ejecuci\'on anterior, solo se incrementa el porcentaje de aciertos cuando el valor de $i$ tambi\'en es incrementado.

\vspace{0.4cm}

\begin{figure*}[htp]
\centering
\includegraphics[width=0.7\textwidth]{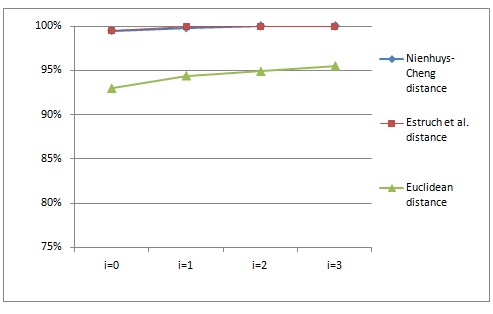}
\caption{Porcentajes de aciertos para el conjunto de datos de {\em mushroom} de acuerdo con la similitud entre atributos.}
\label{fig:Mush-correlation-att}
\end{figure*}

Finalmente, usando el otro m\'etodo de agrupaci\'on, que considera la similitud entre atrubutos, basado en la m\'etrica de Chi-cuadrado (ChiSquaredAttributeEval + Ranker en Weka \cite{hall2009weka}) y los grupos mostrados en la figura \ref{fig:similarity-induced-hierarchy}, se obtienen los resultados presentados en la tabla \ref{tab-Mush-correlation-att} y la figura \ref{fig:Mush-correlation-att}. En estos resultados, las distancias entre t\'erminos aprovechan esta estructura y obtiene nuevamente mejores resultados que la distancia eucl\'idea.  

\vspace{0.4cm}

\begin{table}[htp]
\begin{center}
\begin{tabular}{|l|c|c|c|c|}\hline
  & $i=0$ & $i=1$ & $i=2$ & $i=3$ \\
  &$(\%)$ &$(\%)$ &($\%)$ &$(\%)$\\
\hline\hline
 & & & & \\
Distancia de Nienhuys-Cheng& $99.5$ & $99.8$ & $100$ & $100$   \\
\makebox[4cm][c]{$d_{N}$}&  &  &  &  \\
\hline
  & & & & \\
Distancia de Estruch et al. & $99.5$ & $100$ & $100$ & $100$   \\
\makebox[4cm][c]{$d_{E}$}  &  &  &  &  \\
\hline
 & & & & \\
Distancia eucl\'idea& $93.0$ & $94.4$ & $94.9$ & $95.5$   \\
\makebox[4cm][c]{$d_{U}$}&  &  &  &  \\
\hline
\end{tabular}
\caption{Porcentajes de aciertos para el conjunto de datos de {\em mushroom} de acuerdo con la similitud entre atributos.}
\label{tab-Mush-correlation-att}
\end{center}
\end{table}

\vspace{0.4cm}

\section{Conjunto de datos {\em Soybean}}

El conjunto de datos {\em soybean}, al igual que {\em mushroom}, es un conjunto de datos plano, compuesto de 307 instancias y cada instancia tiene 35 atributos que determinan 19 clases diferentes de enfermedades de la soja. Las ejecuciones realizadas con este conjunto de datos son iguales a las realizadas para {\em mushroom}: 1) sin utilizar ning\'un tipo de jerarqu\'ia, 2) creando una jerarqu\'ia a partir de los nombres y valores de los atributos y 3) creando una jerarqu\'ia de acuerdo con la similitud de los atributos.

\begin{figure*}[htp]
\centering
\includegraphics[width=0.7\textwidth]{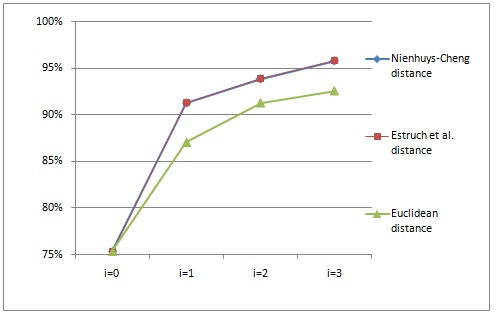}
\caption{Porcentajes de aciertos para el conjunto de datos de {\em soybean} sin jerarqu\'ias.}
\label{fig:soyB-without-hierarchies}
\end{figure*}

La tabla \ref{tab-soyB-without-hierarchies} y la figura \ref{fig:soyB-without-hierarchies} muestra los resultados de las ejecuciones con las tres distancias usando la versi\'on plana del conjunto de datos. En este caso, cuando $i=0$ las tres distancias son iguales; sin embargo, al incrementar $i$ las distancias entre t\'erminos superan un poco la distancia eucl\'idea, aunque sus diferencias no son estad\'isticamente significativas.

\vspace{0.4cm}

\begin{table}[htp]
\begin{center}
\begin{tabular}{|l|c|c|c|c|}\hline
  & $i=0$ & $i=1$ & $i=2$ & $i=3$ \\
  &$(\%)$ &$(\%)$ &($\%)$ &$(\%)$\\
\hline\hline
 & & & & \\
Distancia de Nienhuys-Cheng& $75.3$ & $91.3$ & $93.9$ & $95.8$   \\
\makebox[4cm][c]{$d_{N}$}  &  &  &  &  \\
\hline
& & & & \\
Distancia de Estruch et al.&$75.3$ & $91.3$ & $93.9$ & $95.8$   \\
\makebox[4cm][c]{$d_{E}$}  &  &  &  &  \\
\hline
& & & & \\
Distancia eucl\'idea&$75.3$ & $87.1$ & $91.3$ & $92.6$   \\
\makebox[4cm][c]{$d_{U}$}&  &  &  &  \\
\hline
\end{tabular}
\caption{Porcentajes de aciertos para el conjunto de datos de {\em soybean} sin jerarqu\'ias.}
\label{tab-soyB-without-hierarchies}
\end{center}
\end{table}

\vspace{0.4cm}

\vspace{0.4cm}

\begin{table}[htp]
\begin{center}
\begin{tabular}{|l|c|c|c|c|}\hline
  & $i=0$ & $i=1$ & $i=2$ & $i=3$ \\
  &$(\%)$ &$(\%)$ &($\%)$ &$(\%)$\\
\hline\hline
 & & & & \\
Distancia de Nienhuys-Cheng& $78.6$ & $89.6$ & $92.6$ & $93.2$   \\
\makebox[4cm][c]{$d_{N}$}  &  &  &  &  \\
\hline
 & & & & \\
Distancia de Estruch et al. & $76.0$ & $88.0$ & $91.6$ & $93.5$   \\
\makebox[4cm][c]{$d_{E}$}  &  &  &  &  \\
\hline
 & & & & \\
Distancia eucl\'idea   & $75.3$ & $87.1$ & $91.3$ & $92.6$   \\
\makebox[4cm][c]{$d_{U}$}&  &  &  &  \\
\hline
\end{tabular}
\caption{Porcentajes de aciertos para el conjunto de datos de {\em soybean} con una jerarqu\'ia derivada del nombre y valor de los atributos.}
\label{tab-soyB-att-associated}
\end{center}
\end{table}

\begin{figure*}[htp]
\centering
\includegraphics[width=0.7\textwidth]{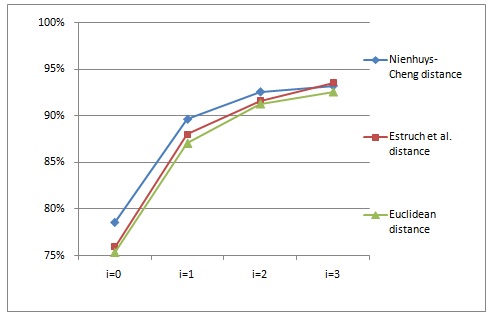}
\caption{Porcentajes de aciertos para el conjunto de datos de {\em soybean} con una jerarqu\'ia derivada del nombre y valor de los atributos.}
\label{fig:soyB-att-associated}
\end{figure*}

La tabla \ref{tab-soyB-att-associated} y la figura \ref{fig:soyB-att-associated} muestran el desempe\~no utilizando una jerarqu\'ia construida a partir de los nombres y valores de los atributos; el comportamiento de las tres distancias es similar y las diferencias no son significativas.

\vspace{0.4cm}

\vspace{0.4cm}

\begin{table}[htp]
\begin{center}
\begin{tabular}{|l|c|c|c|c|}\hline
  & $i=0$ & $i=1$ & $i=2$ & $i=3$ \\
  &$(\%)$ &$(\%)$ &($\%)$ &$(\%)$\\
\hline\hline
 & & & & \\
Distancia de Nienhuys-Cheng & $86.1$ & $99.0$ & $99.4$ & $99.4$   \\
 \makebox[4cm][c]{$d_{N}$}& ${d_{U}}$ & ${d_{U}}$ & ${d_{U}}$ & ${d_{U}}$ \\
\hline
& & & & \\
Distancia de Estruch et al.& $86.1$ & $99.0$ & $99.4$ & $99.4$   \\
\makebox[4cm][c]{$d_{E}$} & ${d_{U}}$ & ${d_{U}}$ & ${d_{U}}$ & ${d_{U}}$ \\
\hline
& & & & \\
Distancia eucl\'idea &$75.3$ & $87.1$ & $91.3$ & $92.6$    \\
\makebox[4cm][c]{$d_{U}$} & ${d_{N}}{d_{E}}$ & ${d_{N}}{d_{E}}$ & ${d_{N}}{d_{E}}$ & ${d_{N}}{d_{E}}$ \\
\hline
\end{tabular}
\caption{Porcentajes de aciertos para el conjunto de datos de {\em soybean} de acuerdo con la similitud entre atributos.}
\label{tab-SoyBean-correlation-att}
\end{center}
\end{table}

\begin{figure*}[htp]
\centering
\includegraphics[width=0.7\textwidth]{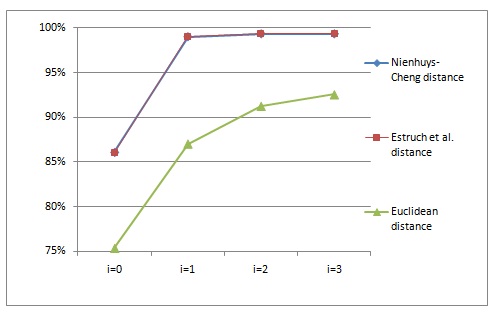}
\caption{Porcentajes de aciertos para el conjunto de datos de {\em soybean} de acuerdo con la similitud entre atributos.}
\label{fig:SoyBean-correlation-att}
\end{figure*}

Finalmente, para este conjunto de datos, utilizando la m\'etrica Chi-cuadrado para identificar la similitud entre atributos, se obtienen los resultados de la tabla \ref{tab-SoyBean-correlation-att} y la figura \ref{fig:SoyBean-correlation-att}. Para esta situaci\'on, las diferencias son estad\'isticamente significativas; las distancias entre t\'erminos obtienen mejores resultados que la distancia eucl\'idea, que siempre es igual y no es afectada por la estructura del conjunto de datos.

\vspace{0.4cm}

Utilizando la similitud entre atributos y las distancias entre t\'erminos se puede mejorar significativamente el desempe\~no de un proceso de clasificaci\'on aplicado sobre conjuntos de datos planos.

\section{Conjunto de datos {\em Demospongiae} }\label{sec:sponge-Dataset}

El conjunto de datos {\em Demospongiae} es una estructura originalmente jer\'arquica con 503 instantacias de esponjas marinas; cada instancia es representada como un \'arbol usando t\'erminos en el lenguaje LISP. Cada \'arbol tiene entre 5 y 8 niveles de profundidad y el n\'umero de hojas var\'ia entre 17 y 51 (ver la figura \ref{fig:sponge-hierarchy}).

De este conjunto de datos, se extrae un documento XML bien formado que preserva la estructura original. Para hacer esta transformaci\'on, cada l\'inea en LISP es convertida en uno o varios elementos XML bien formados, por medio de la asignaci\'on de nombres y valores de cada elemento de cada una de las caracter\'isticas descritas en LISP, acuerdo con su jerarqu\'ia y orden. Algunas transformaciones fueron definidas de la siguiente manera:

\begin{itemize}
\item Cada ejemplo  en  LISP, definido por la etiqueta {\ttfamily `define-episode'}  y un identificador  {\ttfamily `sponge :ID SPONGE-0'} fue transformado por un elemento XML:  {\ttfamily $<$DEFINE-EPISODE$>$}, el cual adiciona un elemento hijo:  {\ttfamily $<$SPONGE\_ID$>$SPONGE-0$<$/SPONGE\_ID$>$}.
\item Cada caracter\'istica definida como  {\ttfamily `EXTERNAL-FEATURES (DEFINE (EXTERNAL-FEATURES)' } es convertida en un elemento principal contenedor de varios elementos XML:  {\ttfamily $<$EXTERNAL- FEATURES$>$}.
\item Cada caracteristica simple como  {\ttfamily `(BODY-SIZE SMALL)' } es convertida en un elemento  {\ttfamily $<$BODY-SIZE$>$SMALL$<$/BODY-SIZE$>$}.
\item Adicionalmente, las coleciones como  {\ttfamily `(SET) GREY WHITISH' }, es convertida en XML de la siguiente manera:  {\ttfamily $<$SET1$>$GREY$<$/SET1$>$ $<$SET2$>$WHITISH$<$/SET2$>$}.
\end{itemize} 

\begin{figure*}[htp]
\centering
\includegraphics[width=0.7\textwidth]{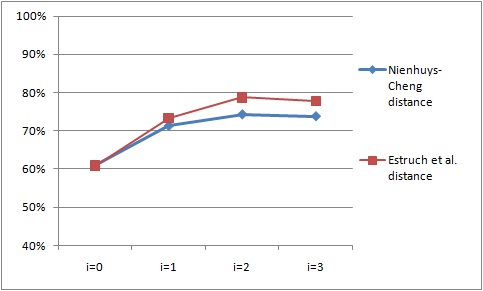}
\caption{Porcentajes de aciertos para el conjunto de datos de {\em sponge} usando su jerarqu\'ia original.}
\label{fig:sponges}
\end{figure*}

La tabla \ref{tab-sponges} y la figura \ref{fig:sponges} muestran los resultados de la clasificaci\'on utilizando el documento XML generado a partir de las transformaciones anteriores. Con este conjunto de datos solo se aplican las distancias entre t\'erminos porque no es posible aplicar directamente la distancia eucl\'idea; la distancia de Estruch et al. muestra mejores resultados que la distancia de Nienhuys-Cheng (cuando $i$ es mayor que 0). Este conjunto de datos es \'util para tener en cuenta las ventajas de las propiedades intr\'insecas de las distancias entre t\'erminos; es un buen ejemplo para ilustrar las diferencias entre este tipo de distancias debido a que los datos son puramente semi-estructurados y se percibe el efecto de las repeticiones.

\begin{table}[htp]
\begin{center}
\begin{tabular}{|l|c|c|c|c|}\hline
  & $i=0$ & $i=1$ & $i=2$ & $i=3$ \\
  &$(\%)$ &$(\%)$ &($\%)$ &$(\%)$\\
\hline\hline
 & & & & \\
Distancia de Nienhuys-Cheng & 60.9 & 71.3 & 74.3 & 73.8   \\
 \makebox[4cm][c]{$d_{N}$}& & & & \\
\hline
& & & & \\
Distancia de Estruch et al.& 60.9& 73.3 & 78.7 & 77.7   \\
\makebox[4cm][c]{$d_{E}$} & & & & \\
\hline
\end{tabular}
\caption{Porcentajes de aciertos para el conjunto de datos de {\em sponge} usando su jerarqu\'ia original.}
\label{tab-sponges}
\end{center}
\end{table}

\chapter{Conclusiones y trabajos futuros}\label{conclusiones}

Las distancias entre t\'erminos, aunque inicialmente definidas en el \'area de programaci\'on inductiva, pueden ser usadas en una gama m\'as amplia de aplicaciones. En este trabajo, por ejemplo, se utilizan diferentes tipos de transformaciones que buscan adaptar datos planos en diferentes grados de estructuras, aplicando las distancias entre t\'erminos. De igual manera, se utiliza la distancia entre t\'erminos sobre un conjunto de datos originalmente jer\'arquicos. Todas estas transformaciones son aplicadas sobre estructuras XML; de esta manera, cualquier conjunto de datos, plano o con alg\'un nivel de jerarqu\'ia, puede ser transformado y, por medio de un proceso de clasificaci\'on (basado en distancias), utilizar las distancias entre t\'erminos.

\section{Discusi\'on}

Con los resultados obtenidos en los experimentos realizados utilizando las diferentes transformaciones y las distancias entre t\'erminos (y compar\'andolas con la distancia eucl\'idea, para el caso de conjuntos de datos planos), se puede identificar los siguientes tres aspectos:

\begin{itemize}
\item Los m\'etodos para la construcci\'on de jerarqu\'ias a partir de los nombres y los valores de atributos no siempre parecen dar buenos resultados. Sin embargo, estas asociaciones intuitivas entre atributos pueden modificar los resultados del proceso de clasificaci\'on y, en algunas ocasiones, de una manera sencilla y utilizando distancias entre t\'erminos, mejora los resultados obtenidos en la aplicaci\'on directa de este proceso sobre datos planos.

\item El proceso basado en la similitud de atributos se caracteriza por agrupar las variables m\'as parecidas entre s\'i y colocar m\'as cerca de la ra\'iz del \'arbol aquellos grupos que est\'an altamente relacionados; su principal beneficio se debe a la ponderaci\'on impl\'icita que asumen las distancias entre t\'erminos donde, com\'unmente, ofrecen mayor relevancia a los atributos que est\'an cerca de la ra\'iz del \'arbol. Sin embargo, la creaci\'on de estructuras altamente jerarquizadas puede generar que atributos con relevancia media y baja no sean considerados adecuadamente, debido a que su ponderaci\'on se hace muy peque\~na; esta raz\'on lleva a la construcci\'on de una estructura con menos niveles de jerarqu\'ia. El uso de este proceso, basado en la similitud entre atributos para construir un dendrograma y a partir de este construir una jerarqu\'ia, mejora los resultados obtenidos sobre la aplicaci\'on de datos planos. 

\item Cuando los datos son originalmente jer\'arquicos no es posible aplicar directamente las distancias definidas para datos proposicionales y es relevante considerar las ventajas ofrecidas por las propiedades de distancias entre t\'erminos con el fin de mejorar los resultados de la clasificaci\'on.
\end{itemize}

Este trabajo ha mostrado una aplicaci\'on prometedora de las distancias entre t\'erminos para diferentes tipos de conjuntos de datos, lo que sugiere que su uso puede ser m\'as amplio de lo que es ahora. Aunque existen distancias entre \'arboles que permiten la comparaci\'on de estructuras jer\'arquicas, como por ejemplo la distancia de edici\'on, estas se basan principalmente en costos de operaciones de inserci\'on y eliminaci\'on nodos y aristas, que dependen del n\'umero de transformaciones para convertir un \'arbol en otro; lo cual es un concepto que, en muchas ocasiones,  pierde el sentido de comparaci\'on jer\'arquica, puesto que no considera el nivel de profundidad de las variables dentro de este tipo de estructuras.

\section{Trabajos futuros}

Existen algunos de aspectos que ser\'ia importante explorar y profundizar en trabajos futuros  y que, seguramente, permitir\'ian mejorar los procesos de aprendizaje por medio de la aplicaci\'on de distancias entre t\'erminos. A continuaci\'on se describen algunos de estos aspectos:

\begin{itemize}
\item \textbf{Manejo de conjuntos:} este trabajo se centra principalmente en la aplicaci\'on distancias sobre estructuras jer\'arquicas que son representadas por medio de XML. Sin embargo, el manejo de conjuntos, como tipos de datos anidados dentro de este tipo de estructuras jer\'arquicas, es un aspecto relevante para los procesos de clasificaci\'on. La utilizaci\'on de m\'etricas sobre este tipo de dato, como por ejemplo distancias que tienen en cuenta la diferencia entre los elementos del conjunto, mejorar\'ia los procesos de clasificaci\'on.  

\item \textbf{Manejo de valores desconocidos de las variables y ausencia de variables:} el desconocimiento del valor de una variable y la ausencia de \'esta son aspectos que deben ser tratados de diferente manera; por ejemplo,  intuitivamente la ausencia de una misma caracter\'istica, entre dos objetos diferentes, representa mayor grado de similitud que el caso en que uno de estos objetos contiene la caracter\'istica y el otro no; por otra parte, el desconocimiento del valor de una misma variable en dos instancias  tiene una alta probabilidad de que efectivamente sean diferentes (esta probabilidad se incrementa, mientras mayor sea el dominio de los posibles valores de la variable en comparaci\'on);  igualmente, pueden generarse m\'ultiples combinaciones entre estos dos aspectos que seguramente, utilizando ponderaciones adecuadas, ayudar\'ian a mejorar los procesos de aprendizaje por medio de la utilizaci\'on de las distancias entre t\'erminos.

\item \textbf{Distancias entre t\'erminos para agrupamiento:} adicionalmente, un trabajo futuro es la aplicaci\'on de distancias entre t\'erminos para  tareas aprendizaje no supervisado; particularmente,  tareas de agrupamiento basado en distancias.

\end{itemize}

\section{Publicaciones}

Fundamentado en el an\'alisis y los resultados obtenidos de este proyecto de investigaci\'on, se desarroll\'o un art\'iculo titulado {\em ``Applying distances between terms to both flat and hierarchical data''} \cite{bedoya2011applying} en conjunto con los profesores Jos\'e Hern\'andez Orallo, C\'esar Ferri y Mar\'ia Jos\'e Ram\'irez Quintana, miembros del Grupo DMIP ({\em Data Mining and Inductive Programming}). Este art\'iculo fue aceptado por el comit\'e del {\em 4th International Workshop on Approaches and Applications of Inductive Programming},  presentado el 19 de julio de 2011 en Odense-Dinamarca (publicado en http://www.cogsys.wiai.uni-bamberg.de/aaip11/AAIP2011.pdf).  

\newpage
\cleardoublepage
\addcontentsline{toc}{chapter}{Bibliograf\'ia}
{
\bibliographystyle{plain}
\bibliography{biblio}
}
\appendix 
\addappheadtotoc 
\appendixpage

\chapter{Detalle de los resultados}\label{Resuldatos}

Este ap\'endice detalla los resultados presentados en el cap\'itulo \ref{experimentos} para los conjuntos de datos planos ({\em mushroom} y {\em soybean}) y originalmente jer\'arquicos ({\em demospongiae}), utilizando el algoritmo de clasificaci\'on $k$-NN con diferentes valores para el {\em par\'ametro de atracci\'on $i$}, aplicando las distancias entre t\'erminos y la distancia eucl\'idea. Para los conjuntos de datos planos se utiliza validaci\'on cruzada con 10 pliegues; para el conjunto de datos originalmente jer\'arquico se utiliza 60\% de ejemplos de entrenamiento y 40\% para ejemplos de prueba.

\section{Conjunto de datos {\em Mushroom}}

Los resultados para este conjunto de datos son calculados utilizando la distancia de Nienhuys-Cheng, la distancia de Estruch et al. y la distancia eucl\'idea sobre los tres tipos de estructuras: 1) plana (sin jerarqu\'ias), 2) jerarqu\'ia inducida por nombres y valores de los atributos y 3) jerarqu\'ia de acuerdo con la similitud de los atributos.

\begin{table}[H]\small
\begin{center}
\begin{tabular}{|c|r|r|r|r|r|r|r|r|r|}
\multicolumn{10}{l}{\normalsize{\textbf{Estructura plana (sin jerarqu\'ias):}}}\\
\multicolumn{10}{l}{}\\
  \hline
\multicolumn{10}{|c|}{Distancia de Nienhuys-Cheng}\\
  \hline
No. & Elementos &\multicolumn{2}{|c|}{$i=0$}&\multicolumn{2}{|c|}{$i=1$}&\multicolumn{2}{|c|}{$i=2$}& \multicolumn{2}{|c|}{$i=3$}\\ 
\cline{3-10}
pliegue & del pliegue & Aciertos &  \makebox[0.8cm][c]{\%}  & Aciertos &  \makebox[0.8cm][c]{\%} & Aciertos &  \makebox[0.8cm][c]{\%} & Aciertos &  \makebox[0.8cm][c]{\%} \\ 
 \hline\hline
0 & 100 & 100 & 100.00 & 100 & 100.00 & 100 & 100.00 & 99 & 99.00 \\ \hline
1 & 100 & 58 & 58.00 & 60 & 60.00 & 60 & 60.00 & 62 & 62.00 \\ \hline
2 & 100 & 94 & 94.00 & 96 & 96.00 & 99 & 99.00 & 100 & 100.00 \\ \hline
3 & 100 & 100 & 100.00 & 100 & 100.00 & 100 & 100.00 & 100 & 100.00 \\ \hline
4 & 100 & 100 & 100.00 & 100 & 100.00 & 100 & 100.00 & 100 & 100.00 \\ \hline
5 & 100 & 93 & 93.00 & 100 & 100.00 & 100 & 100.00 & 100 & 100.00 \\ \hline
6 & 100 & 100 & 100.00 & 100 & 100.00 & 100 & 100.00 & 100 & 100.00 \\ \hline
7 & 100 & 87 & 87.00 & 95 & 95.00 & 100 & 100.00 & 100 & 100.00 \\ \hline
8 & 100 & 98 & 98.00 & 98 & 98.00 & 98 & 98.00 & 98 & 98.00 \\ \hline
9 & 100 & 100 & 100.00 & 100 & 100.00 & 99 & 99.00 & 99 & 99.00 \\ \hline \hline
Total/prom. & 1000 & 930 & 93.00 & 949 & 94.90 & 956 & 95.60 & 958 & 95.80 \\ \hline
\end{tabular}
\vspace{-0.2cm}
\caption{Aciertos para el conjunto de datos {\em mushroom} sin jerarqu\'ias utilizando la distancia de Nienhuys-Cheng}
\label{tab-mush-sj-dn}
\end{center}
\end{table}

\begin{table}[H]\small
\begin{center}
\begin{tabular}{|c|r|r|r|r|r|r|r|r|r|}
  \hline
\multicolumn{10}{|c|}{Distancia de Estruch et al.}\\
  \hline
No. & Elementos &\multicolumn{2}{|c|}{$i=0$}&\multicolumn{2}{|c|}{$i=1$}&\multicolumn{2}{|c|}{$i=2$}& \multicolumn{2}{|c|}{$i=3$}\\ 
\cline{3-10}
pliegue & del pliegue & Aciertos &  \makebox[0.8cm][c]{\%}  & Aciertos &  \makebox[0.8cm][c]{\%} & Aciertos &  \makebox[0.8cm][c]{\%} & Aciertos &  \makebox[0.8cm][c]{\%} \\ 
 \hline\hline
0 & 100 & 100 & 100.00 & 100 & 100.00 & 100 & 100.00 & 99 & 99.00 \\ \hline
1 & 100 & 58 & 58.00 & 60 & 60.00 & 60 & 60.00 & 62 & 62.00 \\ \hline
2 & 100 & 94 & 94.00 & 96 & 96.00 & 99 & 99.00 & 100 & 100.00 \\ \hline
3 & 100 & 100 & 100.00 & 100 & 100.00 & 100 & 100.00 & 100 & 100.00 \\ \hline
4 & 100 & 100 & 100.00 & 100 & 100.00 & 100 & 100.00 & 100 & 100.00 \\ \hline
5 & 100 & 93 & 93.00 & 100 & 100.00 & 100 & 100.00 & 100 & 100.00 \\ \hline
6 & 100 & 100 & 100.00 & 100 & 100.00 & 100 & 100.00 & 100 & 100.00 \\ \hline
7 & 100 & 87 & 87.00 & 95 & 95.00 & 100 & 100.00 & 100 & 100.00 \\ \hline
8 & 100 & 98 & 98.00 & 98 & 98.00 & 98 & 98.00 & 98 & 98.00 \\ \hline
9 & 100 & 100 & 100.00 & 100 & 100.00 & 99 & 99.00 & 99 & 99.00 \\ \hline\hline
Total/prom. & 1000 & 930 & 93.00 & 949 & 94.90 & 956 & 95.60 & 958 & 95.80 \\ \hline
\end{tabular}
\vspace{-0.2cm}
\caption{Aciertos para el conjunto de datos {\em mushroom} sin jerarqu\'ias utilizando la distancia de Estruch et al.}
\label{tab-mush-sj-de}
\end{center}
\end{table}

\begin{table}[H]\small
\begin{center}
\begin{tabular}{|c|r|r|r|r|r|r|r|r|r|}
  \hline
\multicolumn{10}{|c|}{Distancia eucl\'idea}\\
  \hline
No. & Elementos &\multicolumn{2}{|c|}{$i=0$}&\multicolumn{2}{|c|}{$i=1$}&\multicolumn{2}{|c|}{$i=2$}& \multicolumn{2}{|c|}{$i=3$}\\ 
\cline{3-10}
pliegue & del pliegue & Aciertos &  \makebox[0.8cm][c]{\%}  & Aciertos &  \makebox[0.8cm][c]{\%} & Aciertos &  \makebox[0.8cm][c]{\%} & Aciertos &  \makebox[0.8cm][c]{\%} \\ 
 \hline\hline
0 & 100 & 100 & 100.00 & 100 & 100.00 & 100 & 100.00 & 100 & 100.00 \\ \hline
1 & 100 & 58 & 58.00 & 60 & 60.00 & 60 & 60.00 & 60 & 60.00 \\ \hline
2 & 100 & 94 & 94.00 & 94 & 94.00 & 96 & 96.00 & 98 & 98.00 \\ \hline
3 & 100 & 100 & 100.00 & 100 & 100.00 & 100 & 100.00 & 100 & 100.00 \\ \hline
4 & 100 & 100 & 100.00 & 100 & 100.00 & 100 & 100.00 & 100 & 100.00 \\ \hline
5 & 100 & 93 & 93.00 & 98 & 98.00 & 100 & 100.00 & 100 & 100.00 \\ \hline
6 & 100 & 100 & 100.00 & 100 & 100.00 & 100 & 100.00 & 100 & 100.00 \\ \hline
7 & 100 & 87 & 87.00 & 94 & 94.00 & 95 & 95.00 & 99 & 99.00 \\ \hline
8 & 100 & 98 & 98.00 & 98 & 98.00 & 98 & 98.00 & 98 & 98.00 \\ \hline
9 & 100 & 100 & 100.00 & 100 & 100.00 & 100 & 100.00 & 100 & 100.00 \\ \hline\hline
Total/prom. & 1000 & 930 & 93.00 & 944 & 94.40 & 949 & 94.90 & 955 & 95.50 \\ \hline
\end{tabular}
\vspace{-0.2cm}
\caption{Aciertos para el conjunto de datos {\em mushroom} sin jerarqu\'ias utilizando la distancia eucl\'idea}
\label{tab-mush-sj-du}
\end{center}
\end{table}

\begin{table}[H]\small
\begin{center}
\begin{tabular}{|c|r|r|r|r|r|r|r|r|r|}
\multicolumn{10}{l}{\normalsize{\textbf{Jerarqu\'ia inducida por nombres y valores de los atributos:}}}\\
\multicolumn{10}{l}{}\\
  \hline
\multicolumn{10}{|c|}{Distancia de Nienhuys-Cheng}\\
  \hline
No. & Elementos &\multicolumn{2}{|c|}{$i=0$}&\multicolumn{2}{|c|}{$i=1$}&\multicolumn{2}{|c|}{$i=2$}& \multicolumn{2}{|c|}{$i=3$}\\ 
\cline{3-10}
pliegue & del pliegue & Aciertos &  \makebox[0.8cm][c]{\%}  & Aciertos &  \makebox[0.8cm][c]{\%} & Aciertos &  \makebox[0.8cm][c]{\%} & Aciertos &  \makebox[0.8cm][c]{\%} \\ 
 \hline\hline
0 & 100 & 100 & 100.00 & 100 & 100.00 & 100 & 100.00 & 100 & 100.00 \\ \hline
1 & 100 & 100 & 100.00 & 100 & 100.00 & 100 & 100.00 & 100 & 100.00 \\ \hline
2 & 100 & 99 & 99.00 & 99 & 99.00 & 100 & 100.00 & 100 & 100.00 \\ \hline
3 & 100 & 99 & 99.00 & 99 & 99.00 & 99 & 99.00 & 99 & 99.00 \\ \hline
4 & 100 & 100 & 100.00 & 100 & 100.00 & 100 & 100.00 & 100 & 100.00 \\ \hline
5 & 100 & 100 & 100.00 & 100 & 100.00 & 100 & 100.00 & 100 & 100.00 \\ \hline
6 & 100 & 100 & 100.00 & 100 & 100.00 & 100 & 100.00 & 100 & 100.00 \\ \hline
7 & 100 & 100 & 100.00 & 100 & 100.00 & 100 & 100.00 & 100 & 100.00 \\ \hline
8 & 100 & 100 & 100.00 & 100 & 100.00 & 100 & 100.00 & 100 & 100.00 \\ \hline
9 & 100 & 100 & 100.00 & 100 & 100.00 & 100 & 100.00 & 100 & 100.00 \\ \hline \hline
Total/prom. & 1000 & 998 & 99.80 & 998 & 99.80 & 999 & 99.90 & 999 & 99.90 \\ \hline
\end{tabular}
\vspace{-0.2cm}
\caption{Aciertos para el conjunto de datos {\em mushroom} con jerarqu\'ia inducida por nombres y valores de los atributos utilizando la distancia de Nienhuys-Cheng}
\label{tab-mush-jn-dn}
\end{center}
\end{table}

\begin{table}[H]\small
\begin{center}
\begin{tabular}{|c|r|r|r|r|r|r|r|r|r|}
  \hline
\multicolumn{10}{|c|}{Distancia de Estruch et al.}\\
  \hline
No. & Elementos &\multicolumn{2}{|c|}{$i=0$}&\multicolumn{2}{|c|}{$i=1$}&\multicolumn{2}{|c|}{$i=2$}& \multicolumn{2}{|c|}{$i=3$}\\ 
\cline{3-10}
pliegue & del pliegue & Aciertos &  \makebox[0.8cm][c]{\%}  & Aciertos &  \makebox[0.8cm][c]{\%} & Aciertos &  \makebox[0.8cm][c]{\%} & Aciertos &  \makebox[0.8cm][c]{\%} \\ 
 \hline\hline
0 & 100 & 100 & 100.00 & 100 & 100.00 & 100 & 100.00 & 100 & 100.00 \\ \hline
1 & 100 & 100 & 100.00 & 100 & 100.00 & 100 & 100.00 & 100 & 100.00 \\ \hline
2 & 100 & 99 & 99.00 & 99 & 99.00 & 100 & 100.00 & 100 & 100.00 \\ \hline
3 & 100 & 99 & 99.00 & 99 & 99.00 & 98 & 98.00 & 98 & 98.00 \\ \hline
4 & 100 & 100 & 100.00 & 100 & 100.00 & 100 & 100.00 & 100 & 100.00 \\ \hline
5 & 100 & 100 & 100.00 & 100 & 100.00 & 100 & 100.00 & 100 & 100.00 \\ \hline
6 & 100 & 100 & 100.00 & 100 & 100.00 & 100 & 100.00 & 100 & 100.00 \\ \hline
7 & 100 & 100 & 100.00 & 100 & 100.00 & 100 & 100.00 & 100 & 100.00 \\ \hline
8 & 100 & 100 & 100.00 & 100 & 100.00 & 100 & 100.00 & 100 & 100.00 \\ \hline
9 & 100 & 100 & 100.00 & 100 & 100.00 & 100 & 100.00 & 100 & 100.00 \\ \hline \hline
Total/prom. & 1000 & 998 & 99.80 & 998 & 99.80 & 998 & 99.80 & 998 & 99.80 \\ \hline
\end{tabular}
\vspace{-0.2cm}
\caption{Aciertos  para el conjunto de datos {\em mushroom} con jerarqu\'ia inducida por nombres y valores de los atributos utilizando la distancia de Estruch et al.}
\label{tab-mush-jn-de}
\end{center}
\end{table}

\begin{table}[H]\small
\begin{center}
\begin{tabular}{|c|r|r|r|r|r|r|r|r|r|}
  \hline
\multicolumn{10}{|c|}{Distancia eucl\'idea}\\
  \hline
No. & Elementos &\multicolumn{2}{|c|}{$i=0$}&\multicolumn{2}{|c|}{$i=1$}&\multicolumn{2}{|c|}{$i=2$}& \multicolumn{2}{|c|}{$i=3$}\\ 
\cline{3-10}
pliegue & del pliegue & Aciertos &  \makebox[0.8cm][c]{\%}  & Aciertos &  \makebox[0.8cm][c]{\%} & Aciertos &  \makebox[0.8cm][c]{\%} & Aciertos &  \makebox[0.8cm][c]{\%} \\ 
 \hline\hline
0 & 100 & 100 & 100.00 & 100 & 100.00 & 100 & 100.00 & 100 & 100.00 \\ \hline
1 & 100 & 58 & 58.00 & 60 & 60.00 & 60 & 60.00 & 60 & 60.00 \\ \hline
2 & 100 & 94 & 94.00 & 94 & 94.00 & 96 & 96.00 & 98 & 98.00 \\ \hline
3 & 100 & 100 & 100.00 & 100 & 100.00 & 100 & 100.00 & 100 & 100.00 \\ \hline
4 & 100 & 100 & 100.00 & 100 & 100.00 & 100 & 100.00 & 100 & 100.00 \\ \hline
5 & 100 & 93 & 93.00 & 98 & 98.00 & 100 & 100.00 & 100 & 100.00 \\ \hline
6 & 100 & 100 & 100.00 & 100 & 100.00 & 100 & 100.00 & 100 & 100.00 \\ \hline
7 & 100 & 87 & 87.00 & 94 & 94.00 & 95 & 95.00 & 99 & 99.00 \\ \hline
8 & 100 & 98 & 98.00 & 98 & 98.00 & 98 & 98.00 & 98 & 98.00 \\ \hline
9 & 100 & 100 & 100.00 & 100 & 100.00 & 100 & 100.00 & 100 & 100.00 \\ \hline \hline
Total/prom. & 1000 & 930 & 93.00 & 944 & 94.40 & 949 & 94.90 & 955 & 95.50 \\ \hline
\end{tabular}
\vspace{-0.2cm}
\caption{Aciertos para el conjunto de datos {\em mushroom} con jerarqu\'ia inducida por nombres y valores de los atributos utilizando la distancia eucl\'idea}
\label{tab-mush-jn-du}
\end{center}
\end{table}

\begin{table}[H]\small
\begin{center}
\begin{tabular}{|c|r|r|r|r|r|r|r|r|r|}
\multicolumn{10}{l}{\normalsize{\textbf{Jerarqu\'ia de similitud de los atributos:}}}\\
\multicolumn{10}{l}{}\\
  \hline
\multicolumn{10}{|c|}{Distancia de Nienhuys-Cheng}\\
  \hline
No. & Elementos &\multicolumn{2}{|c|}{$i=0$}&\multicolumn{2}{|c|}{$i=1$}&\multicolumn{2}{|c|}{$i=2$}& \multicolumn{2}{|c|}{$i=3$}\\ 
\cline{3-10}
pliegue & del pliegue & Aciertos &  \makebox[0.8cm][c]{\%}  & Aciertos &  \makebox[0.8cm][c]{\%} & Aciertos &  \makebox[0.8cm][c]{\%} & Aciertos &  \makebox[0.8cm][c]{\%} \\ 
 \hline\hline
0 & 100 & 100 & 100.00 & 100 & 100.00 & 100 & 100.00 & 100 & 100.00 \\ \hline
1 & 100 & 99 & 99.00 & 100 & 100.00 & 100 & 100.00 & 100 & 100.00 \\ \hline
2 & 100 & 98 & 98.00 & 98 & 98.00 & 100 & 100.00 & 100 & 100.00 \\ \hline
3 & 100 & 98 & 98.00 & 100 & 100.00 & 100 & 100.00 & 100 & 100.00 \\ \hline
4 & 100 & 100 & 100.00 & 100 & 100.00 & 100 & 100.00 & 100 & 100.00 \\ \hline
5 & 100 & 100 & 100.00 & 100 & 100.00 & 100 & 100.00 & 100 & 100.00 \\ \hline
6 & 100 & 100 & 100.00 & 100 & 100.00 & 100 & 100.00 & 100 & 100.00 \\ \hline
7 & 100 & 100 & 100.00 & 100 & 100.00 & 100 & 100.00 & 100 & 100.00 \\ \hline
8 & 100 & 100 & 100.00 & 100 & 100.00 & 100 & 100.00 & 100 & 100.00 \\ \hline
9 & 100 & 100 & 100.00 & 100 & 100.00 & 100 & 100.00 & 100 & 100.00 \\ \hline \hline
Total/prom. & 1000 & 995 & 99.50 & 998 & 99.80 & 1000 & 100.00 & 1000 & 100.00 \\ \hline
\end{tabular}
\vspace{-0.2cm}
\caption{Aciertos para el conjunto de datos {\em mushroom} con jerarqu\'ia de similitud de los atributos utilizando la distancia de Nienhuys-Cheng}
\label{tab-mush-jsim-dn}
\end{center}
\end{table}

\begin{table}[H]\small
\begin{center}
\begin{tabular}{|c|r|r|r|r|r|r|r|r|r|}
  \hline
\multicolumn{10}{|c|}{Distancia de Estruch et al.}\\
  \hline
No. & Elementos &\multicolumn{2}{|c|}{$i=0$}&\multicolumn{2}{|c|}{$i=1$}&\multicolumn{2}{|c|}{$i=2$}& \multicolumn{2}{|c|}{$i=3$}\\ 
\cline{3-10}
pliegue & del pliegue & Aciertos &  \makebox[0.8cm][c]{\%}  & Aciertos &  \makebox[0.8cm][c]{\%} & Aciertos &  \makebox[0.8cm][c]{\%} & Aciertos &  \makebox[0.8cm][c]{\%} \\ 
 \hline\hline
0 & 100 & 100 & 100.00 & 100 & 100.00 & 100 & 100.00 & 100 & 100.00 \\ \hline
1 & 100 & 99 & 99.00 & 100 & 100.00 & 100 & 100.00 & 100 & 100.00 \\ \hline
2 & 100 & 97 & 97.00 & 100 & 100.00 & 100 & 100.00 & 100 & 100.00 \\ \hline
3 & 100 & 99 & 99.00 & 100 & 100.00 & 100 & 100.00 & 100 & 100.00 \\ \hline
4 & 100 & 100 & 100.00 & 100 & 100.00 & 100 & 100.00 & 100 & 100.00 \\ \hline
5 & 100 & 100 & 100.00 & 100 & 100.00 & 100 & 100.00 & 100 & 100.00 \\ \hline
6 & 100 & 100 & 100.00 & 100 & 100.00 & 100 & 100.00 & 100 & 100.00 \\ \hline
7 & 100 & 100 & 100.00 & 100 & 100.00 & 100 & 100.00 & 100 & 100.00 \\ \hline
8 & 100 & 100 & 100.00 & 100 & 100.00 & 100 & 100.00 & 100 & 100.00 \\ \hline
9 & 100 & 100 & 100.00 & 100 & 100.00 & 100 & 100.00 & 100 & 100.00 \\ \hline \hline
Total/prom. & 1000 & 995 & 99.50 & 1000 & 100.00 & 1000 & 100.00 & 1000 & 100.00 \\ \hline
\end{tabular}
\vspace{-0.2cm}
\caption{Aciertos  para el conjunto de datos {\em mushroom} con jerarqu\'ia de similitud de los atributos utilizando la distancia de Estruch et al.}
\label{tab-mush-jsim-de}
\end{center}
\end{table}

\begin{table}[H]\small
\begin{center}
\begin{tabular}{|c|r|r|r|r|r|r|r|r|r|}
  \hline
\multicolumn{10}{|c|}{Distancia eucl\'idea}\\
  \hline
No. & Elementos &\multicolumn{2}{|c|}{$i=0$}&\multicolumn{2}{|c|}{$i=1$}&\multicolumn{2}{|c|}{$i=2$}& \multicolumn{2}{|c|}{$i=3$}\\ 
\cline{3-10}
pliegue & del pliegue & Aciertos &  \makebox[0.8cm][c]{\%}  & Aciertos &  \makebox[0.8cm][c]{\%} & Aciertos &  \makebox[0.8cm][c]{\%} & Aciertos &  \makebox[0.8cm][c]{\%} \\ 
 \hline\hline
0 & 100 & 100 & 100.00 & 100 & 100.00 & 100 & 100.00 & 100 & 100.00 \\ \hline
1 & 100 & 58 & 58.00 & 60 & 60.00 & 60 & 60.00 & 60 & 60.00 \\ \hline
2 & 100 & 94 & 94.00 & 94 & 94.00 & 96 & 96.00 & 98 & 98.00 \\ \hline
3 & 100 & 100 & 100.00 & 100 & 100.00 & 100 & 100.00 & 100 & 100.00 \\ \hline
4 & 100 & 100 & 100.00 & 100 & 100.00 & 100 & 100.00 & 100 & 100.00 \\ \hline
5 & 100 & 93 & 93.00 & 98 & 98.00 & 100 & 100.00 & 100 & 100.00 \\ \hline
6 & 100 & 100 & 100.00 & 100 & 100.00 & 100 & 100.00 & 100 & 100.00 \\ \hline
7 & 100 & 87 & 87.00 & 94 & 94.00 & 95 & 95.00 & 99 & 99.00 \\ \hline
8 & 100 & 98 & 98.00 & 98 & 98.00 & 98 & 98.00 & 98 & 98.00 \\ \hline
9 & 100 & 100 & 100.00 & 100 & 100.00 & 100 & 100.00 & 100 & 100.00 \\ \hline \hline
Total/prom. & 1000 & 930 & 93.00 & 944 & 94.40 & 949 & 94.90 & 955 & 95.50 \\ \hline
\end{tabular}
\vspace{-0.2cm}
\caption{Aciertos para el conjunto de datos {\em mushroom} con jerarqu\'ia de similitud de los atributos utilizando la distancia eucl\'idea}
\label{tab-mush-jsim-du}
\end{center}
\end{table}

\section{Conjunto de datos {\em Soybean}}

Los resultados del conjunto de datos {\em soybean}, al igual que {\em mushroom}, son calculados utilizando las distancias de Nienhuys-Cheng, la distancia de Estruch et al. y la distancia eucl\'idea sobre sobre una estructura plana (sin jerarqu\'ias), una jerarqu\'ia inducida por nombres (y valores de los atributos) y una jerarqu\'ia de acuerdo con la similitud de los atributos.

\begin{table}[H]\small
\begin{center}
\begin{tabular}{|c|r|r|r|r|r|r|r|r|r|}
\multicolumn{10}{l}{\normalsize{\textbf{Estructura plana (sin jerarqu\'ias):}}}\\
\multicolumn{10}{l}{}\\
  \hline
\multicolumn{10}{|c|}{Distancia de Nienhuys-Cheng}\\
  \hline
No. & Elementos &\multicolumn{2}{|c|}{$i=0$}&\multicolumn{2}{|c|}{$i=1$}&\multicolumn{2}{|c|}{$i=2$}& \multicolumn{2}{|c|}{$i=3$}\\ 
\cline{3-10}
pliegue & del pliegue & Aciertos &  \makebox[0.8cm][c]{\%}  & Aciertos &  \makebox[0.8cm][c]{\%} & Aciertos &  \makebox[0.8cm][c]{\%} & Aciertos &  \makebox[0.8cm][c]{\%} \\ 
 \hline\hline
0 & 31 & 25 & 80.65 & 30 & 96.77 & 31 & 100.00 & 31 & 100.00 \\ \hline
1 & 31 & 22 & 70.97 & 29 & 93.55 & 29 & 93.55 & 29 & 93.55 \\ \hline
2 & 31 & 19 & 61.29 & 26 & 83.87 & 28 & 90.32 & 29 & 93.55 \\ \hline
3 & 31 & 19 & 61.29 & 29 & 93.55 & 29 & 93.55 & 29 & 93.55 \\ \hline
4 & 31 & 21 & 67.74 & 25 & 80.65 & 26 & 83.87 & 29 & 93.55 \\ \hline
5 & 31 & 24 & 77.42 & 26 & 83.87 & 29 & 93.55 & 30 & 96.77 \\ \hline
6 & 31 & 28 & 90.32 & 30 & 96.77 & 30 & 96.77 & 31 & 100.00 \\ \hline
7 & 31 & 25 & 80.65 & 30 & 96.77 & 30 & 96.77 & 30 & 96.77 \\ \hline
8 & 31 & 24 & 77.42 & 27 & 87.10 & 28 & 90.32 & 28 & 90.32 \\ \hline
9 & 28 & 24 & 85.71 & 28 & 100.00 & 28 & 100.00 & 28 & 100.00 \\ \hline \hline
Total/prom. & 307 & 231 & 75.35 & 280 & 91.29 & 288 & 93.87 & 294 & 95.81 \\ \hline
\end{tabular}
\vspace{-0.2cm}
\caption{Aciertos para el conjunto de datos {\em soybean} sin jerarqu\'ias utilizando la distancia de Nienhuys-Cheng}
\label{tab-soyb-sj-dn}
\end{center}
\end{table}

\begin{table}[H]\small
\begin{center}
\begin{tabular}{|c|r|r|r|r|r|r|r|r|r|}
  \hline
\multicolumn{10}{|c|}{Distancia de Estruch et al.}\\
  \hline
No. & Elementos &\multicolumn{2}{|c|}{$i=0$}&\multicolumn{2}{|c|}{$i=1$}&\multicolumn{2}{|c|}{$i=2$}& \multicolumn{2}{|c|}{$i=3$}\\ 
\cline{3-10}
pliegue & del pliegue & Aciertos &  \makebox[0.8cm][c]{\%}  & Aciertos &  \makebox[0.8cm][c]{\%} & Aciertos &  \makebox[0.8cm][c]{\%} & Aciertos &  \makebox[0.8cm][c]{\%} \\ 
 \hline\hline
0 & 31 & 25 & 80.65 & 30 & 96.77 & 31 & 100.00 & 31 & 100.00 \\ \hline
1 & 31 & 22 & 70.97 & 29 & 93.55 & 29 & 93.55 & 29 & 93.55 \\ \hline
2 & 31 & 19 & 61.29 & 26 & 83.87 & 28 & 90.32 & 29 & 93.55 \\ \hline
3 & 31 & 19 & 61.29 & 29 & 93.55 & 29 & 93.55 & 29 & 93.55 \\ \hline
4 & 31 & 21 & 67.74 & 25 & 80.65 & 26 & 83.87 & 29 & 93.55 \\ \hline
5 & 31 & 24 & 77.42 & 26 & 83.87 & 29 & 93.55 & 30 & 96.77 \\ \hline
6 & 31 & 28 & 90.32 & 30 & 96.77 & 30 & 96.77 & 31 & 100.00 \\ \hline
7 & 31 & 25 & 80.65 & 30 & 96.77 & 30 & 96.77 & 30 & 96.77 \\ \hline
8 & 31 & 24 & 77.42 & 27 & 87.10 & 28 & 90.32 & 28 & 90.32 \\ \hline
9 & 28 & 24 & 85.71 & 28 & 100.00 & 28 & 100.00 & 28 & 100.00 \\ \hline \hline
Total/prom. & 307 & 231 & 75.35 & 280 & 91.29 & 288 & 93.87 & 294 & 95.81 \\ \hline
\end{tabular}
\vspace{-0.2cm}
\caption{Aciertos para el conjunto de datos {\em soybean} sin jerarqu\'ias utilizando la distancia de Estruch et al.}
\label{tab-soyb-sj-de}
\end{center}
\end{table}

\begin{table}[H]\small
\begin{center}
\begin{tabular}{|c|r|r|r|r|r|r|r|r|r|}
  \hline
\multicolumn{10}{|c|}{Distancia eucl\'idea}\\
  \hline
No. & Elementos &\multicolumn{2}{|c|}{$i=0$}&\multicolumn{2}{|c|}{$i=1$}&\multicolumn{2}{|c|}{$i=2$}& \multicolumn{2}{|c|}{$i=3$}\\ 
\cline{3-10}
pliegue & del pliegue & Aciertos &  \makebox[0.8cm][c]{\%}  & Aciertos &  \makebox[0.8cm][c]{\%} & Aciertos &  \makebox[0.8cm][c]{\%} & Aciertos &  \makebox[0.8cm][c]{\%} \\ 
 \hline\hline
0 & 31 & 25 & 80.65 & 29 & 93.55 & 30 & 96.77 & 31 & 100.00 \\ \hline
1 & 31 & 22 & 70.97 & 25 & 80.65 & 29 & 93.55 & 29 & 93.55 \\ \hline
2 & 31 & 19 & 61.29 & 25 & 80.65 & 26 & 83.87 & 26 & 83.87 \\ \hline
3 & 31 & 19 & 61.29 & 28 & 90.32 & 29 & 93.55 & 29 & 93.55 \\ \hline
4 & 31 & 21 & 67.74 & 24 & 77.42 & 25 & 80.65 & 25 & 80.65 \\ \hline
5 & 31 & 24 & 77.42 & 26 & 83.87 & 26 & 83.87 & 28 & 90.32 \\ \hline
6 & 31 & 28 & 90.32 & 30 & 96.77 & 30 & 96.77 & 30 & 96.77 \\ \hline
7 & 31 & 25 & 80.65 & 26 & 83.87 & 30 & 96.77 & 30 & 96.77 \\ \hline
8 & 31 & 24 & 77.42 & 27 & 87.10 & 27 & 87.10 & 28 & 90.32 \\ \hline
9 & 28 & 24 & 85.71 & 27 & 96.43 & 28 & 100.00 & 28 & 100.00 \\ \hline \hline
Total/prom. & 307 & 231 & 75.35 & 267 & 87.06 & 280 & 91.29 & 284 & 92.58 \\ \hline
\end{tabular}
\vspace{-0.2cm}
\caption{Aciertos para el conjunto de datos {\em soybean} sin jerarqu\'ias utilizando la distancia eucl\'idea}
\label{tab-soyb-sj-du}
\end{center}
\end{table}

\begin{table}[H]\small
\begin{center}
\begin{tabular}{|c|r|r|r|r|r|r|r|r|r|}
\multicolumn{10}{l}{\normalsize{\textbf{Jerarqu\'ia inducida por nombres y valores de los atributos:}}}\\
\multicolumn{10}{l}{}\\
  \hline
\multicolumn{10}{|c|}{Distancia de Nienhuys-Cheng}\\
  \hline
No. & Elementos &\multicolumn{2}{|c|}{$i=0$}&\multicolumn{2}{|c|}{$i=1$}&\multicolumn{2}{|c|}{$i=2$}& \multicolumn{2}{|c|}{$i=3$}\\ 
\cline{3-10}
pliegue & del pliegue & Aciertos &  \makebox[0.8cm][c]{\%}  & Aciertos &  \makebox[0.8cm][c]{\%} & Aciertos &  \makebox[0.8cm][c]{\%} & Aciertos &  \makebox[0.8cm][c]{\%} \\ 
 \hline\hline
0 & 31 & 28 & 90.32 & 30 & 96.77 & 30 & 96.77 & 30 & 96.77 \\ \hline
1 & 31 & 23 & 74.19 & 28 & 90.32 & 29 & 93.55 & 29 & 93.55 \\ \hline
2 & 31 & 24 & 77.42 & 26 & 83.87 & 28 & 90.32 & 28 & 90.32 \\ \hline
3 & 31 & 23 & 74.19 & 27 & 87.10 & 28 & 90.32 & 28 & 90.32 \\ \hline
4 & 31 & 21 & 67.74 & 25 & 80.65 & 26 & 83.87 & 27 & 87.10 \\ \hline
5 & 31 & 24 & 77.42 & 27 & 87.10 & 27 & 87.10 & 27 & 87.10 \\ \hline
6 & 31 & 26 & 83.87 & 29 & 93.55 & 30 & 96.77 & 31 & 100.00 \\ \hline
7 & 31 & 24 & 77.42 & 30 & 96.77 & 30 & 96.77 & 30 & 96.77 \\ \hline
8 & 31 & 24 & 77.42 & 26 & 83.87 & 28 & 90.32 & 28 & 90.32 \\ \hline
9 & 28 & 24 & 85.71 & 27 & 96.43 & 28 & 100.00 & 28 & 100.00 \\ \hline \hline
Total/prom. & 307 & 241 & 78.57 & 275 & 89.64 & 284 & 92.58 & 286 & 93.23 \\ \hline
\end{tabular}
\vspace{-0.2cm}
\caption{Aciertos para el conjunto de datos {\em soybean} con jerarqu\'ia inducida por nombres y valores de los atributos utilizando la distancia de Nienhuys-Cheng}
\label{tab-soyb-jn-dn}
\end{center}
\end{table}

\begin{table}[H]\small
\begin{center}
\begin{tabular}{|c|r|r|r|r|r|r|r|r|r|}
  \hline
\multicolumn{10}{|c|}{Distancia de Estruch et al.}\\
  \hline
No. & Elementos &\multicolumn{2}{|c|}{$i=0$}&\multicolumn{2}{|c|}{$i=1$}&\multicolumn{2}{|c|}{$i=2$}& \multicolumn{2}{|c|}{$i=3$}\\ 
\cline{3-10}
pliegue & del pliegue & Aciertos &  \makebox[0.8cm][c]{\%}  & Aciertos &  \makebox[0.8cm][c]{\%} & Aciertos &  \makebox[0.8cm][c]{\%} & Aciertos &  \makebox[0.8cm][c]{\%} \\ 
 \hline\hline
0 & 31 & 27 & 87.10 & 29 & 93.55 & 30 & 96.77 & 31 & 100.00 \\ \hline
1 & 31 & 23 & 74.19 & 28 & 90.32 & 29 & 93.55 & 29 & 93.55 \\ \hline
2 & 31 & 22 & 70.97 & 27 & 87.10 & 28 & 90.32 & 28 & 90.32 \\ \hline
3 & 31 & 24 & 77.42 & 27 & 87.10 & 28 & 90.32 & 28 & 90.32 \\ \hline
4 & 31 & 21 & 67.74 & 25 & 80.65 & 25 & 80.65 & 27 & 87.10 \\ \hline
5 & 31 & 23 & 74.19 & 24 & 77.42 & 24 & 77.42 & 25 & 80.65 \\ \hline
6 & 31 & 25 & 80.65 & 28 & 90.32 & 31 & 100.00 & 31 & 100.00 \\ \hline
7 & 31 & 25 & 80.65 & 28 & 90.32 & 30 & 96.77 & 30 & 96.77 \\ \hline
8 & 31 & 20 & 64.52 & 27 & 87.10 & 28 & 90.32 & 30 & 96.77 \\ \hline
9 & 28 & 23 & 82.14 & 27 & 96.43 & 28 & 100.00 & 28 & 100.00 \\ \hline \hline
Total/prom. & 307 & 233 & 75.96 & 270 & 88.03 & 281 & 91.61 & 287 & 93.55 \\ \hline
\end{tabular}
\vspace{-0.2cm}
\caption{Aciertos  para el conjunto de datos {\em soybean} con jerarqu\'ia inducida por nombres y valores de los atributos utilizando la distancia de Estruch et al.}
\label{tab-soyb-jn-de}
\end{center}
\end{table}

\begin{table}[H]\small
\begin{center}
\begin{tabular}{|c|r|r|r|r|r|r|r|r|r|}
  \hline
\multicolumn{10}{|c|}{Distancia eucl\'idea}\\
  \hline
No. & Elementos &\multicolumn{2}{|c|}{$i=0$}&\multicolumn{2}{|c|}{$i=1$}&\multicolumn{2}{|c|}{$i=2$}& \multicolumn{2}{|c|}{$i=3$}\\ 
\cline{3-10}
pliegue & del pliegue & Aciertos &  \makebox[0.8cm][c]{\%}  & Aciertos &  \makebox[0.8cm][c]{\%} & Aciertos &  \makebox[0.8cm][c]{\%} & Aciertos &  \makebox[0.8cm][c]{\%} \\ 
 \hline\hline
0 & 31 & 25 & 80.65 & 29 & 93.55 & 30 & 96.77 & 31 & 100.00 \\ \hline
1 & 31 & 22 & 70.97 & 25 & 80.65 & 29 & 93.55 & 29 & 93.55 \\ \hline
2 & 31 & 19 & 61.29 & 25 & 80.65 & 26 & 83.87 & 26 & 83.87 \\ \hline
3 & 31 & 19 & 61.29 & 28 & 90.32 & 29 & 93.55 & 29 & 93.55 \\ \hline
4 & 31 & 21 & 67.74 & 24 & 77.42 & 25 & 80.65 & 25 & 80.65 \\ \hline
5 & 31 & 24 & 77.42 & 26 & 83.87 & 26 & 83.87 & 28 & 90.32 \\ \hline
6 & 31 & 28 & 90.32 & 30 & 96.77 & 30 & 96.77 & 30 & 96.77 \\ \hline
7 & 31 & 25 & 80.65 & 26 & 83.87 & 30 & 96.77 & 30 & 96.77 \\ \hline
8 & 31 & 24 & 77.42 & 27 & 87.10 & 27 & 87.10 & 28 & 90.32 \\ \hline
9 & 28 & 24 & 85.71 & 27 & 96.43 & 28 & 100.00 & 28 & 100.00 \\ \hline \hline
Total/prom. & 307 & 231 & 75.35 & 267 & 87.06 & 280 & 91.29 & 284 & 92.58 \\ \hline
\end{tabular}
\vspace{-0.2cm}
\caption{Aciertos para el conjunto de datos {\em soybean} con jerarqu\'ia inducida por nombres y valores de los atributos utilizando la distancia eucl\'idea}
\label{tab-soyb-jn-du}
\end{center}
\end{table}

\begin{table}[H]\small
\begin{center}
\begin{tabular}{|c|r|r|r|r|r|r|r|r|r|}
\multicolumn{10}{l}{\normalsize{\textbf{Jerarqu\'ia de similitud de los atributos:}}}\\
\multicolumn{10}{l}{}\\
  \hline
\multicolumn{10}{|c|}{Distancia de Nienhuys-Cheng}\\
  \hline
No. & Elementos &\multicolumn{2}{|c|}{$i=0$}&\multicolumn{2}{|c|}{$i=1$}&\multicolumn{2}{|c|}{$i=2$}& \multicolumn{2}{|c|}{$i=3$}\\ 
\cline{3-10}
pliegue & del pliegue & Aciertos &  \makebox[0.8cm][c]{\%}  & Aciertos &  \makebox[0.8cm][c]{\%} & Aciertos &  \makebox[0.8cm][c]{\%} & Aciertos &  \makebox[0.8cm][c]{\%} \\ 
 \hline\hline
0 & 31 & 26 & 83.87 & 31 & 100.00 & 31 & 100.00 & 31 & 100.00 \\ \hline
1 & 31 & 27 & 87.10 & 31 & 100.00 & 31 & 100.00 & 31 & 100.00 \\ \hline
2 & 31 & 25 & 80.65 & 30 & 96.77 & 30 & 96.77 & 30 & 96.77 \\ \hline
3 & 31 & 25 & 80.65 & 30 & 96.77 & 30 & 96.77 & 30 & 96.77 \\ \hline
4 & 31 & 28 & 90.32 & 30 & 96.77 & 31 & 100.00 & 31 & 100.00 \\ \hline
5 & 31 & 26 & 83.87 & 31 & 100.00 & 31 & 100.00 & 31 & 100.00 \\ \hline
6 & 31 & 28 & 90.32 & 31 & 100.00 & 31 & 100.00 & 31 & 100.00 \\ \hline
7 & 31 & 27 & 87.10 & 31 & 100.00 & 31 & 100.00 & 31 & 100.00 \\ \hline
8 & 31 & 25 & 80.65 & 31 & 100.00 & 31 & 100.00 & 31 & 100.00 \\ \hline
9 & 28 & 27 & 96.43 & 28 & 100.00 & 28 & 100.00 & 28 & 100.00 \\ \hline \hline
Total/prom. & 307 & 264 & 86.09 & 304 & 99.03 & 305 & 99.35 & 305 & 99.35 \\ \hline
\end{tabular}
\vspace{-0.2cm}
\caption{Aciertos para el conjunto de datos {\em soybean} con jerarqu\'ia de similitud de los atributos utilizando la distancia de Nienhuys-Cheng}
\label{tab-soyb-jsim-dn}
\end{center}
\end{table}

\begin{table}[H]\small
\begin{center}
\begin{tabular}{|c|r|r|r|r|r|r|r|r|r|}
  \hline
\multicolumn{10}{|c|}{Distancia de Estruch et al.}\\
  \hline
No. & Elementos &\multicolumn{2}{|c|}{$i=0$}&\multicolumn{2}{|c|}{$i=1$}&\multicolumn{2}{|c|}{$i=2$}& \multicolumn{2}{|c|}{$i=3$}\\ 
\cline{3-10}
pliegue & del pliegue & Aciertos &  \makebox[0.8cm][c]{\%}  & Aciertos &  \makebox[0.8cm][c]{\%} & Aciertos &  \makebox[0.8cm][c]{\%} & Aciertos &  \makebox[0.8cm][c]{\%} \\ 
 \hline\hline
0 & 31 & 26 & 83.87 & 31 & 100.00 & 31 & 100.00 & 31 & 100.00 \\ \hline
1 & 31 & 27 & 87.10 & 31 & 100.00 & 31 & 100.00 & 31 & 100.00 \\ \hline
2 & 31 & 25 & 80.65 & 30 & 96.77 & 30 & 96.77 & 30 & 96.77 \\ \hline
3 & 31 & 25 & 80.65 & 30 & 96.77 & 30 & 96.77 & 30 & 96.77 \\ \hline
4 & 31 & 28 & 90.32 & 30 & 96.77 & 31 & 100.00 & 31 & 100.00 \\ \hline
5 & 31 & 26 & 83.87 & 31 & 100.00 & 31 & 100.00 & 31 & 100.00 \\ \hline
6 & 31 & 28 & 90.32 & 31 & 100.00 & 31 & 100.00 & 31 & 100.00 \\ \hline
7 & 31 & 27 & 87.10 & 31 & 100.00 & 31 & 100.00 & 31 & 100.00 \\ \hline
8 & 31 & 25 & 80.65 & 31 & 100.00 & 31 & 100.00 & 31 & 100.00 \\ \hline
9 & 28 & 27 & 96.43 & 28 & 100.00 & 28 & 100.00 & 28 & 100.00 \\ \hline \hline
Total/prom. & 307 & 264 & 86.09 & 304 & 99.03 & 305 & 99.35 & 305 & 99.35 \\ \hline
\end{tabular}
\vspace{-0.2cm}
\caption{Aciertos  para el conjunto de datos {\em soybean} con jerarqu\'ia de similitud de los atributos utilizando la distancia de Estruch et al.}
\label{tab-soyb-jsim-de}
\end{center}
\end{table}

\begin{table}[H]\small
\begin{center}
\begin{tabular}{|c|r|r|r|r|r|r|r|r|r|}
  \hline
\multicolumn{10}{|c|}{Distancia eucl\'idea}\\
  \hline
No. & Elementos &\multicolumn{2}{|c|}{$i=0$}&\multicolumn{2}{|c|}{$i=1$}&\multicolumn{2}{|c|}{$i=2$}& \multicolumn{2}{|c|}{$i=3$}\\ 
\cline{3-10}
pliegue & del pliegue & Aciertos &  \makebox[0.8cm][c]{\%}  & Aciertos &  \makebox[0.8cm][c]{\%} & Aciertos &  \makebox[0.8cm][c]{\%} & Aciertos &  \makebox[0.8cm][c]{\%} \\ 
 \hline\hline
0 & 31 & 25 & 80.65 & 29 & 93.55 & 30 & 96.77 & 31 & 100.00 \\ \hline
1 & 31 & 22 & 70.97 & 25 & 80.65 & 29 & 93.55 & 29 & 93.55 \\ \hline
2 & 31 & 19 & 61.29 & 25 & 80.65 & 26 & 83.87 & 26 & 83.87 \\ \hline
3 & 31 & 19 & 61.29 & 28 & 90.32 & 29 & 93.55 & 29 & 93.55 \\ \hline
4 & 31 & 21 & 67.74 & 24 & 77.42 & 25 & 80.65 & 25 & 80.65 \\ \hline
5 & 31 & 24 & 77.42 & 26 & 83.87 & 26 & 83.87 & 28 & 90.32 \\ \hline
6 & 31 & 28 & 90.32 & 30 & 96.77 & 30 & 96.77 & 30 & 96.77 \\ \hline
7 & 31 & 25 & 80.65 & 26 & 83.87 & 30 & 96.77 & 30 & 96.77 \\ \hline
8 & 31 & 24 & 77.42 & 27 & 87.10 & 27 & 87.10 & 28 & 90.32 \\ \hline
9 & 28 & 24 & 85.71 & 27 & 96.43 & 28 & 100.00 & 28 & 100.00 \\ \hline \hline
Total/prom. & 307 & 231 & 75.35 & 267 & 87.06 & 280 & 91.29 & 284 & 92.58 \\ \hline
\end{tabular}
\vspace{-0.2cm}
\caption{Aciertos para el conjunto de datos {\em soybean} con jerarqu\'ia de similitud de los atributos utilizando la distancia eucl\'idea}
\label{tab-soyb-jsim-du}
\end{center}
\end{table}

\section{Conjunto de datos {\em Demospongiae}}

Los resultados para este conjunto de datos ({\em demospongiae}) son calculados utilizando las distancias de Nienhuys-Cheng y la distancia de Estruch et al.

\vspace{0.4cm}

\begin{table}[H]\small
\begin{center}
\begin{tabular}{|l|r|r|r|r|r|r|r|r|}
\cline{2-9}
\multicolumn{1}{c}{} &\multicolumn{8}{|c|}{Distancia de Nienhuys-Cheng}\\
\cline{2-9}
\multicolumn{1}{c}{} & \multicolumn{2}{|c|}{$i=0$}&\multicolumn{2}{|c|}{$i=1$}&\multicolumn{2}{|c|}{$i=2$}& \multicolumn{2}{|c|}{$i=3$}\\ 
\cline{2-9}
\multicolumn{1}{c}{} & \multicolumn{1}{|c|}{No. ejemplos} & \makebox[0.8cm][c]{\%} & No. ejemplos &  \makebox[0.8cm][c]{\%} & No. ejemplos &  \makebox[0.8cm][c]{\%} & No. ejemplos &  \makebox[0.8cm][c]{\%} \\ 
 \hline  \hline
Aciertos & 123 & 60.89 & 144 & 71.29 & 150 & 74.26 & 149 & 73.76 \\ \hline
Errores & 79 & 39.11 & 58 & 28.71 & 52 & 25.74 & 53 & 26.24 \\ \hline \hline
Total & 202 &  & 202 &  & 202 &  & 202 &  \\ \hline
\end{tabular}
\vspace{-0.2cm}
\caption{Aciertos y errores para el conjunto de datos {\em demospongiae} utilizando la distancia de Nienhuys-Cheng}
\label{tab-spong-dn}
\end{center}
\end{table}

\begin{table}[H]\small
\begin{center}
\begin{tabular}{|l|r|r|r|r|r|r|r|r|r|}
  \hline
\multicolumn{10}{|c|}{Distancia de Nienhuys-Cheng}\\
  \hline
\multirow{2}{*}{\makebox[2cm][c]{Clase}}  & \makebox[1.2cm][c]{No.} &\multicolumn{2}{|c|}{$i=0$}&\multicolumn{2}{|c|}{$i=1$}&\multicolumn{2}{|c|}{$i=2$}& \multicolumn{2}{|c|}{$i=3$}\\ 
\cline{3-10}
& \makebox[1.2cm][c]{ejemplos} & Aciertos &  \makebox[0.8cm][c]{\%}  & Aciertos &  \makebox[0.8cm][c]{\%} & Aciertos &  \makebox[0.8cm][c]{\%} & Aciertos &  \makebox[0.8cm][c]{\%} \\ 
 \hline\hline
\tiny{ASTROPHORIDA} & 36 & 23 & 63.89 & 30 & 83.33 & 30 & 83.33 & 29 & 80.56 \\ \hline
\tiny{AXINELLIDA} & 28 & 20 & 71.43 & 24 & 85.71 & 24 & 85.71 & 24 & 85.71 \\ \hline
\tiny{DICTYOCERATIDA} & 20 & 20 & 100.00 & 20 & 100.00 & 20 & 100.00 & 20 & 100.00 \\ \hline
\tiny{HADROMERIDA} & 48 & 28 & 58.33 & 29 & 60.42 & 29 & 60.42 & 29 & 60.42 \\ \hline
\tiny{HALICHONDRIDA} & 16 & 1 & 6.25 & 8 & 50.00 & 9 & 56.25 & 9 & 56.25 \\ \hline
\tiny{HAPLOSCLERIDA} & 18 & 12 & 66.67 & 12 & 66.67 & 15 & 83.33 & 14 & 77.78 \\ \hline
\tiny{POECILOSCLERIDA} & 36 & 19 & 52.78 & 21 & 58.33 & 23 & 63.89 & 24 & 66.67 \\ \hline \hline
Total & 202 & 123 &  & 144 &  & 150 &  & 149 &  \\ \hline
\end{tabular}
\vspace{-0.2cm}
\caption{Aciertos por cada clase para el conjunto de datos {\em demospongiae} utilizando la distancia de Nienhuys-Cheng}
\label{tab-spong-clases-dn}
\end{center}
\end{table}

\begin{table}[H]\small
\begin{center}
\begin{tabular}{|l|r|r|r|r|r|r|r|r|}
\cline{2-9}
\multicolumn{1}{c}{} &\multicolumn{8}{|c|}{Distancia de Estruch et al.}\\
\cline{2-9}
\multicolumn{1}{c}{} & \multicolumn{2}{|c|}{$i=0$}&\multicolumn{2}{|c|}{$i=1$}&\multicolumn{2}{|c|}{$i=2$}& \multicolumn{2}{|c|}{$i=3$}\\ 
\cline{2-9}
\multicolumn{1}{c}{} & \multicolumn{1}{|c|}{No. ejemplos} & \makebox[0.8cm][c]{\%} & No. ejemplos &  \makebox[0.8cm][c]{\%} & No. ejemplos &  \makebox[0.8cm][c]{\%} & No. ejemplos &  \makebox[0.8cm][c]{\%} \\ 
 \hline  \hline
 Aciertos & 123 & 60.89 & 148 & 73.27 & 159 & 78.71 & 157 & 77.72 \\ \hline
 Errores & 79 & 39.11 & 54 & 26.73 & 43 & 21.29 & 45 & 22.28 \\ \hline \hline
Total &  202 &  & 202 &  & 202 &  & 202 &  \\ \hline
\end{tabular}
\vspace{-0.2cm}
\caption{Aciertos y errores para el conjunto de datos {\em demospongiae} utilizando la distancia de Estruch et al.}
\label{tab-spong-de}
\end{center}
\end{table}

\begin{table}[H]\small
\begin{center}
\begin{tabular}{|l|r|r|r|r|r|r|r|r|r|}
  \hline
\multicolumn{10}{|c|}{Distancia de Estruch et al.}\\
  \hline
\multirow{2}{*}{\makebox[2cm][c]{Clase}}  & \makebox[1.2cm][c]{No.} &\multicolumn{2}{|c|}{$i=0$}&\multicolumn{2}{|c|}{$i=1$}&\multicolumn{2}{|c|}{$i=2$}& \multicolumn{2}{|c|}{$i=3$}\\ 
\cline{3-10}
& \makebox[1.2cm][c]{ejemplos} & Aciertos &  \makebox[0.8cm][c]{\%}  & Aciertos &  \makebox[0.8cm][c]{\%} & Aciertos &  \makebox[0.8cm][c]{\%} & Aciertos &  \makebox[0.8cm][c]{\%} \\ 
 \hline\hline
\tiny{ASTROPHORIDA} & 36 & 24 & 66.67 & 32 & 88.89 & 32 & 88.89 & 31 & 86.11 \\ \hline
\tiny{AXINELLIDA} & 28 & 20 & 71.43 & 23 & 82.14 & 23 & 82.14 & 23 & 82.14 \\ \hline
\tiny{DICTYOCERATIDA} & 20 & 20 & 100.00 & 20 & 100.00 & 20 & 100.00 & 20 & 100.00 \\ \hline
\tiny{HADROMERIDA} & 48 & 26 & 54.17 & 33 & 68.75 & 33 & 68.75 & 33 & 68.75 \\ \hline
\tiny{HALICHONDRIDA} & 16 & 0 & 0.00 & 4 & 25.00 & 8 & 50.00 & 8 & 50.00 \\ \hline
\tiny{HAPLOSCLERIDA} & 18 & 12 & 66.67 & 15 & 83.33 & 18 & 100.00 & 18 & 100.00 \\ \hline
\tiny{POECILOSCLERIDA} & 36 & 21 & 58.33 & 21 & 58.33 & 25 & 69.44 & 24 & 66.67 \\ \hline
\tiny{Total} & 202 & 123 &  & 148 &  & 159 &  & 157 &  \\ \hline
\end{tabular}
\vspace{-0.2cm}
\caption{Aciertos por cada clase para el conjunto de datos {\em demospongiae} utilizando la distancia de Estruch et al.}
\label{tab-spong-clases-dn}
\end{center}
\end{table}


\chapter{Descripci\'on de la implementaci\'on de las distancias entre t\'erminos}\label{implementacion}

En este ap\'endice se describe la implementaci\'on de las distancias de Nienhuys-Cheng y Estruch et al. utilizando XML como estructura de representaci\'on de datos.

\section {Distancia de Nienhuys-Cheng}

La distancia de Nienhuys-Cheng utiliza la profundidad de las ocurrencias de manera que las diferencias que est\'an cerca de los s\'imbolos ra\'iz cuentan m\'as que aquellas que no lo est\'an. Dadas dos expresiones, $s = s_0(s_1, \ldots, s_n)$ y $t =t_0(t_1, \ldots, t_n)$, esta distancia est\'a definida de la siguiente manera \cite{estruchnew}:

\[d_N(s, t) =\left\{
\begin{array} {ll}
0, & \mbox{si $s = t$}\\
1, & \mbox{si $\neg Compatible(s,t)$}\\
\frac{1}{2n}\sum_{i=1}^{n} d(s_i, t_i), & \mbox{otro caso}
\end{array}
\right.\]

\noindent Dos expresiones son incompatibles si los s\'imbolos ra\'iz o la aridad de las dos expresiones son diferentes.

Para implementar la distancia Nienhuys-Cheng entre dos ejemplos, representados en una estructura XML, es necesario recorrer (recursivamente) todos los elementos de un \'arbol e ir compar\'andolos con los elementos del segundo, aplicando la funci\'on de distancia. 

Inicialmente se determina el n\'umero de hijos de cada \'arbol y puede ocurrir una de las siguientes tres situaciones: 1) ninguno de los dos elementos tiene hijos,  2) los dos elementos tienen diferente n\'umero de hijos, \'o 3) los dos elementos tienen el mismo n\'umero de hijos. Para la primera situaci\'on, si el contenido de los dos elementos son iguales su distancia es 0; si por el contrario son diferentes, su distancia es 1; para el segundo caso, los elementos (o t\'erminos) son incompatibles y su distancia siempre es 1; en el tercer caso es necesario calcular recursivamente las distancias de cada elemento, sumarlas entre s\'i y finalmente multiplicarlas por el factor $\frac{1}{2n}$, donde $n$ es el n\'umero de hijos del elemento.

\begin{figure*}[h]
  \centering
   \subfloat[]{\label{fig:ejemplo1-jerarquias-XML}\includegraphics[width=0.57\textwidth]{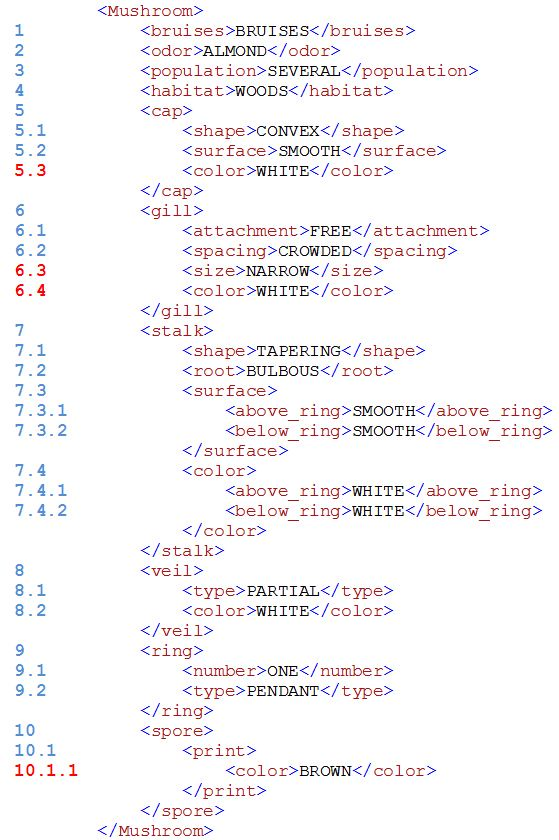}}
   \subfloat[]{\label{fig:ejemplo2-jerarquias-XML}\includegraphics[width=0.47\textwidth]{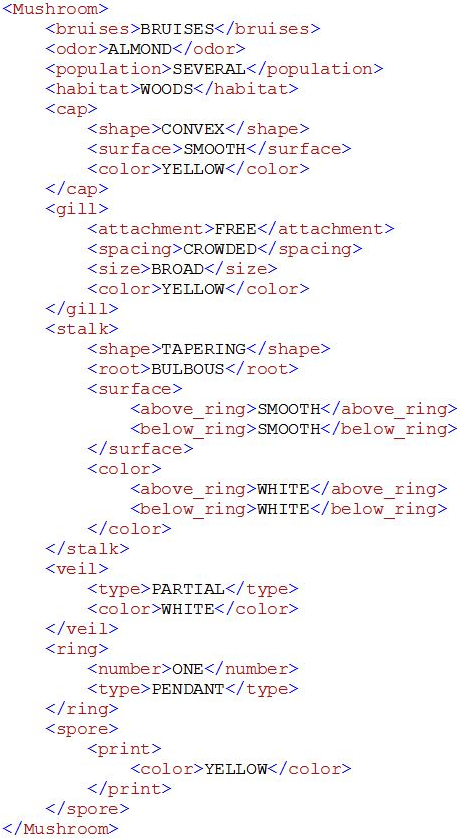}}  
   \caption{Ejemplo de dos jerarqu\'ias en XML para calcular distancias entre t\'erminos}
   \label{fig:ejemplo-jerarquias-XML}
   \vspace{-0.2cm}
 \end{figure*}

La figura \ref{fig:ejemplo-jerarquias-XML} muestra dos ejemplos del conjunto de datos de {\em mushroom} representados en XML. Debido a que la distancia entre dos elementos iguales es $0$, entonces se identifican \'unicamente los elementos que son diferentes; para el ejemplo de la figura \ref{fig:ejemplo-jerarquias-XML}: $size$ y tres elementos $color$ pertenecientes a $cap$, $gill$ y $spore$; luego se aplica la funci\'on de distancia sobre estos elementos (las distancias entre los elementos $color$, para los tres casos, son iguales a $1$; por esta raz\'on no se diferencian en este ejemplo; sin embargo, si el valor para alguno de estos elementos fuera una estructura, las distancias ser\'ian diferentes y deben calcularse cada una por separado):

\[
d_N(color_1, color_2)=d_N(size_1, size_2)=1
\]

\noindent Despu\'es de calculada la distancia para $color$ y $size$, es posible calcular las distancias para $cap$ y $gill$:  

\[
d_N(cap_1, cap_2)= \frac{1}{2\cdot 3}\bigl(d(color_1, color_2)\bigr)=\frac{1}{6}\cdot 1 = \frac{1}{6}
\]

\[
d_N(gill_1, gill_2)= \frac{1}{2\cdot 4}\Bigl(\bigl(d(size_1, size_2)\bigr)+\bigl(d(color_1, color_2)\bigr)\Bigr)=\frac{1}{8}\cdot (1+1) = \frac{1}{4}
\]

\noindent Para calcular la distancia de $spore$ primero se debe calcular $print$:

\[
d_N(print_1, print_2)=\frac{1}{2\cdot 1}\bigl(d(color_1, color_2)\bigr)=\frac{1}{2}\cdot 1 = \frac{1}{2}
\]

\[
d_N(spore_1, spore_2)=\frac{1}{2\cdot 1}\bigl(d(print_1, print_2)\bigr)=\frac{1}{2}\cdot \frac{1}{2} = \frac{1}{4}
\]

\noindent Finalmente, utilizando las distancias de $cap$, $gill$ y $spore$ se calcula la distancia entre estas dos instancias: 

\[
d_N(gill_1, gill_2)=\frac{1}{2\cdot 10}\Bigl(\bigl(d(cap_1, cap_2)\bigr)+\bigl(d(gill_1, gill_2)\bigr)+\bigl(d(spore_1, spore_2)\bigr)\Bigr) 
\]

\[
=\frac{1}{20}\cdot \biggl(\frac{1}{6} + \frac{1}{4} +  \frac{1}{4}\biggr)= \frac{1}{30}
\]

\section{Distancia de  Estruch et al.}

La distancia entre t\'erminos propuesta por Estruch et al. \cite{estruchnew} tiene en cuenta los siguentes aspectos: el contexto de las diferencias, la complejidad sint\'actica mediante el tama\~no de una expresi\'on y la ponderaci\'on asociada con las repeticiones utilizando relaciones de equivalencia. A continuaci\'on se describe la implementaci\'on de estos aspectos.

\begin{itemize}
\item \textbf{Diferencias sint\'acticas entre expresiones:}
\\ \\
La diferencia sint\'actica es un conjunto de ocurrencias incompatibles entre dos expresiones. Las ocurrencias son una secuencia de n\'umeros positivos (separados por puntos), encabezada por el s\'imbolo $\lambda$, que representan rutas de acceso en un \'arbol o un t\'ermino (por ejemplo, sea $t=p(a,f(b))$, entonces  el conjunto de ocurrencias $O(t)=\{\lambda, 1, 2, 2.1 \}$). 

Para implementar la diferencia sint\'actica, entre dos ejemplos, es necesario recorrer el documento XML y seleccionar las rutas en las cuales sus elementos son incompatibles y adicionarlos a una lista (de diferencias sint\'acticas). La incompatibilidad entre dos elementos se presenta cuando el n\'umero de hijos, etiquetas o valores en la misma ruta son diferentes; para el ejemplo de la figura \ref{fig:ejemplo-jerarquias-XML}, la lista de diferencias sint\'acticas es la siguiente:

\vspace{0.2cm}

\[
O(Mushroom_1, Mushroom_2)=\{ \lambda, 5.3, 6.3, 6.4, 10.1.1 \}
\]

\vspace{0.2cm}

\item \textbf{Relaciones de equivalencia de clases:}

Las diferencias repetidas entre los t\'erminos encontrados en la diferencia sint\'actica son definidas como relaciones de equivalencia. La implementaci\'on de estas equivalencias de clases est\'a definida por medio de la identificaci\'on de ocurrencias con repeticiones comunes (para las dos instancias) contenidas en la diferencia sint\'actica; es decir, todos los elementos de cada ejemplo encontrados en la diferencia sint\'actica, de acuerdo con su ruta, son comparados entre s\'i identificando sus repeticiones; posteriormente, estas repeticiones tambi\'en son comparadas para los dos ejemplos y en caso de coincidir, sus rutas son asignadas a una misma clase.  

Inicialmente se construye una matriz cuadrada por cada ejemplo; la dimensi\'on de esta matriz es igual al n\'umero de rutas existentes en la diferencia sint\'actica, donde cada fila y columna representa uno de estos elementos para un ejemplo determinado. Luego, se compara cada elemento con todos los dem\'as: si tanto la etiqueta como el contenido del elemento son iguales, entonces la posici\'on en la matriz correspondiente a estos elementos es marcada con 1, en caso contrario es marcada con 0. Cuando el elemento es comparado con \'el mismo (que corresponde a la diagonal de la matriz), este valor no ser\'a tenido en cuenta y por este motivo ser\'a marcado con el valor 0. Utilizando el ejemplo anterior (donde la diferencia sint\'actica es: $O(Mushroom_1, Mushroom_2)=\{ \lambda, 5.3, 6.3, 6.4, 10.1.1 \}$, entonces se crean dos matrices como se muestra en la figura \ref{fig:matriz-repeticiones} indicando cu\'ales elementos son iguales:

\vspace{0.2cm}

\begin{figure}[h]\small
\centering
\subfloat[]{\label{fig:matriz-rep1}
\begin{tabular}{|l|c|c|c|c|}\hline
 & 5.3 & 6.3 & 6.4 & 10.1.1 \\
\hline\hline
5.3  & 0 & 0 & 1 & 0   \\
\hline
6.3  & 0 & 0 & 0 & 0   \\
\hline
6.4  & 1 & 0 & 0 & 0   \\
\hline
10.1.1  & 0 & 0 & 0 & 0   \\
\hline
\end{tabular}}
\hspace{0.2cm}
\subfloat[]{\label{fig:matriz-rep2}
\begin{tabular}{|l|c|c|c|c|}\hline
 & 5.3 & 6.3 & 6.4 & 10.1.1 \\
\hline\hline
5.3  & 0 & 0 & 1 & 1   \\
\hline
6.3  & 0 & 0 & 0 & 0   \\
\hline
6.4  & 1 & 0 & 0 & 1   \\
\hline
10.1.1  & 1 & 0 & 1 & 0   \\
\hline
\end{tabular}}
\caption{Matriz de repeticiones para las jerarquias de la figura \ref{fig:ejemplo-jerarquias-XML} }
\label{fig:matriz-repeticiones}
\end{figure}

Para la matriz del ejemplo \ref{fig:matriz-rep2} los elementos $5.3$, $6.4$ y $10.1.1$ son iguales (en nombre de la etiqueta y valor). Por otra parte, para la matriz del ejemplo \ref{fig:matriz-rep1}, \'unicamente los elementos $5.3$ y $6.4$ son iguales. Utilizando las matrices de los ejemplos anteriores, es posible construir una tercera matriz que contenga como resultado las repeticiones comunes. Para construir esta matriz resultante es necesario comparar cada celda y s\'olo ser\'an marcadas con 1 aquellas que, en ambos casos, contengan el valor 1; es decir, las repeticiones son las mismas para ambos ejemplos. Utilizando las tablas de la figura \ref{fig:matriz-repeticiones}, \'unicamente ser\'ian marcadas las celdas $5.3$ y $6.4$. 

\vspace{0.2cm}

\begin{table}[H]\small
\begin{center}
\begin{tabular}{|l|c|c|c|c|}\hline
 & 5.3 & 6.3 & 6.4 & 10.1.1 \\
\hline\hline
5.3  & 0 & 0 & 1 & 0   \\
\hline
6.3  & 0 & 0 & 0 & 0   \\
\hline
6.4  & 1 & 0 & 0 & 0   \\
\hline
10.1.1  & 0 & 0 & 0 & 0   \\
\hline
\end{tabular}
\caption{Matriz de repeticiones com\'unes de la figura \ref{fig:matriz-repeticiones}}
\label{tab-rep-comunes}
\end{center}
\end{table}

Utilizando la matriz resultante \ref{tab-rep-comunes} es posible asignar cada ruta dentro de una clase; la manera de hacerlo es colocando en una misma clase todas las rutas con repeticiones comunes; luego las rutas de los elementos restantes de la diferencia sint\'actica son asignadas, cada una, en una clase aparte. La tabla \ref{tab-ubicacion-clase} muestra la aisgnaci\'on de clases para cada ruta:   

\vspace{0.2cm}

\begin{table}[H]\small
\begin{center}
\begin{tabular}{|c|c|}\hline
 Ruta & Clase \\
\hline\hline
5.3  & 1   \\
\hline
6.4  & 1   \\
\hline
6.3  & 2  \\
\hline
10.1.1 & 3   \\
\hline
\end{tabular}
\caption{Rutas por cada clase utilizando la matriz \ref{tab-rep-comunes}}
\label{tab-ubicacion-clase}
\end{center}
\end{table}

Esta tabla de equivalencias de clases \ref{tab-ubicacion-clase} es guardada en una estructura en memoria y, posteriormente, utilizada para calcular la ponderaci\'on de las ocurrencias.

\item \textbf{Tama\~no de una expresi\'on}:

El tama\~no de una expresi\'on $t = t_0(t_1,\dots,t_n)$ est\'a definido por la siguiente funci\'on \cite{estruchnew}:

\vspace{0.2cm}

\[
Size'(t)=\frac{1}{4}Size(t)
\]

donde:

\[
Size(t_0(t_1,\ldots,t_n))= \left \{
\begin{array}{l}
1,\,n=0 \\
1+\frac{\sum_{i=1}^n Size(t_i)}{2(n+1)}, \,n>0
\end{array}\right .
\]

\vspace{0.2cm}

$Size$ es una funci\'on recursiva que se implementa obteniendo el n\'umero de hijos de un elemento; si el elemento no tiene hijos, su tama\~no es 1; por el contrario, si el elemento tiene hijos, se calcula recursivamente el tama\~no de todos los elementos hijos y se suman entre s\'i; luego, el valor obtenido divide por el factor $2(n+1)$, donde $n$ es el n\'umero de hijos y, finalmente, se le suma 1 a este valor. $Size'$ es calculado f\'acilmente dividiendo el valor de  $Size$ entre $4$.

Los tama\~nos son calculados para todos los elementos de las dos instancias comparadas, relacionados en la lista de diferencias sint\'acticas y se almacenan en memoria para ser utilizados posteriormente. La tabla \ref{tab-size} muestra los tama\~nos $Size'$ para los elementos de la tabla \ref{tab-ubicacion-clase}.

\begin{table}[H]\small
\begin{center}
\begin{tabular}{|c|c|c|c|}\hline
 Ruta & Clase & $size'$ & $size'$  \\
 & & ejemplo 1 & ejemplo 2  \\
\hline\hline
5.3  & 1 & 0.25 & 0.25  \\
\hline
6.4  & 1 & 0.25 & 0.25   \\
\hline
6.3  & 2 & 0.25 & 0.25 \\
\hline
10.1.1 & 3 & 0.25 & 0.25   \\
\hline
\end{tabular}
\caption{Tama\~no de los elementos relacionados en la tabla \ref{tab-ubicacion-clase}}
\label{tab-size}
\end{center}
\end{table}

\item \textbf{Valor del contexto de ocurrencia}:

El valor de contexto de ocurrencia de una expresi\'on $t$, denotado por $C(o;t)$, es definido de la siguiente manera \cite{estruchnew}:

\[
C(o;t)=\left \{
\begin{array}{l}
1,\,o=\lambda \\
2^{Length(o)}\cdot\prod_{\forall o' \in Pre(o)}(Arity(t|_{o'})+1),\,\textrm{otro caso} \\
\end{array}\right .
\]

\vspace{0.2cm}

donde la longitud de una ocurrencia, $Length(o)$, es el n\'umero de \'items en $o$. Por ejemplo, $Length(5.3) = 2$, $Length(10.1.1) = 3$.

El valor del contexto de ocurrencia se refiere a las relaci\'on entre un subt\'ermino $t|_0$ y $t$ en el sentido en que, para un valor alto de $C(o;t)$ corresponde a una posici\'on profunda de $t|_0$ en t o la existencia de supert\'erminos de $t|_0$ con un gran n\'umero de argumentos \cite{estruchnew}.

Para implementar el valor del contexto, por cada ocurrencia encontrada en la diferencia sint\'actica, se recorre la rama del \'arbol ascendentemente calculando la aridad de cada padre y sum\'andole $1$; luego, todas las aridades son multiplicadas entre s\'i;  finalmente, se multiplica el valor anterior por $2^{Length(o)}$. Para aquellos casos en que la ocurrencia sea $\lambda$, de acuerdo con la definici\'on, el valor de contexto ser\'a 1. 

La tabla \ref{tab-contexto} muestra el valor de contexto calculado para los elementos de la tabla \ref{tab-ubicacion-clase}:

\vspace{0.2cm}

\begin{table}[H]\small
\begin{center}
\begin{tabular}{|c|c|c|c|c|}\hline
 Ruta & Clase & $size'$ & $size'$ & Valor \\
 & & ejemplo 1 & ejemplo 2 &  de contexto \\
\hline\hline
5.3  & 1 & 0.25 & 0.25 & 176\\
\hline
6.4  & 1 & 0.25 & 0.25  & 220\\
\hline
6.3  & 2 & 0.25 & 0.25 & 220\\
\hline
10.1.1 & 3 & 0.25 & 0.25 & 352  \\
\hline
\end{tabular}
\caption{Valor de contexto de los elementos relacionados en la tabla \ref{tab-ubicacion-clase}}
\label{tab-contexto}
\end{center}
\end{table}

\item \textbf{Ponderaci\'on de las ocurrencias}:

La funci\'on de ponderaci\'on $w$ es una funci\'on que, dada dos expresiones $s$ y $t$, se define de la siguiente manera \cite{estruchnew}:

 \[\forall o\in O^\star_i(s,t),\,\, w(o)=\frac{3f_i(o)+1}{4f_i(o)}\]

Esta funci\'on asocia ponderaciones a las ocurrencias de manera que, mientras mayor sea el valor de contexto de una ocurrencia, menor es la ponderaci\'on asignada a esta; es decir, la diferencia sint\'actica menos significativa que se refiere a la ocurrencia \cite{estruchnew}.

Para la implementaci\'on de la ponderaci\'on, por cada ocurrencia relacionada en la diferencia sint\'actica, primero se calcula el valor de $f_i(o)$ que corresponde al valor de posici\'on de cada ocurrencia dentro de la clase obtenida en la relaci\'on de equivalencia. En el ejemplo anterior,  $f_1(5.3) = 1$ y $f_1(6.4) = 2$, debido a que la equivalencia de clase $1$ contiene dos ocurrencias; para la clase $2$ y $3$, que solo tienen una ocurrencia, entonces $f_2(6.3) = f_3(11.1.1) = 1$.

Despu\'es de calculado  el valor $f_i(o)$, se aplica la funci\'on de ponderaci\'on. La tabla \ref{tab-ponderacion} muestra la ponderaci\'on calculada utilizando las relaciones de equivalencia obtenidas en la tabla \ref{tab-ubicacion-clase}.

\vspace{0.2cm}

\begin{table}[H]\small
\begin{center}
\begin{tabular}{|c|c|c|c|c|c|}\hline
 Ruta & Clase & $size'$ & $size'$ & Valor & Ponderaci\'on\\
 & & ejemplo 1 & ejemplo 2 &  de contexto & \\
\hline\hline
5.3  & 1 & 0.25 & 0.25 & 176 & 1\\
\hline
6.4  & 1 & 0.25 & 0.25  & 220 & 0.875\\
\hline
6.3  & 2 & 0.25 & 0.25 & 220 & 1\\
\hline
10.1.1 & 3 & 0.25 & 0.25 & 352 & 1\\
\hline
\end{tabular}
\caption{Ponderaci\'on de las ocurrencias de los elementos relacionados en la tabla \ref{tab-ubicacion-clase} }
\label{tab-ponderacion}
\end{center}
\end{table}

\item \textbf{Distancia de Estruch et al.}:

Dada dos expresiones $s$ y $t$, la distancia de Estruch et al. se define de la siguiente manera:

\vspace{0.2cm}

 \[
d_E(s,t) = \sum_{o \in O^\star(s,t)} \frac{w(o)}{C(o)}\bigl (size'(s|_{o})+size'(t|_{o})\bigr )
\]

\vspace{0.2cm}

Finalmente, para calcular la distancia de Estruch et al. se suman los tama\~nos de todos los elemntos relacionados en la diferencia sint\'actica, se multiplican por la ponderaci\'on y se dividen por el valor de contexto; para obtener la distancia, se suman estos valores entre s\'i. Para el ejemplo anterior (mostrado en la tabla \ref{tab-ponderacion}), la distancia es igual a $0.00852272$.

\end{itemize}

\end{document}